\documentclass{article}

\usepackage{microtype}
\usepackage{graphicx}
\usepackage{subcaption}
\usepackage{booktabs} % for professional tables

\usepackage{hyperref}

\usepackage[accepted]{icml2026}

\usepackage{amsmath}
\usepackage{amssymb}
\usepackage{mathtools}
\usepackage{amsthm}

\usepackage[capitalize,noabbrev]{cleveref}

\usepackage{adjustbox}
\usepackage{xcolor,colortbl}
\usepackage{caption}
\usepackage{subcaption}
\usepackage{amsmath}
\usepackage{bbm}
\usepackage{xspace}
\usepackage{enumitem}
\usepackage{pifont}
\def\ours{MoRAM\xspace}

\definecolor{lightgrey}{rgb}{0.95, 0.95, 0.95}
\definecolor{softblue}{rgb}{0.22, 0.59, 1.00}
\definecolor{softorange}{rgb}{1.00, 0.52, 0.00}
\definecolor{softgreen}{rgb}{0.45, 0.62, 0.5}
\definecolor{lightorange}{RGB}{255, 223, 186}

\usepackage{multirow}
\usepackage{colortbl}
\usepackage[dvipsnames]{xcolor}
\usepackage{makecell}
\usepackage{rotating}

\usepackage{amsmath,amsfonts,bm}

\def\eqref#1{equation~\ref{#1}}
\def\Eqref#1{Eq.~(\ref{#1})}
\def\1{\bm{1}}

\def\rs{{\textnormal{s}}}

\def\rw{{\textnormal{w}}}

\def\rvx{{\mathbf{x}}}

\DeclareMathAlphabet{\mathsfit}{\encodingdefault}{\sfdefault}{m}{sl}
\SetMathAlphabet{\mathsfit}{bold}{\encodingdefault}{\sfdefault}{bx}{n}

\newcommand{\softmax}{\mathrm{softmax}}

\usepackage{xcolor}         % colors
\usepackage[percent]{overpic}
\definecolor{RowCOLOR}{RGB}{230,230,250}
\definecolor{CellCOLOR}{RGB}{250,230,245}
\newcommand{\highblue}[1]{{\textbf{\color[RGB]{30, 85, 170}#1}}}
\newcommand{\highred}[1]{{\textbf{\color[RGB]{220, 20, 60}#1}}}
\def\eg{\textit{e.g.}}
\def\ie{\textit{i.e.}}

\theoremstyle{plain}
\newtheorem{theorem}{Theorem}[section]

\theoremstyle{definition}
\newtheorem{definition}[theorem]{Definition}

\theoremstyle{remark}
\newtheorem{remark}[theorem]{Remark}

\usepackage[textsize=tiny]{todonotes}

\icmltitlerunning{Little By Little: Continual Learning via Incremental Mixture of Rank-1 Associative Memory Experts}

\begin{document}

\twocolumn[
  \icmltitle{Little by Little: Continual Learning via \\Incremental Mixture of Rank-1 Associative Memory Experts}

  \icmlsetsymbol{equal}{*}

  \begin{icmlauthorlist}
    \icmlauthor{Haodong Lu}{unsw,csiro}
    \icmlauthor{Chongyang Zhao}{unsw}
    \icmlauthor{Minhui Xue}{csiro}
    \icmlauthor{Lina Yao}{unsw}
    \icmlauthor{Kristen Moore}{csiro}
    \icmlauthor{Dong Gong}{unsw}
\end{icmlauthorlist}

\icmlaffiliation{unsw}{University of New South Wales, Sydney, Australia}
\icmlaffiliation{csiro}{CSIRO}

\icmlcorrespondingauthor{Dong Gong}{dong.gong@unsw.edu.au}

  \icmlkeywords{Machine Learning, ICML, Continual Learning, CLIP, LLM, MoE, LoRA, Memory}

  \vskip 0.3in
]

\printAffiliationsAndNotice{}  % no special notice (required even if empty)

\begin{abstract}
Continual learning (CL) with large pre-trained models aims to incrementally acquire knowledge without catastrophic forgetting.
Existing LoRA-based Mixture-of-Experts (MoE) methods expand capacity by adding isolated new experts while freezing old ones, but still suffer from redundancy, interference, routing ambiguity, and consequent forgetting. 
We investigate the issues stemming from \emph{coarse-grained} expert granularity. Coarse-grained experts (\eg, high-rank LoRA) encode low-specialty information, leading to expert duplication/interference and routing degradation/confusion as experts accumulate. 
In this work, we propose \textbf{\ours} (\textbf{M}ixture \textbf{o}f \textbf{R}ank-1 \textbf{A}ssociative \textbf{M}emory). 
Grounded in the view that weight matrices act as {linear associative memories}, \ours achieves CL as incremental expansion of reusable \emph{atomic rank-1 experts as memory}. Each rank-1 adapter acts as a fine-grained \emph{MoE} expert or an associative \emph{memory} unit. 
By viewing rank-1 experts as key-value memory pairs, we eliminate explicit MoE-LoRA routers with self-activation, where each memory atom evaluates its relevance via its intrinsic key.
The inference process thus becomes a content-addressable retrieval and recall over the incrementally accumulated memory of learning snapshots. 
Extensive experiments on CLIP and LLMs show that \ours significantly outperforms state-of-the-art methods, achieving a better plasticity–stability trade-off, stronger generalization, and reduced forgetting.
Project page: \href{https://artificer-ai-lab.github.io/MoRAM/}{\texttt{artificer-ai-lab.github.io/MoRAM}}.
\end{abstract}
\vspace{-0.3cm}
~~~~~\textit{``Little by little, we gave you everything you ever dreamed of ...'' --- ``Little by little'', Oasis.}
\vspace{0.3cm}

\section{Introduction}
\label{sec:intro}
Continual learning (CL) \citep{hadsell2020embracing,de2021continual,ding2022don} aims to enable models to incrementally and efficiently acquire new knowledge from a stream of tasks and data, without catastrophic forgetting \citep{mccloskey1989catastrophic,nguyen2019toward} or the need for repeated fine-tuning on all previously seen data \citep{l2p,wang2024self}. 
With the ascendancy of Large Pre-trained Models (PTMs) in vision \citep{dosovitskiy2020image,clip} and language \citep{raffel2020exploring,grattafiori2024llama}, a specific challenge has arisen: \emph{how do we efficiently insert specific new capabilities into a massive, static memory structure without disrupting its existing generalized knowledge?}

\begin{figure*}[t]
    \centering
    % \vspace{-0.3cm}
    \includegraphics[width=0.94\linewidth]{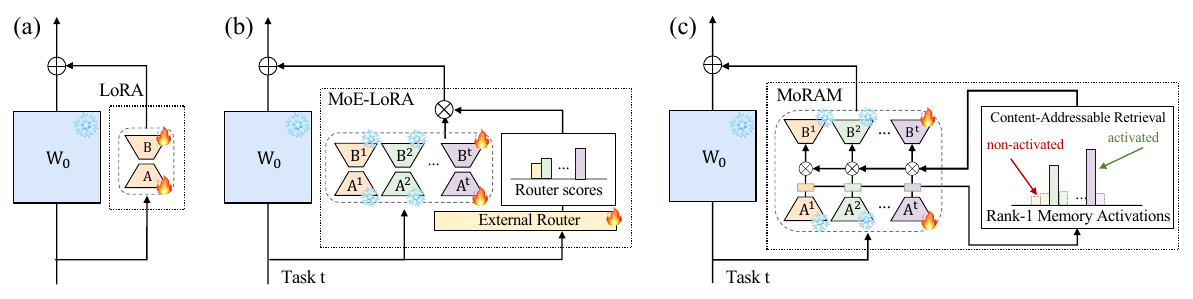}
    % \vspace{-0.13cm}
    \caption{Conceptual illustration of CL with (a) LoRA, (b) MoE-LoRA, and (c) \ours (Ours).}
    \label{fig:concept}
    % \vspace{-0.2cm}
\end{figure*}

A dominant approach for adapting PTMs is Parameter-Efficient Fine-Tuning (PEFT), particularly Low-Rank Adaptation (LoRA, Fig.~\ref{fig:concept}(a)) \citep{lora}. While efficient, standard LoRA applies updates as ``dense'' modifications to the weight space. This creates a fundamental conflict: to learn a new task, the method blindly and inevitably alters the representation space used by previous tasks, leading to interference and forgetting \citep{lora_learn_less}. 
To mitigate this, recent works have adopted Mixture-of-Experts (MoE) frameworks with LoRA adapters as experts (Fig.~\ref{fig:concept}(b)) \citep{dou2023loramoe,wu2024mixture}, an approach now widely used in CL \citep{boosting,wang2024self,yang2024moral,chen2024llava,theory}.
These works either pre-define an MoE with LoRA for CL \citep{yang2024moral}, or incrementally add experts \citep{wang2024self} or task-specific routers \citep{boosting}, assuming MoE benefits CL by isolating task interference. Such methods \citep{rusu2016progressive,wang2024self,qiao2024learn,boosting} freeze old components and add new ones (e.g., experts or routers) to reduce forgetting. Despite design differences, we collectively refer to a plain and general design with LoRA-based MoE as MoE-LoRA. In MoE-LoRA models, each expert is a LoRA adapter with pre-defined ranks (in subspaces), and the router selects among experts with each LoRA adapter as a unit.

However, current MoE-LoRA approaches operate at a \emph{coarse granularity} that limits their effectiveness in CL. 
Each expert is a multi-rank LoRA adapter treated as a single unit, yielding coarse-grained routing and learning.
This forces each expert to capture a broad range of information, resulting in weaker specialization and limited combinatorial expressivity \cite{he2024mixture,ludziejewski2024scaling,dai2024deepseekmoe}. 
In continual learning with incrementally and dynamically added experts, coarse-grained MoE introduces three key challenges:

\noindent\textbf{(1) Interference:} A coarse expert contains a mix of low-specialty knowledge. Activating coarse-grained experts (\ie, LoRA adapters with large rank) for a specific input inevitably triggers irrelevant subspaces, causing interference.

\noindent\textbf{(2) Redundancy:} New experts cannot selectively reuse specific ``atoms'' of knowledge from old experts; if the limited combination cannot cover the new task, they must relearn entire blocks, leading to inefficient capacity usage.

\noindent\textbf{(3) Routing/Retrieval Collapse:} The low-specialization of coarse-grained experts further confuses the router. 
As the number of experts grows, routers increasingly struggle to reliably index the expanding expert pool (\ie, routing ambiguity), leading to routing drift and collapse of old experts \cite{zhao2026token}, and ultimately accelerating forgetting. 

The coarse-grained experts and these resulting challenges limit the potential of the promising MoE-LoRA design.

To address the limitations, we fundamentally rethink adaptation by synthesizing the intrinsic dimensionality of PTMs \citep{aghajanyan2020intrinsic} through the lens of Linear Associative Memory \citep{kohonen1972correlation}. While model weights reside in a high-dimensional parameter space, they operate on a low-rank manifold, effectively functioning as a composition of atomic key-value pairs of a memory system \cite{kohonen1972correlation}. 
From this perspective, the optimal continual learning update is not a dense, persistent modification of weights or coarse-grained adapters, but an associative memory of \emph{expandable}, \emph{test-time retrievable} atomic units.
Conventional dense updates alter the model’s global memory structure, more easily causing catastrophic forgetting. MoE-LoRA offers an alternative but remains limited by brittle coarse-grained expert expansion and retrieval.

From a memory-augmentation perspective, we cast \emph{continual learning as maintaining and expanding a parametric memory} — a growing collection of fine-grained, atomic units stored outside the base model's weights — and cast \emph{inference as input-specific retrieval from this memory}. 
Each memory unit encodes an incremental learning snapshot, and relevant units are retrieved at test time to specialize the model for the current input, \ie, recall and reuse of memorized snapshots.

Based on this insight, we propose \textbf{\ours} (\textbf{M}ixture \textbf{o}f \textbf{R}ank-1 \textbf{A}ssociative \textbf{M}emory), a framework that continually updates PTMs 
by incrementing an associative memory system with atomic rank-1 experts ``\emph{little by little}'' (Fig. \ref{fig:concept}(c)). 
The associative memory can also be seen as a mixture-of-expert model with rank-1 adapters as experts. Each rank-1 adapter acts as a fine-grained MoE expert or an atomic memory unit.
By treating rank-1 experts (each comprising two vectors) as key–value memory pairs, we naturally eliminate the need for explicit routers in MoE-LoRA, with a proposed self-activation mechanism. 
Each expert can thus evaluate its own relevance to the input via its intrinsic key.
This shifts the paradigm from \textit{address-based routing} (learning where to send data) to \textit{content-addressable retrieval} (inputs automatically triggering the correct memory). By eliminating explicit routers, it avoids routing collapse and forgetting, improving retrieval reliability, scalability, and efficiency.

Our contributions are as follows: 
(1) 
We introduce \ours, a novel CL framework that treats learning as expanding a parametric memory of rank-1 associative units and inference as test-time memory retrieval, enabling fine-grained knowledge reuse with minimal interference.
(2) 
With rank-1 memory atoms, the model naturally forms a fine-grained MoE. We propose a sparse self-activation routing mechanism, eliminating external routers and enabling sparse content-based retrieval.
(3) We conduct extensive evaluations on both vision–language (CLIP) and large language model (LLM) benchmarks. Empirical results demonstrate that \ours significantly outperforms state-of-the-art methods, achieving superior plasticity-stability trade-offs, while effectively preserving pre-trained generalization capabilities.

\section{Related Work}
\noindent\textbf{Continual learning}  
enables sequential knowledge acquisition without forgetting. 
Experience replay (ER) methods \citep{chaudhry2018riemannian,chaudhry2018efficient,aljundi2019gradient,liu2020mnemonics,der,yan2022learning,luo2023class,tong2025coreset,tong2026model} interleave past examples with new data. 
Parameter regularization \citep{kirkpatrick2017overcoming,zenke2017continual,aljundi2018memory,aljundi2019task,npcl,zhao2024learning} penalizes updates to critical weights. 
Dynamic networks \citep{wang2022beef,wang2022foster,zhou2022model,l2p,sprompts,dualprompt,codaprompt,wang2024self,mcdonnell2024ranpac,inflora,ease,zhao2026token} allocate new capacity on the fly and preserve dedicated pathways for prior tasks.

\noindent\textbf{Continual learning of PTMs.}  
For CL on vision–language CLIP model \citep{garg2023tic,jha2024clapclip,zhang2024overcoming}, methods like ZSCL \citep{zscl} retain zero-shot performance during adaptation, and follow-up work \citep{boosting,rail,codyra,wu2025sd,tang2025mind} continually fine-tunes while leveraging frozen pre-trained predictions. The X-TAIL benchmark \citep{rail} further challenges models by mixing domain labels at test time. In language models (LMs) \citep{de2019episodic,qin2021lfpt5,razdaibiedina2023progressive,wang2023orthogonal,qiao2024learn}, continual learning uses capacity expansion or task-specific submodules to reduce interference. 

\noindent\textbf{Low‐rank adaptation}  
(LoRA) \citep{lora} is widely used for parameter-efficient fine-tuning of large pre-trained models. Building on this foundation, recent methods have reformulated LoRA’s updates via SVD-based initialization and dynamic rank scheduling \citep{sora,adalora,alora,moslora,milora,pissa}, demonstrating that task adaptation primarily relies on finding a compact subspace. In this work, we offer a complementary perspective by treating both the pre-trained weight matrix and its low-rank updates through the lens of linear associative memory \citep{kohonen1972correlation,anderson1972simple,li2018measuring,aghajanyan2020intrinsic}. Under this formulation, a rank-$r$ update corresponds to $r$ new memory entries into the matrix, where each rank-1 component is an atomic memory slot.

\noindent\textbf{Mixture-of-Experts (MoE) with LoRA.}
MoE scales capacity by routing inputs to sparse expert subnetworks via load-balancing \citep{shazeer2017outrageously,lepikhin2020gshard,fedus2022switch,dai2024deepseekmoe}. This paradigm has been adapted for fine-tuning \citep{dou2023loramoe,chen2023octavius,li2024mixlora,zhou2025same} and CL \citep{boosting,wang2024self,yang2024moral,chen2024llava} by treating adapters as experts, typically frozen per task to prevent forgetting. In contrast, we decompose rank-$r$ updates into $r$ atomic rank-1 components and compute an input-dependent mixture over these fine-grained experts, significantly enhancing specialization and diversity.

\section{Methods}
\subsection{Preliminaries}
\label{sec:preliminaries}

\noindent\textbf{Continual learning.} In CL, a model sequentially learns $T$ tasks. For task $t\in \{1,\dots,T\}$, let $\mathcal{D}^t={\{}(\mathbf{x}^t_i,y^t_i){\}}_{i=1}^{N^t}$, where $\mathbf{x}^t_i\in\mathbb{R}^{n\times d}$, $y^t_i\in\mathcal{C}^t$, and $N^t$ is the number of examples. In the memory-free setting, the model may access only $\mathcal{D}^t$ and cannot access data from any $\mathcal{D}^{u}$ with $u<t$.

\noindent\textbf{Low-rank adaptation.}
LoRA \citep{lora} parameterizes a low‐rank update to a pre‐trained weight matrix $\mathbf W_0\in\mathbb R^{d_\text{out} \times d_\text{in}}$ by introducing two factors
$\mathbf{B} \in \mathbb{R}^{d_\text{out} \times r}$ and $\mathbf{A} \in \mathbb{R}^{r \times d_\text{in}}$, such that $\Delta \mathbf{W} = \mathbf{B} \mathbf{A}$, where $r \ll \min(d_\text{in},d_\text{out})$. The updated weight matrix is then defined as:
\begin{equation}
\label{eq:lora}
\mathbf{W} = \mathbf{W}_0 + \Delta \mathbf{W} = \mathbf{W}_0 + \mathbf{B} \mathbf{A}.
\end{equation}
In this formulation, the original weights $\mathbf{W}_0$ remain fixed, and only $\mathbf{B}$ and $\mathbf{A}$ are trained, reducing the number of trainable parameters from $d_\text{in}d_\text{out}$ to $r(d_\text{in}+d_\text{out})$. 
While parameter-efficient, standard LoRA applies the rank-$r$ update densely to every input. This would result in updates for a new task inevitably perturbing the subspaces used by prior tasks, leading to interference.

\noindent\textbf{Mixture‐of‐Experts LoRA.}  
Building on the Mixture‐of‐Experts (MoE) paradigm, a common generic framework of MoE‐LoRA \citep{boosting,wang2024self} treats each LoRA as an independent expert.
Suppose after \(T\) tasks, we have \(T\) LoRA experts 
$\{(\mathbf A^{1},\mathbf B^{1}),\dots,(\mathbf A^{T},\mathbf B^{T})\}$. For an input token \(x\in\mathbb R^{d_{\rm in}}\), the overall LoRA update in this framework is given by  
$ \Delta\mathbf W = \sum\nolimits_{i=1}^T R(x)_i \mathbf B^i\,\mathbf A^i,$
where the mixture weight \(R(x)\in\mathbb R^T\) is produced by a learned router  
$R(\cdot) = \softmax(x \mathbf{W}_r)$
and \(\mathbf W_r\in\mathbb R^{d_{\rm in}\times T}\) contains the router’s trainable parameters.  Each LoRA's contribution is weighted by the learnable router, enabling the model to dynamically select and combine the most relevant low‐rank updates for each token. In practice, a \texttt{top-k} masking is typically applied to mixture weights to enforce sparsity, activating only the \(k\) most relevant experts. However, this design remains coarse-grained. Each expert is a dense rank-$r$ block, and the external router $R(\mathbf{x})$ must learn to map inputs to these overlapping experts. As the number of tasks grows, this external mapping becomes ambiguous, leading to routing collapse and forgetting.

\begin{figure*}[t]
    \centering
    % \vspace{-0.25cm}
    \includegraphics[width=\linewidth]{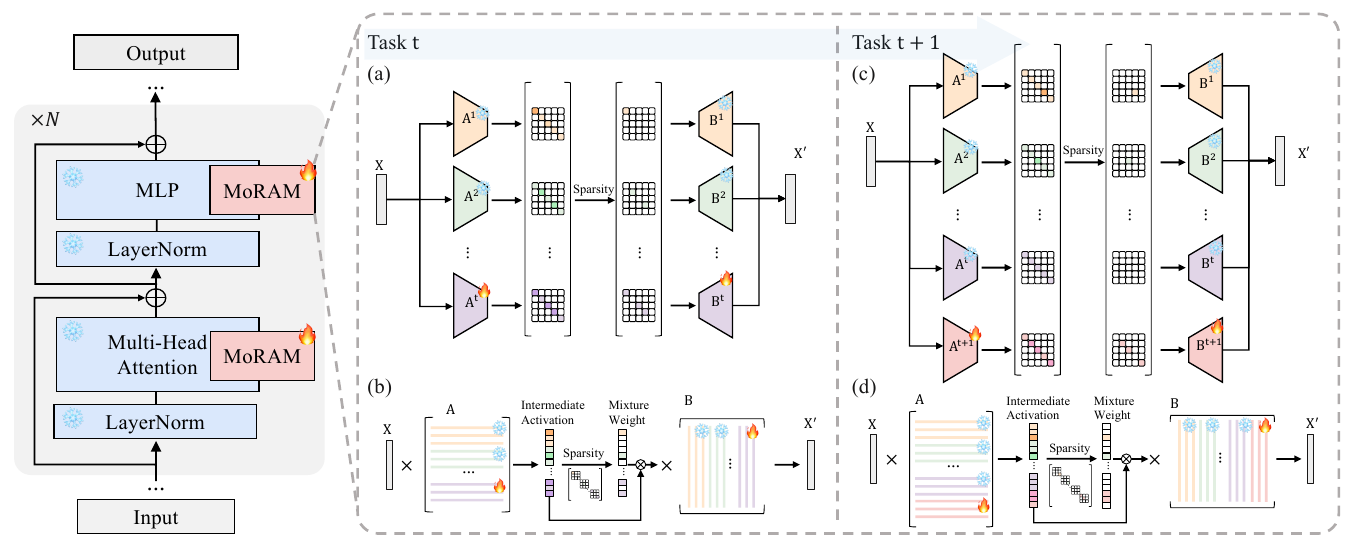}
    % \vspace{-0.15cm}
    \caption{Overview of \ours. For each new task, we freeze the atoms learned on previous tasks and introduce $r$ new rank-1 updates. Our sparse self-activated mixture‐of‐ranks framework jointly considers all old and new atoms, adaptively inferring a sparse mixture weight for each atom. Panels (a,c) illustrate \ours conceptually and (b,d) detail its computation for tasks $t$ and $t+1$, respectively.}
    \label{fig:overview}
    % \vspace{-0.2cm}
\end{figure*}

\subsection{Weights as Linear Associative Memory}
\label{sec:associative_memory}

\noindent To resolve the limitations of coarse granularity and routing ambiguity inherent in dense rank-$r$ updates, we depart from the conventional view of LoRA as monolithic blocks, instead reconceptualizing both the weight matrix and its updates as a \emph{Linear Associative Memory}.

\begin{definition}[Weight Matrix as Linear Associative Memory]
\label{def:assoc_mem}
Consider a weight matrix $\mathbf{W} \in \mathbb{R}^{d_{\text{out}} \times d_{\text{in}}}$ of rank $m$. Through the lens of linear associative memory \citep{kohonen1972correlation,anderson1972simple}, it acts as a retrieval system composed of $m$ atomic key-value pairs $\{(\mathbf{k}_i, \mathbf{v}_i)\}_{i=1}^m$, where $\mathbf{k}_i \in \mathbb{R}^{d_{\text{in}}}$ and $\mathbf{v}_i \in \mathbb{R}^{d_{\text{out}}}$ such that $\mathbf{W} \approx \sum_{i=1}^m \mathbf{v}_i \mathbf{k}_i^\top$. For an input hidden state for a token $\mathbf{x} \in \mathbb{R}^{d_{\text{in}}}$, the matrix-vector product effectively performs a content-addressable read operation:
\begin{equation}
\label{eq:associative_read}
\mathbf{y} = \mathbf{W}\mathbf{x} \approx \sum\nolimits_{i=1}^m \mathbf{v}_{i} (\mathbf{k}_{i}^\top \mathbf{x}),
\end{equation}
where the inner product $(\mathbf{k}_{i}^\top \mathbf{x})$ computes the \textit{relevance} (or activation strength) of the $i$-th memory slot to the input, which weights the retrieval of the value vector $\mathbf{v}_{i}$.
\end{definition}

\begin{remark}[Key-Value Memory of Weight Matrix]
It is crucial to distinguish Definition \ref{def:assoc_mem} from the key-value mechanism in Self-Attention \citep{vaswani2017attention}. In attention, keys and values are \textit{dynamic} projections of the input sequence generated at runtime. In contrast, under the linear associative memory view, the keys $\mathbf{k}_i$ and values $\mathbf{v}_i$ are \textit{static} parameters intrinsic to the weight matrix, representing knowledge patterns acquired during pre-training.
\end{remark}

\subsection{Proposed \ours}
\label{sec:mix_r}

\subsubsection{Mixture of Rank-1 Memory Experts}

\noindent\textbf{Fine-grained rank-1 memory augmentation.} 
Grounded in the associative memory view (Definition \ref{def:assoc_mem}), we reconceptualize the fine-tuning process. Instead of formulating the update $\Delta \mathbf{W}$ as a single-unit rank-$r$ matrix, we treat it as a collection of retrievable rank-1 memory augmentations. Formally, we define the update as the aggregation of $r$ atomic rank-1 key-value pair updates:
\begin{equation}
\label{eq:lora_keyvalue}
    \Delta\mathbf{W}\mathbf{x} = \sum\nolimits_{i=1}^{r} \underbrace{\mathbf{B}_{:,i}}_{\text{Value } \mathbf{v}_i} (\underbrace{\mathbf{A}_{i,:}}_{\text{Key } \mathbf{k}_i^\top} \mathbf{x}),
\end{equation}
where the row vector $\mathbf{A}_{i,:}$ acts as the \textbf{Key} determining the relevance of the atom to the input, and the column vector $\mathbf{B}_{:,i}$ acts as the \textbf{Value} storing the retrieved knowledge.
Crucially, this shifts the paradigm from matrix adaptation to \textit{memory expansion}: the update is no longer a rigid block, but a flexible set of fine-grained atomic memories.

Despite this granular potential, standard LoRA and its variants \citep{lora,pissa,milora} apply updates densely: every rank contributes to every input. Even in MoE-LoRA, while experts are separated at rank-$r$ adapter level, the constituent rank-1 memories \textit{within} each expert remain entangled:
(1) \emph{Interference}: Irrelevant knowledge subspaces within a chosen expert are forced to activate for mismatched inputs.
(2) \emph{Routing collapse}: It overlooks the intrinsic capacity of the key vectors $\mathbf{A}$ to act as \emph{content retrieval keys}, instead relying on indiscriminate dense activation or redundant external routers.

\noindent\textbf{MoRAM formulation.}
To address this, we propose \textbf{\ours} (\textbf{M}ixture \textbf{o}f \textbf{R}ank-1 \textbf{A}ssociative \textbf{M}emory). We dismantle the rigid adapter structure, redefining the model's adaptation parameters as a dynamic collection of atomic memory experts $\mathcal{M}_t = \{(\mathbf{B}_{:,i}, \mathbf{A}_{i,:})\}_{i=1}^{r_t}$, where $r_t$ denotes the total accumulated rank-1 pairs available at task $t$.
For an input $\mathbf{x}$ during task $t$ (with accumulated memory atoms $r_t$), the effective update is defined as a sparse, input-dependent mixture:
\begin{equation}
\label{eq:dyra-mix}
\Delta\mathbf{W}^t = \sum\nolimits_{i=1}^{r_t} \rw_i \mathbf{B}_{:,i} \mathbf{A}_{i,:},
\end{equation}
where $\rw_i \in \mathbb{R}$ represents the computed retrieval confidence of the $i$-th memory atom on each input token hidden state $\rvx$.
This formulation allows the model to freeze specific ``atomic'' memories (preserving old tasks) while inserting new ones on demand of input, or to jointly activate a combination of old and new memory slots to handle shared concepts.

\subsubsection{Self-Activation for MoRAM Routing}
\label{sec:self_act}

Relying on the key-value form of the rank-1 adapter, we propose a self-activation mechanism for routing the memory experts. We derive the mixing weights directly from the intrinsic activation of each atomic memory, leveraging the key vectors $\mathbf{A}_{i,:}$ for memory retrieval. Unlike standard MoE approaches that rely on additional routing networks, our self-activation routing performs routing as a content-addressable retrieval, reducing the possibility of forgetting caused by an additional router.

\noindent\textbf{Self-activated relevance scoring.}
By eliminating the external router, \ours{} avoids the optimization instability and forgetting associated with auxiliary networks. Instead, we derive mixing weights directly from the intrinsic alignment between the input and the memory keys.
Formally, for an input token $\mathbf{x} \in \mathbb{R}^{d_{\text{in}}}$ and the set of accumulated memory atoms up to task $t$, we compute the raw relevance score $\rs_i$ for the $i$-th atom as:
\begin{equation}
\label{eq:raw_score}
    \rs_i \;=\; \frac{\mathbf{A}_{i,:} \mathbf{x}}{\sqrt{\sum_{j=1}^{r_t} (\mathbf{A}_{j,:} \mathbf{x})^2}},
\end{equation}
where the numerator represents the relevance to the memory key $\mathbf{A}_{i,:}$, the denominator performs an $\ell_2$-normalization across the entire memory ensemble for numerical stability.
Empirically, we find that this intrinsic scoring mechanism matches or exceeds the performance of external routers (see Table \ref{tab:mix_strategy}), confirming that the memory keys alone contain sufficient information to determine their own utility.

\subsubsection{Sparse Expert Routing and Mixture}
While \Eqref{eq:raw_score} measures relevance, naive dense activation leads to low specialty and induces interference and computational overhead. We therefore employ sparse routing to enhance the specialization of the fine-grained experts.

\noindent\textbf{Sparse memory expert selection.}
To prevent interference between tasks and ensure a budgeted computational cost, we enforce a sparsity constraint via \texttt{top-k} masking. Given the raw relevance scores $\rs$, we retain only the $k$ entries with the highest activation and mask the rest:
\begin{equation}
    [\mathrm{TopK}(\rs, k)]_i =
\begin{cases}
\rs_i, & \text{if } \rs_i \in \text{top-}k(\rs),\\
-\infty, & \text{otherwise.}
\end{cases}
\end{equation}

This masking ensures that for any given input, at most $k$ out of the total $r_t$ accumulated memory atoms are eligible for activation. This encourages specialization: only a small set of the most relevant atoms are emphasized and trained to capture each kind of input-specific dynamics, and it prevents tiny, noisy activations from spuriously triggering unrelated memory atoms (\eg, those frozen from prior tasks).

\noindent\textbf{Sharpness enhancement.}
To further encourage specialization and concentrate the update on the most relevant atoms, we apply temperature-scaled softmax normalization to enhance the sharpness of the mixture weights:
\begin{equation}
\label{eq:softmax_temp}
    \rw_i \;=\; \text{softmax}\Bigl(\frac{\mathrm{TopK}(\rs, k)}{\tau_{\text{\ours}}}\Bigr)_i,
\end{equation}
where $\tau_{\text{\ours}}$ is a scalar temperature hyperparameter. In the forward pass, a lower $\tau_{\text{\ours}}$ acts as a contrast enhancer, concentrating probability mass on the definitive specialist atoms. In the backward pass, it functions as a gradient router: by sharpening the distribution, stronger learning signals are directed exclusively to the winning atoms. This effectively isolates them from irrelevant updates and accelerates specialization, ensuring that memory atoms only evolve when they are truly relevant to the current data distribution.

\begin{table*}[!t]
\centering
\caption{Comparisons on X-TAIL for each domain in terms of ``Transfer'', ``Average'', and ``Last'' scores (\%). The \highred{best} and the \highblue{second best} results are highlighted in \highred{red} and \highblue{blue}, respectively. 
   }
    % \vspace{-0.2cm}
   \label{tab:few_xtail}
    \resizebox{\linewidth}{!}{
	\begin{tabular}{l>{\centering\arraybackslash}p{1cm} >{\centering\arraybackslash}p{1cm}>{\centering\arraybackslash}p{1cm} >{\centering\arraybackslash}p{1cm} >{\centering\arraybackslash}p{1cm} >{\centering\arraybackslash}p{1cm} >{\centering\arraybackslash}p{1cm} >{\centering\arraybackslash}p{1cm} >{\centering\arraybackslash}p{1cm} >{\centering\arraybackslash}p{1cm} >{\centering\arraybackslash}p{1.6cm}}
 
		\toprule
               {\quad} \makecell[c]{Method} & \makecell[c]{\rotatebox{90}{Aircraft}} & \makecell[c]{\rotatebox{90}{Caltech}}  & \makecell[c]{\rotatebox{90}{DTD}} & \makecell[c]{\rotatebox{90}{EuroSAT}} & \makecell[c]{\rotatebox{90}{Flowers}} & \makecell[c]{\rotatebox{90}{Food}} & \makecell[c]{\rotatebox{90}{MNIST}} & \makecell[c]{\rotatebox{90}{OxPet}} & \makecell[c]{\rotatebox{90}{Cars}} & \makecell[c]{\rotatebox{90}{SUN397}} & \makecell[c]{\textit{Average}} \\
  
		\midrule
            \rowcolor{RowCOLOR}\multicolumn{12}{l}{\emph{CLIP}}\\
        {\quad}Zero-shot & 23.5 & 76.8 & 37.3 & 36.7 & 63.6 & 84.0 & 46.7 & 86.7 & 66.1 & 63.7 & {\cellcolor{CellCOLOR!50}}58.5\\
         
        {\quad}Fine-tune & 39.6	&84.7 &70.0	&94.7 &97.0	&85.8 &97.6	&93.4 &81.0	&74.7 & {\cellcolor{CellCOLOR!50}}81.9\\
   
            \midrule
            \rowcolor{RowCOLOR}\multicolumn{12}{l}{\emph{Transfer}}\\
            {\quad}Zero-shot \citep{clip} & -- & \highred{76.8}&\highblue{37.3}& {36.7} &{63.6}& 84.0 &\highred{46.7}&86.7&\highred{66.1}& \highred{63.7} & {\cellcolor{CellCOLOR!50}}{62.4}\\
            {\quad}LwF \citep{li2017learning}& -- & 66.6 & 26.9 & 19.5 & 51.0 & 78.4 & 26.6 & 68.9 & 35.5 & 56.1& {\cellcolor{CellCOLOR!50}}47.7\\
            {\quad}WiSE-FT \citep{wortsman2022robust} & -- & 70.1 & 31.9 & 25.3 & 56.3 & 79.8 & 29.9 & 74.9 & 45.6 & 56.8 & {\cellcolor{CellCOLOR!50}}52.3\\
            {\quad}iCaRL \citep{rebuffi2017icarl}& -- & 71.7 & 35.0 & {43.0} & 63.4 & \highred{86.9} & {43.9} & {87.8} & 63.7 & 60.0 & {\cellcolor{CellCOLOR!50}}61.7\\
            {\quad}ZSCL \citep{zscl} & -- & 73.3 & 32.6 & 36.8 & 62.1 & 83.8 & 42.1 & 83.6 & 56.5 & 60.2 & {\cellcolor{CellCOLOR!50}}59.0\\
            {\quad}MoE-Adapter \citep{boosting} & -- & 71.0 & 34.9 & 19.2 & 63.0 & \highblue{86.6} & 20.0 & 87.2 & 63.7 & 58.6 & {\cellcolor{CellCOLOR!50}}56.0\\
            {\quad}RAIL-Primal \citep{rail}& -- & \highred{76.8}&\highblue{37.3}& {36.7} &{63.6}& 84.0 &\highred{46.7}&86.7&\highred{66.1}& \highred{63.7} & {\cellcolor{CellCOLOR!50}}{62.4}\\
            {\quad}{CoDyRA \citep{codyra}} & -- & {74.3} & {36.8} & \highblue{44.2} &\highred{69.9}& 83.5 &42.8& \highred{88.9}& {64.6} & \highblue{63.4} & {\cellcolor{CellCOLOR!50}}\highblue{63.2}\\
           \midrule
            \rowcolor[RGB]{253, 253, 255} {\quad}{\ours} & -- & \highblue{74.5} & \highred{38.1} & \highred{46.9} & \highblue{65.3} & 82.9 & \highblue{45.8} & \highblue{88.2} & \highblue{65.1} & 62.9 & {\cellcolor{CellCOLOR!50}}\highred{63.3} \\
           
           \midrule
            \rowcolor{RowCOLOR}\multicolumn{12}{l}{\emph{Average}}\\
            {\quad}LwF \citep{li2017learning}& 24.7 & 79.7 & 38.3 & 36.9 & 63.9 & 81.0 & 36.5 & 71.9 & 42.7 & 56.7& {\cellcolor{CellCOLOR!50}}53.2\\
            {\quad}WiSE-FT \citep{wortsman2022robust} & 27.1 & 76.5 & 40.9 & 31.3 & 68.7 & 81.6 & 31.4 & 74.7 & 51.7 & 58.4 & {\cellcolor{CellCOLOR!50}}54.2\\
            {\quad}iCaRL \citep{rebuffi2017icarl}& 25.4 & 72.1 & 37.5 & 51.6 & 65.1 & \highred{87.1} & 59.1 & 88.0 & 63.7 & 60.1 &{\cellcolor{CellCOLOR!50}} 61.0\\
            {\quad}ZSCL \citep{zscl} & 36.0 & 75.0 & 40.7 & 40.5 & 71.0 & 85.3 & 46.3 & 83.3 & 60.7 & 61.5 & {\cellcolor{CellCOLOR!50}}60.0\\
            {\quad}MoE-Adapter \citep{boosting}& \highblue{43.6} & 77.9 & 52.1 & 34.7 & 75.9 & \highblue{86.3} & 45.2 & 87.4 & 66.6 & 60.2 & {\cellcolor{CellCOLOR!50}}63.0\\
            {\quad}RAIL-Primal \citep{rail} & 42.4 & \highred{89.8} & 55.7 & 68.5 & \highred{84.0} & 83.3 & \highblue{65.3} & 85.8 & \highblue{67.9} & \highred{64.5} & {\cellcolor{CellCOLOR!50}}70.7\\
            % \midrule
             {\quad}{CoDyRA \citep{codyra}} & 41.4 & 81.0 & \highblue{58.7} & \highblue{77.8} &{83.4} & 84.6 &{64.5}& \highred{90.4}&67.2&\highblue{64.4} & {\cellcolor{CellCOLOR!50}}\highblue{71.3}\\
              
              \midrule
              \rowcolor[RGB]{253, 253, 255} {\quad}{\ours} & \highred{44.1} & \highblue{81.6} & \highred{64.6} & \highred{79.6} & \highblue{83.9} & 84.4 & \highred{66.5} & \highblue{89.7} & \highred{68.4} & 64.1 &{\cellcolor{CellCOLOR!50}}\highred{72.7} \\

             \midrule
            \rowcolor{RowCOLOR}\multicolumn{12}{l}{\emph{Last}}\\
            {\quad}LwF \citep{li2017learning}& 25.5 & 72.1 & 38.9 & 55.4 & 65.5 & \highblue{87.3} & 81.9 & 88.6 & 63.6 & 61.5& {\cellcolor{CellCOLOR!50}}64.0\\
            {\quad}WiSE-FT \citep{wortsman2022robust} & 21.8 & 76.8 & 42.9 & 20.8 & 77.5 & 84.9 & 30.7 & 76.6 & 75.8 & 72.5 & {\cellcolor{CellCOLOR!50}}58.0\\
            {\quad}iCaRL \citep{rebuffi2017icarl}& 25.5 & 72.1 & 38.9 & 55.4 & 65.5 & \highblue{87.3} & 81.9 & 88.6 & 63.6 & 61.5 & {\cellcolor{CellCOLOR!50}}64.0\\
            {\quad}ZSCL \citep{zscl} & 33.1 & 75.3 & 43.5 & 35.2 & 74.6 & \highred{87.4} & 50.4 & 84.2 & 77.3 & 73.4 & {\cellcolor{CellCOLOR!50}}63.4\\
            {\quad}MoE-Adapter  \citep{boosting} & \highred{43.2} & 78.7 & 57.6 & 32.8 & 79.4 & 86.0 & 86.7 & 87.8 & \highblue{78.2} & \highblue{74.2} & {\cellcolor{CellCOLOR!50}}70.5\\
            {\quad}RAIL-Primal \citep{rail} & \highblue{41.7} & \highred{94.0} & \highblue{66.0} & 86.4 & \highred{97.2} & 82.4 & {93.1} & 83.6 & 75.0 & 71.3 & {\cellcolor{CellCOLOR!50}}{79.1}\\
            
            {\quad}{CoDyRA \citep{codyra}} & 37.7& \highblue{81.5}&	65.1&	\highblue{89.9}&	91.4&	85.5&	\highblue{96.8}&	\highred{93.3}&	77.3 &	{73.5} & {\cellcolor{CellCOLOR!50}}\highblue{79.2} \\
            \midrule
            \rowcolor[RGB]{253, 253, 255} {\quad}{\ours} & 37.7 & \highblue{81.5} & \highred{70.7} & \highred{92.4} & \highblue{95.0} & 86.0 & \highred{97.6} & \highblue{92.6} & \highred{81.0} & \highred{74.7} & {\cellcolor{CellCOLOR!50}}\highred{80.9} \\

             \bottomrule
	\end{tabular}}
    % \vspace{-0.2cm}
\end{table*}

\noindent\textbf{Threshold-based expert selection.}
To maximize retrieval precision during inference, we apply an additional threshold-based expert selection, via a relevance threshold $\delta$ to the normalized scores, enabling adaptively pruning memory atoms with weak activation signals. 
Since the  number of useful experts may differ depending on layers and input, at inference time, we apply  this threshold-based filter to further adaptively select experts from the top-k experts, while reducing both computational overhead and noise:
\begin{equation}
     \rw_i := \mathbbm{1}\{\rs_i \ge \delta\} \odot \rw_i.
\end{equation}
This operation yields a highly sparse, input-dependent set comprising only the most significant memory experts.

\subsubsection{Continual Learning with MoRAM}
\noindent\textbf{Incremental memory expansion.} 
Unlike methods that add multi-rank adapter blocks as a whole unit, \ours expands memory ``little by little.'' For each new task, we introduce a set of $r$ new atomic rank-1 pairs (Keys $\mathbf{A}$ and Values $\mathbf{B}$), freeze all prior atoms, and let the self-activation mechanism jointly route across the union of all old and new memories. This seamlessly integrates new knowledge while preserving the reusability of prior tasks.  
The learning process becomes an incremental accumulation of fine-grained \emph{learning snapshots} as atomic parametric memory units. 
This avoids persistent weight updates and enables fine-grained test-time \emph{recall} and \emph{reuse} of learning snapshots in memory. 
The sparse mixture of rank-1 units enables flexible memory retrieval and reuse.
To enhance efficiency and enable sub-linear expansion rate, \ours allows low-utility atoms to be pruned after training with minimal performance degradation (more in Appendix \ref{supp:compute}).

\noindent\textbf{Training objectives.} 
In our experiments, we only use the model’s standard training objective, without any extra regularization or load‐balancing constraints. 
Benefiting from the fine-grained modeling with rank-1 experts, the experts tend to be specialized and the routing tends to be balanced and stable naturally. 
Even without additional regularization, the specialized expert self-activation reduces router degradation and forgetting, while we believe further regularization might be more helpful. 
In practice, we empirically observe that conventional load-balancing regularization \citep{shazeer2017outrageously,lepikhin2020gshard,fedus2022switch,dai2024deepseekmoe} for MoE may not be necessary for the experimented setup (Appendix \ref{supp:load_balance}). And imposing regularization on the self-activation routing improperly tends to hinder the learning of the key vectors, since the key vectors for routing are also crucial for information representation.

\begin{table*}[!t]
    \centering
        \caption{Comparison with a broad range of CL methods on the TRACE benchmark.
        % using the LLaMA-3.2-1B-Instruct. 
        We report Overall Performance (OP (\%) $\uparrow$) and Backward Transfer (BWT (\%) $\downarrow$). Results are averaged over three runs with standard deviations. The best results are highlighted in bold.}
        % \vspace{-0.2cm}
    \label{tab:llm_cl}
    \resizebox{\linewidth}{!}{
    \small
\begin{tabular}{@{}lccccccccccc@{}}
\toprule
 & FIX(ICL) & SeqLoRA & OGD & GEM & EWC & L2P & DualPrompt & HiDeLoRA & O-LoRA & TreeLoRA & \ours \\ \midrule
\rowcolor{RowCOLOR}\multicolumn{12}{l}{meta-llama /   LLaMA-2-7B-Chat} \\ \midrule
OP & 38.94 $\pm$ 0.3 & 34.3 $\pm$ 1.2 & 42.09 $\pm$ 1.6 & 40.08 $\pm$ 1.6 & 42.36 $\pm$ 1.2 & 36.23 $\pm$ 0.8 & 37.69 $\pm$ 1.2 & 41.60 $\pm$ 0.8 & 42.78 $\pm$ 0.8 & 43.52 $\pm$ 1.0 & \multicolumn{1}{c}{\cellcolor{CellCOLOR!50} \textbf{44.54 $\pm$ 0.9}} \\
BWT & -- & 18.5 $\pm$ 0.8 & 8.06 $\pm$ 1.2 & 6.77 $\pm$ 1.2 & 5.97 $\pm$ 0.8 & 8.25 $\pm$ 0.8 & 8.03 $\pm$ 0.8 & 7.12 $\pm$ 0.4 & 7.16 $\pm$ 0.4 & 3.46 $\pm$ 0.4 & \multicolumn{1}{c}{\cellcolor{CellCOLOR!50} \textbf{1.37 $\pm$ 0.3}}  \\ \midrule
\rowcolor{RowCOLOR}\multicolumn{12}{l}{google / Gemma-2B-it} \\ \midrule
OP & 32.3 $\pm$ 0.2 & 31.89 $\pm$ 0.8 & 32.85 $\pm$ 1.4 & 26.48 $\pm$ 1.5 & 28.35 $\pm$ 1.6 & 31.14 $\pm$ 1.2 & 32.42 $\pm$ 1.0 & 33.25 $\pm$ 0.9 & 33.73 $\pm$ 0.8 & 33.41 $\pm$ 0.9 & \multicolumn{1}{c}{\cellcolor{CellCOLOR!50} \textbf{36.27 $\pm$ 0.7}}  \\
BWT & -- & 15.28 $\pm$ 0.4 & 12.27 $\pm$ 0.9 & 18.25 $\pm$ 0.9 & 16.96 $\pm$ 1.2 & 15.77 $\pm$ 0.7 & 14.25 $\pm$ 0.5 & 13.66 $\pm$ 0.5 & 12.36 $\pm$ 0.4 & 8.50 $\pm$ 0.5 & \multicolumn{1}{c}{\cellcolor{CellCOLOR!50} \textbf{2.74 $\pm$ 0.4}} \\ \midrule
\rowcolor{RowCOLOR}\multicolumn{12}{l}{meta-llama /   LLaMA-3-1B-Instruct} \\ \midrule
OP & 31.16 $\pm$ 0.4 & 29.73 $\pm$ 1.6 & 30.12 $\pm$ 2.0 & 32.19 $\pm$ 2.0 & 31.96 $\pm$ 1.6 & 29.38 $\pm$ 1.2 & 30.76 $\pm$ 1.2 & 33.73 $\pm$ 1.2 & 32.94 $\pm$ 0.8 & 36.14 $\pm$ 0.7 & \multicolumn{1}{c}{\cellcolor{CellCOLOR!50}\textbf{37.77 $\pm$ 0.8}} \\
BWT & -- & 17.03 $\pm$ 1.2 & 15.2 $\pm$ 1.6 & 10.74 $\pm$ 1.6 & 11.62 $\pm$ 1.2 & 13.57 $\pm$ 0.8 & 11.34 $\pm$ 0.8 & 12.36 $\pm$ 0.8 & 12.89 $\pm$ 1.2 & 7.36 $\pm$ 0.8 & \multicolumn{1}{c}{\cellcolor{CellCOLOR!50} \textbf{3.12 $\pm$ 0.8} }   \\ \bottomrule
\end{tabular}
    }
    % \vspace{-0.2cm}
\end{table*}

\begin{table*}[t]
\centering
\small
\caption{Standard fine-tuning of Llama-3.1-8B on CodeAlpaca. We report zero-shot in-domain performance on HumanEval (Pass@1) for the code generation and out-of-domain accuracy on selected MMLU subjects (formal logic, philosophy, world religions, economics, public relations, STEM, physics, machine learning). The last two columns report trainable parameters (for \ours: added / activated).}
\label{tab:llm_ood_main}
\resizebox{\linewidth}{!}{%
\begin{tabular}{@{}l
  c
  *{9}{c}
  c
  c@{}}
\toprule
\multirow{2}{*}{Method} & \multirow{2}{*}{\begin{tabular}{c}HumanEval\\(Pass@1)\end{tabular}} 
  & \multicolumn{9}{c}{Out-of-Domain (Acc.)} 
  & \multirow{2}{*}{\begin{tabular}{c}Params\\(M)\end{tabular}} 
  & \multirow{2}{*}{\(\%\)Params} \\
\cmidrule(lr){3-11}
  & 
  & Logic & Phil. & Reli.
  & Econ. & Pub. Rel. & STEM 
  & Phys. & ML & MMLU 
  &  &  \\
\midrule
\rowcolor{CellCOLOR}
Llama-3.1-8B
  & 38.40  
  & 42.06       & \highred{71.06} & \highred{83.63}   
  & 70.17 & \highred{68.18}  & 54.84      
  & 39.22   & \highblue{40.18} & 63.45    
  & —     & —        \\
\midrule
LoRA (\(r=4\))  
  & 41.46  
  & 39.68       & 70.09           & 81.87   
  & 71.43  &65.45  & 54.77      
  & \highblue{45.10}   & 40.17       & 63.28    
  & 10.5  & 0.13\%   \\
LoRA (\(r=8\))  
  & 44.51  
  & 39.68       & \highblue{70.74}& 81.87   
  & 71.85 & 64.54  & 54.17      
  & 42.16   & 39.29       & 63.03    
  & 21.0  & 0.26\%   \\
LoRA (\(r=16\)) 
  & \highblue{45.73}
  & 41.27       & 68.49           & 80.70   
  & \highblue{72.69} & \highblue{66.36} & 54.96
  & 44.11   & 38.39       & 63.35    
  & 41.9  & 0.52\%   \\
LoRA (\(r=32\)) 
  & \highred{47.56}
  & \highblue{42.85} & 69.45           & 81.87   
  & 72.27  & \highblue{66.36} & \highblue{55.44}
  & \highblue{45.10} & 39.29       & \highblue{63.59}    
  & 83.9  & 1.03\%   \\
\midrule
\rowcolor{RowCOLOR}
\ours 
  & \highred{47.56}
  & \highred{48.41} & 70.09           & \highblue{82.46} 
  & \highred{73.53} & \highred{68.18} & \highred{55.53}
  & \highred{46.08}   & \highred{41.96} & \highred{63.70}
  & 41.9/\textbf{26.2} & 0.52\%/\textbf{0.32\%} \\
\bottomrule
\end{tabular}%
}
\end{table*}

\section{Experiments}
\label{sec:experiments}

In this paper, we conduct experiments across a diverse set of tasks, including continual learning for both vision-language CLIP models and LMs, and analyze catastrophic forgetting during fine‐tuning. 
Detailed implementation settings and more experiment results are provided in Appendix \ref{supp:settings}.

\subsection{Experimental Results}
\noindent\textbf{Continual Learning of CLIP.}
We evaluate X-TAIL performance in Table~\ref{tab:few_xtail}, reporting \textit{Transfer}, \textit{Last}, and \textit{Average} accuracy (MTIL results in Appendix Table \ref{tab:few_mtil_full}). While prior methods \citep{boosting,rail} are constrained by the base model's zero-shot ceiling, \ours follows \citep{zscl,codyra} to continuously adapt, surpassing this limit and achieving superior Last and Average scores. Critically, \ours enhances representation without the external domain-IDs or feature banks required by \citep{boosting,rail}, and improves upon the fixed weighting in \citep{codyra} by utilizing dynamic, input-dependent gating for finer expert specialization.

\noindent\textbf{Continual Learning of LLMs.}
In Table \ref{tab:llm_cl}, we evaluate \ours on LLaMA \citep{touvron2023llama,grattafiori2024llama} and Gemma \citep{team2024gemma} using the TRACE benchmark \citep{wang2023trace}. Traditional regularization \citep{kirkpatrick2017overcoming}, rehearsal \citep{lopez2017gradient}, and prompt-based \citep{l2p} methods struggle to scale, often underperforming simple In-Context Learning \citep[ICL;][]{brown2020language}. Among LoRA variants, naive SeqLoRA suffers significant forgetting, while recent approaches like O-LoRA \citep{wang2023orthogonal}, HiDeLoRA \citep{wang2025hide}, and TreeLoRA \citep{qian2025treelora} rely on rigid orthogonality or complex structural expansions. In contrast, \ours performs robustly on LLMs, achieving results competitive with existing methods. We provide additional evaluations on language classification tasks in Tables \ref{tab:llm_cl_main} and \ref{tab:supp_llm_cl_long} of Appendix.

\label{sec:llm_cl}

\noindent\textbf{Forgetting and generalization after standard fine‐tuning.}  
Table \ref{tab:llm_ood_main} examines how standard fine-tuning affects both in-domain performance and out-of-domain generalization. \textit{(1) Forgetting:} Dense LoRA updates overwrite pre-trained knowledge, degrading performance on semantically distant topics (\eg, Religions). \textit{(2) Positive transfer:} Conversely, code adaptation improves logically related tasks (\eg, STEM, Logic), indicating valid transfer potential.

In contrast, \ours{} leverages its atomic memory structure, treating updates as independent rank-1 memory atoms rather than entangled blocks, the model isolates new knowledge accumulation from existing representations. This granular independence prevents the catastrophic overwriting of unrelated concepts (\eg, Religions) while the content-addressable self-activation ensures that relevant atoms are reused for logical tasks (\eg, STEM, Logic). Consequently, \ours{} achieves superior out-of-domain accuracy using roughly one-third of the active parameters required by a standard rank-32 LoRA.

\begin{figure*}[t]
  \centering
  % first subfigure
  \vspace{-0.2cm}
  \begin{subfigure}[c]{\textwidth}
    \centering
    \begin{overpic}[width=0.9\linewidth]{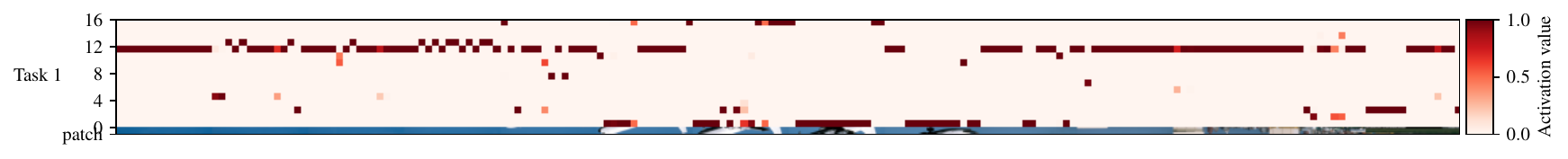}
    \put(38,1){\color{orange}\framebox(3,8){}}
    \put(43,1){\color{orange}\framebox(7,8){}}
    \put(51,1){\color{orange}\framebox(5,8){}}
    \put(59,1){\color{orange}\framebox(4,8){}}
    \end{overpic}%
    \raisebox{0.1cm}{%
        \includegraphics[width=0.07\linewidth]{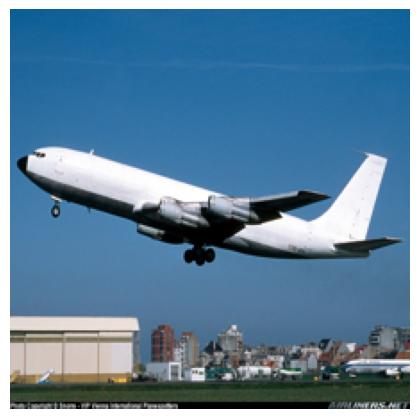}%
    }%
    \\
    % \vspace{-0.2cm}
    \caption{Memory atom activations of \ours on data from Task 1 after learning Task 1.}
    \label{fig:acti_1}
  \end{subfigure}
  % second subfigure
  \begin{subfigure}[c]{\textwidth}
    \centering
    \begin{overpic}[width=0.9\linewidth]{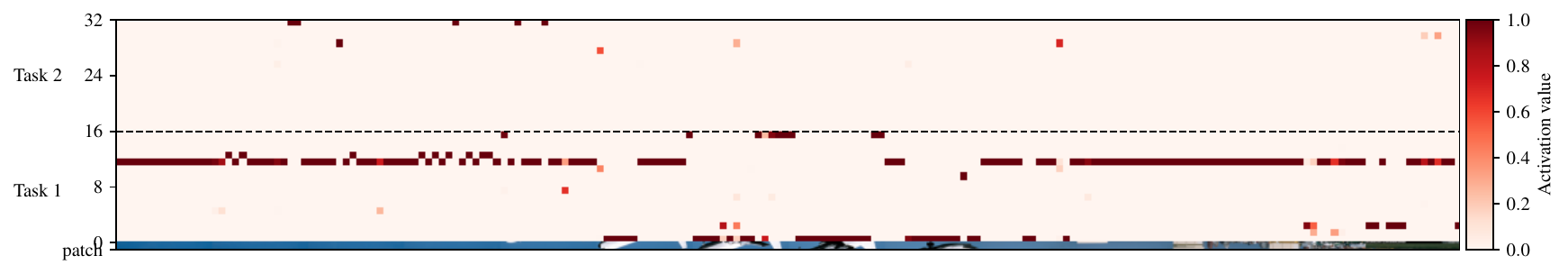}
    \put(38,1){\color{orange}\framebox(3,15.4){}}
    \put(43,1){\color{orange}\framebox(7,15.4){}}
    \put(51,1){\color{orange}\framebox(5,15.4){}}
    \put(59,1){\color{orange}\framebox(4,15.4){}}
    \end{overpic}%
    \raisebox{0.6cm}{%
        \includegraphics[width=0.07\linewidth]{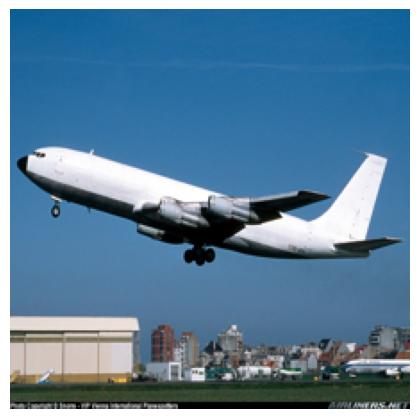}%
    }%
    % \\ \vspace{-0.2cm}
    \caption{Memory atom activations of \ours on data from Task 1 after learning Task 2.}
    \label{fig:acti_2}
  \end{subfigure}
  % third subfigure
  \begin{subfigure}[c]{\textwidth}
    \centering
    \begin{overpic}[width=0.9\linewidth]{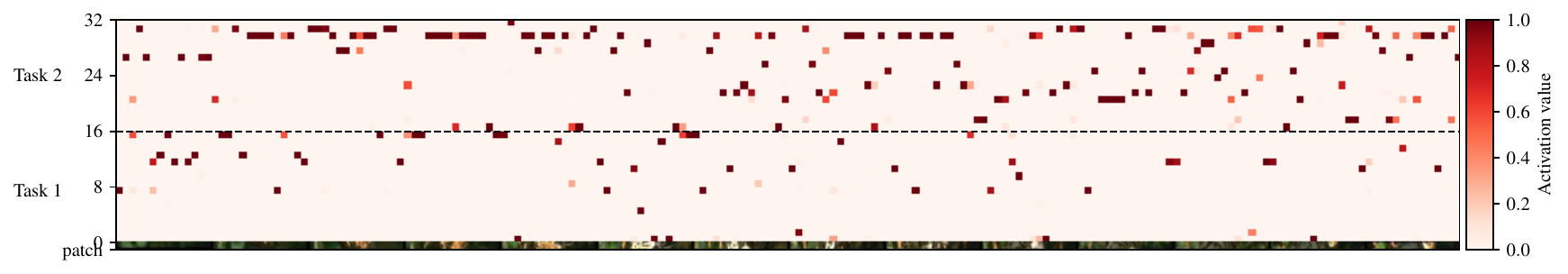}
    \put(21.5,1){\color{orange}\framebox(3,15.4){}}
    \put(27,1){\color{orange}\framebox(4,15.4){}}
    \put(33,1){\color{orange}\framebox(3,15.4){}}
    \put(40,1){\color{orange}\framebox(3,15.4){}}
    \put(46,1){\color{orange}\framebox(3,15.4){}}
    \put(52,1){\color{orange}\framebox(3,15.4){}}
    \put(57,1){\color{orange}\framebox(4, 15.4){}}
    \put(63,1){\color{orange}\framebox(4, 15.4){}}
    \put(70,1){\color{orange}\framebox(3, 15.4){}}
    \end{overpic}%
    \raisebox{0.6cm}{%
        \includegraphics[width=0.07\linewidth]{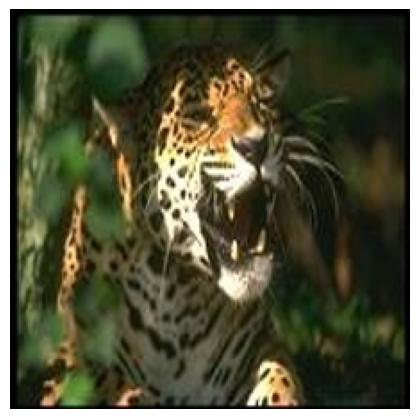}%
    }%
    % \\ 
    % \vspace{-0.2cm}
    \caption{Memory atom activations of \ours on data from Task 2 after learning Task 2.}
    \label{fig:acti_3}
  \end{subfigure}
  % \vspace{-0.25cm}
    \caption{Visualization of \ours's memory atom activations during Task 1 and Task 2 training. Activations are extracted from the K projection in the attention module (layer 8) of the image encoder. Corresponding image patches are shown below each activation map, with regions relevant to each class marked by orange bounding boxes. Zoom in for details. More visualizations are in Figs.~\ref{fig:supp_rank_visual} and \ref{fig:supp_rank_visual2} of the Appendix, demonstrating forgetting mitigation and knowledge reuse.}
  \label{fig:main_rank_visual}
  % \vspace{-0.2cm}
\end{figure*}
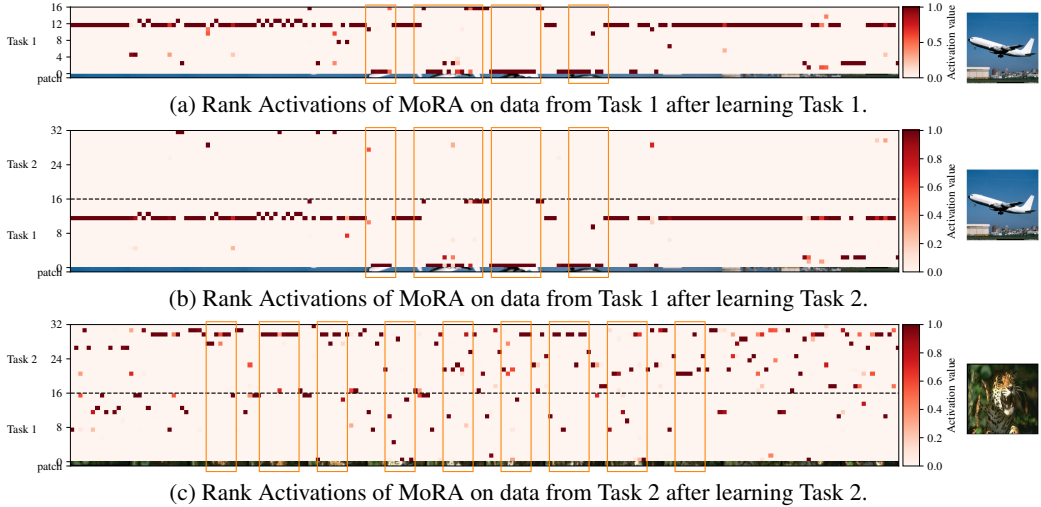

\begin{figure*}[h]
    \begin{minipage}{0.43\textwidth}
    \centering
        \captionof{table}{Memory retrieval (routing) strategies.}
        \label{tab:mix_strategy}
        \resizebox{\linewidth}{!}{
\begin{tabular}{lccc}
    \toprule
    \textbf{Routing} & \textbf{Transfer} & \textbf{Average} & \textbf{Last} \\
    \midrule
    \rowcolor{RowCOLOR}\multicolumn{4}{l}{\textit{Coarse-Grained (Rank-$r$ Experts)}} \\
    MoE-LoRA (Baseline) & 62.56 & 69.45 & 74.53 \\
    \hspace{2mm} \textit{w/} Temperature Scaling ($\tau^*$) & 62.48 & 69.40 & 74.57 \\
    \midrule
    \rowcolor{RowCOLOR}\multicolumn{4}{l}{\textit{Fine-Grained (Rank-1 Memory Experts)}} \\
    External Router (Learned $\mathbf W_r$) & 60.09 & 65.97 & 69.76 \\
    Self-Activated Retrieval & 60.26 & 65.94 & 69.85 \\
    \hspace{2mm} \textit{w/} Sparsity Constraint (Top-$k$) & 60.69 & 66.52 & 70.62 \\
    \hspace{2mm} \textit{w/} Temperature Scaling ($\tau_\text{\ours}$) & 62.07 & 71.15 & 79.62 \\
    \hspace{2mm} \textit{w/} Threshold-based Selection ($\delta$) & 60.78 & 66.83 & 71.08 \\
    \hspace{2mm} \textbf{\ours{} (Full)} & \textbf{63.30} & \textbf{72.70} & \textbf{80.90} \\
    \bottomrule
\end{tabular}
        
        }
        % \vspace{-0.2cm}
    \end{minipage} 
    \hfill
    \phantomcaption
    \vspace{-0.2cm}
    \begin{minipage}{0.54\textwidth}
      \centering
      \begin{subfigure}[t]{0.33\linewidth}
        \centering
        \includegraphics[width=\linewidth]{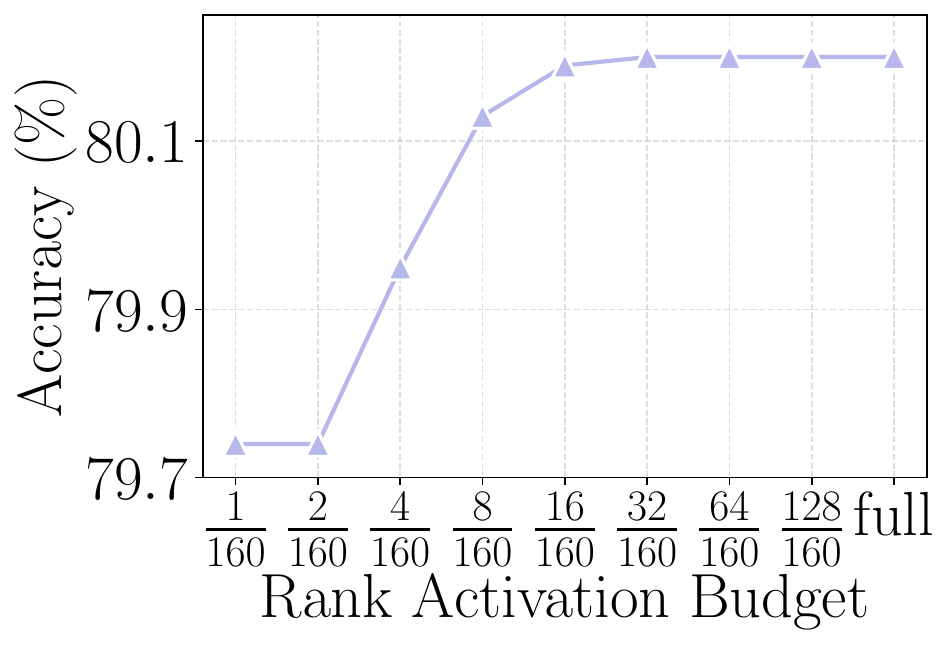}
        \vspace{-0.6cm}
        \caption{Acti. Budget}
        \label{fig:xtail-topk}
      \end{subfigure}\hfill
      \begin{subfigure}[t]{0.33\linewidth}
        \centering
        \includegraphics[width=\linewidth]{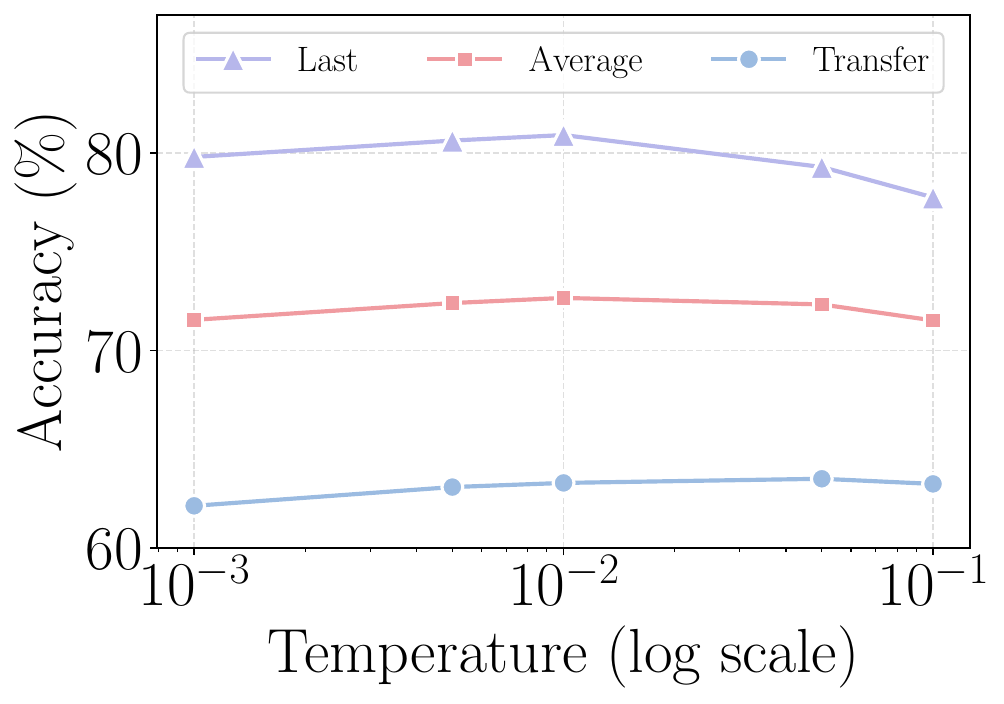}
        \vspace{-0.6cm}
        \caption{Temp. $\tau_\text{\ours}$}
        \label{fig:xtail-temp}
      \end{subfigure}\hfill
      \begin{subfigure}[t]{0.33\linewidth}
        \centering
        \includegraphics[width=\linewidth]{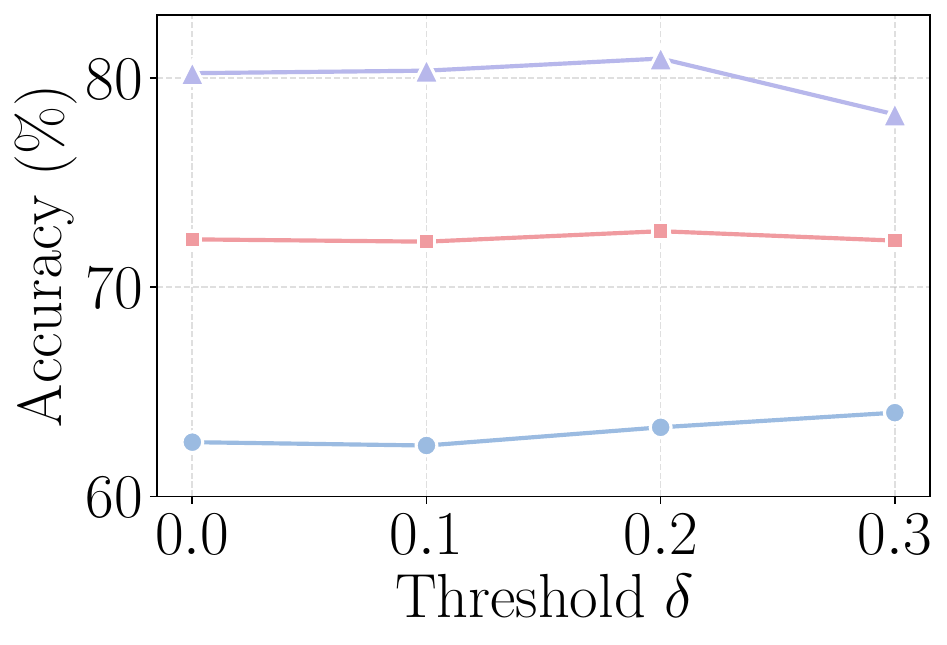}
        \vspace{-0.6cm}
        \caption{Thre. $\delta$}
        \label{fig:xtail-thre}
      \end{subfigure}
      % \vspace{-0.1cm}
      \addtocounter{figure}{-1}
      \caption{Ablation on (a) rank activation budget, (b) temperature $\tau_\text{\ours}$, and (c) threshold $\delta$.}
      \label{fig:xtail-ablation}
    \end{minipage}
\end{figure*}

\subsection{Visualizations of Rank-1 Memory Activations}
\label{sec:rank_vis}
Figure~\ref{fig:main_rank_visual} shows memory atom activations recorded during continual learning. These visualizations illustrate two properties of \ours: (1) individual ranks specialize on distinct input patterns, and (2) the model substantially reduces cross-task interference, thereby mitigating forgetting. 
Extended visualizations across more tasks and scenarios appear in Fig.~\ref{fig:supp_rank_visual} and~\ref{fig:supp_rank_visual2} in the Appendix.

\noindent\textbf{Each memory atom specializes in distinct input patterns.}
In Fig.~\ref{fig:acti_1}, airplane patches strongly activate memory atom 0, while blue-sky backgrounds predominantly activate memory atom 11. In Fig.~\ref{fig:acti_3}, memory atoms 19, 20, and 29 (orange boxes) respond to jaguar patches. The more complex backgrounds in Fig.~\ref{fig:acti_3} (\eg, leaves, shadows) yield a richer, more distributed pattern than the simple blue sky in Fig.~\ref{fig:acti_1}, highlighting MoRAM's capacity to model contextual complexity. Some memory atoms learned earlier are also reused on later tasks (Fig.~\ref{fig:supp_rank_visual2}, Appendix), indicating transfer of shared semantics.

\noindent\textbf{Reduced interference and mitigated forgetting.}
Comparing the same input after Task 1 (Fig.~\ref{fig:acti_1}) and after Task 2 (Fig.~\ref{fig:acti_2}) shows almost identical activations (more in Fig.~\ref{fig:supp_rank_visual}, Appendix): memory atom 0 still responds to airplane semantics and memory atom 11 to blue-sky patches. This stability indicates that our self-activated, sparse mixture of rank-1 atoms prevents later updates from overwriting earlier task representations, reducing interference and mitigating forgetting.

\subsection{Ablation Studies}
\noindent\textbf{Ablation of memory retrieval strategies.}
Table \ref{tab:mix_strategy} analyzes the impact of different retrieval mechanisms under a controlled setup.
(1) \textit{MoE-LoRA (Coarse Baseline)} employs a router to weight entire adapters. Its coarse granularity forces the simultaneous activation of conflicting subspaces, leading to significant interference and forgetting. We further apply the same temperature scaling (\Eqref{eq:softmax_temp}) to this baseline and report the best-performing setting. Performance remains flat across eight temperatures ($\tau \in \{0.001, 0.005, 0.01, 0.05, 0.1, 0.3, 0.5, 0.7\}$), confirming that coarse experts bundle heterogeneous knowledge into indivisible blocks where sharpening selection alone cannot improve specialization.
(2) \textit{External Router} applies a separately learned router to individual rank-1 atoms. However, decoupling the routing logic from the memory content leads to \textit{retrieval collapse}: as the number of atoms grows, the external router struggles to index them precisely, resulting in worse performance than the coarse baseline.
(3) \textit{Self-Activated Retrieval} replaces the external router with intrinsic key-value matching (Eq.~\ref{eq:raw_score}). By ensuring the retrieval condition is strictly aligned with the expert's content, it resolves the routing ambiguity without extra parameters.
(4) \textit{Sparsity Constraint (Top-$k$)} enforces a strict activation budget. This serves as a gating mechanism that prevents memory collisions and reduces interference during training.
(5) \textit{Temperature Scaling} sharpens the retrieval distribution. This concentrates gradient flow on the most relevant ``specialist'' atoms, accelerating their adaptation while protecting shared memories.
(6) \textit{Threshold-based selection} prunes weak activation signals ($\delta$) strictly at test time, maximizing retrieval precision by eliminating noise.

\noindent\textbf{Rank Activation Budget.}
We probe atom usage by varying the activation budget on a 10-task checkpoint (Fig.~\ref{fig:xtail-topk}). To isolate this effect, we disable other sparsity mechanisms (\eg, thresholding). Performance improves as the budget increases from 2 to 16 (10\% of total atoms) before plateauing, confirming that a small, sparse subset of specialized atoms suffices to cover diverse inputs. Crucially, accuracy remains robust even at high budgets, demonstrating that self-activation naturally prioritizes relevant signals while suppressing noise, even without strict sparsity constraints.

\noindent\textbf{Sharpness enhancement.} 
Temperature scaling (\Eqref{eq:softmax_temp}) modulates activation sparsity: lower $\tau_\text{\ours}$ concentrates probability mass, while higher $\tau_\text{\ours}$ broadens expert participation. Fig. \ref{fig:xtail-temp} demonstrates that higher $\tau_\text{\ours}$ significantly boosts transfer to unseen tasks (even surpassing SOTA in Table~\ref{tab:few_xtail}). However, we select a moderate $\tau_\text{\ours}$ of 0.01 to optimally balance specificity (retention) and generalization (transfer), ensuring strong overall performance.

\noindent\textbf{Threshold-based rank selection.}
Figure \ref{fig:xtail-thre} shows the effect of the test-time rank selection threshold \(\delta\). Applying a modest threshold removes low-activation ranks and reduces noisy contributions, which improves both downstream adaptation and out-of-domain generalization. 

\section{Conclusion}
We presented \ours, a framework that challenges the coarse-grained design of standard MoE-LoRA. By redefining adaptation as the accumulation of atomic associative memories, we demonstrated that high-performing CL does not require complex external routers, but rather precise, content-addressable retrieval of low-rank atomic memories.
\\
\noindent\textbf{Future works.}
\ours controls mixture sharpness and sparsity via top-k selection, temperature scaling, and thresholding. While effective, the sparse activation mechanism can be further improved by incorporating awareness of the input data distribution. Learning this temperature or adapting the mixture’s sharpness based on data offers a promising avenue for future research. We also aim to extend the method’s applicability to a broader range of PTMs and applications.

\section*{Acknowledgements}
This work was partially supported by the ARC DECRA Fellowship (DE230101591), and the ARC Discovery Project Grant (DP260103379). H. Lu is affiliated with CSIRO Data61 through a PhD scholarship and acknowledges the support of the Google PhD Fellowship.

\section*{Impact Statement}

MoRAM’s sparse mixture of rank-1 atoms method makes continual adaptation of large vision–language and language models far more efficient and reliable. By updating only the most relevant low-rank subspaces, it reduces compute and memory requirements, enabling on-device personalization in domains like education, healthcare, and finance. This democratizes access to powerful, continually evolved models for smaller teams and resource-constrained settings, driving innovation and improving quality of life. Preserving pre-trained knowledge also ensures stability in safety-critical systems, such as medical diagnostics and autonomous vehicles, where forgetting of pre-trained capabilities can be dangerous.

The focus of this paper is fundamental research, and is broadly applicable to model fine-tuning techniques. Thus, it is possible that \ours's lightweight updates could be misused to insert stealthy backdoors or reinforce hidden biases that are hard to detect. Because it leaves most pre-trained parameters untouched, existing biases may persist or become entrenched. To mitigate these risks, we recommend strict audit logging of rank-level updates, anomaly detection on sparse changes, robust access controls, and routine bias and fairness assessments alongside any deployment.

\bibliography{example_paper}

@String(ECCV= {Eur. Conf. Comput. Vis.})

@String(ECCV  = {ECCV})

@inproceedings{clip,
  title={Learning transferable visual models from natural language supervision},
  author={Radford, Alec and Kim, Jong Wook and Hallacy, Chris and Ramesh, Aditya and Goh, Gabriel and Agarwal, Sandhini and Sastry, Girish and Askell, Amanda and Mishkin, Pamela and Clark, Jack and others},
  booktitle={International conference on machine learning},
  pages={8748--8763},
  year={2021},
  organization={PMLR}
}

@article{maji2013fine,
  title={Fine-grained visual classification of aircraft},
  author={Maji, Subhransu and Rahtu, Esa and Kannala, Juho and Blaschko, Matthew and Vedaldi, Andrea},
  journal={arXiv preprint arXiv:1306.5151},
  year={2013}
}

@inproceedings{fei2004learning,
  title={Learning generative visual models from few training examples: An incremental bayesian approach tested on 101 object categories},
  author={Fei-Fei, Li and Fergus, Rob and Perona, Pietro},
  booktitle={2004 conference on computer vision and pattern recognition workshop},
  pages={178--178},
  year={2004},
  organization={IEEE}
}

@inproceedings{cimpoi2014describing,
  title={Describing textures in the wild},
  author={Cimpoi, Mircea and Maji, Subhransu and Kokkinos, Iasonas and Mohamed, Sammy and Vedaldi, Andrea},
  booktitle={Proceedings of the IEEE conference on computer vision and pattern recognition},
  pages={3606--3613},
  year={2014}
}

@article{helber2019eurosat,
  title={Eurosat: A novel dataset and deep learning benchmark for land use and land cover classification},
  author={Helber, Patrick and Bischke, Benjamin and Dengel, Andreas and Borth, Damian},
  journal={IEEE Journal of Selected Topics in Applied Earth Observations and Remote Sensing},
  volume={12},
  number={7},
  pages={2217--2226},
  year={2019},
  publisher={IEEE}
}

@inproceedings{nilsback2008automated,
  title={Automated flower classification over a large number of classes},
  author={Nilsback, Maria-Elena and Zisserman, Andrew},
  booktitle={2008 Sixth Indian conference on computer vision, graphics \& image processing},
  pages={722--729},
  year={2008},
  organization={IEEE}
}

@inproceedings{bossard2014food,
  title={Food-101--mining discriminative components with random forests},
  author={Bossard, Lukas and Guillaumin, Matthieu and Van Gool, Luc},
  booktitle={Proceedings of the European conference on computer vision (ECCV)},
  pages={446--461},
  year={2014}
}

@article{deng2012mnist,
  title={The mnist database of handwritten digit images for machine learning research [best of the web]},
  author={Deng, Li},
  journal={IEEE signal processing magazine},
  volume={29},
  number={6},
  pages={141--142},
  year={2012},
  publisher={IEEE}
}

@inproceedings{parkhi2012cats,
  title={Cats and dogs},
  author={Parkhi, Omkar M and Vedaldi, Andrea and Zisserman, Andrew and Jawahar, CV},
  booktitle={2012 IEEE conference on computer vision and pattern recognition},
  pages={3498--3505},
  year={2012},
  organization={IEEE}
}

@inproceedings{krause20133d,
  title={3d object representations for fine-grained categorization},
  author={Krause, Jonathan and Stark, Michael and Deng, Jia and Fei-Fei, Li},
  booktitle={Proceedings of the IEEE international conference on computer vision workshops},
  pages={554--561},
  year={2013}
}

@inproceedings{xiao2010sun,
  title={Sun database: Large-scale scene recognition from abbey to zoo},
  author={Xiao, Jianxiong and Hays, James and Ehinger, Krista A and Oliva, Aude and Torralba, Antonio},
  booktitle={2010 IEEE computer society conference on computer vision and pattern recognition},
  pages={3485--3492},
  year={2010},
  organization={IEEE}
}

@article{li2017learning,
  title={Learning without forgetting},
  author={Li, Zhizhong and Hoiem, Derek},
  journal={IEEE transactions on pattern analysis and machine intelligence},
  volume={40},
  number={12},
  pages={2935--2947},
  year={2017},
  publisher={IEEE}
}

@inproceedings{rebuffi2017icarl,
	title={icarl: Incremental classifier and representation learning},
	author={Rebuffi, Sylvestre-Alvise and Kolesnikov, Alexander and Sperl, Georg and Lampert, Christoph H},
	booktitle={Proceedings of the IEEE conference on Computer Vision and Pattern Recognition},
	pages={2001--2010},
	year={2017}
}

@article{ding2022don,
  title={Don't Stop Learning: Towards Continual Learning for the CLIP Model},
  author={Ding, Yuxuan and Liu, Lingqiao and Tian, Chunna and Yang, Jingyuan and Ding, Haoxuan},
  journal={arXiv preprint arXiv:2207.09248},
  year={2022}
}

@inproceedings{wortsman2022robust,
  title={Robust fine-tuning of zero-shot models},
  author={Wortsman, Mitchell and Ilharco, Gabriel and Kim, Jong Wook and Li, Mike and Kornblith, Simon and Roelofs, Rebecca and Lopes, Raphael Gontijo and Hajishirzi, Hannaneh and Farhadi, Ali and Namkoong, Hongseok and others},
  booktitle={Proceedings of the IEEE/CVF Conference on Computer Vision and Pattern Recognition},
  pages={7959--7971},
  year={2022}
}

@article{zscl,
  title={Preventing Zero-Shot Transfer Degradation in Continual Learning of Vision-Language Models},
  author={Zheng, Zangwei and Ma, Mingyuan and Wang, Kai and Qin, Ziheng and Yue, Xiangyu and You, Yang},
  journal={arXiv preprint arXiv:2303.06628},
  year={2023}
}

@inproceedings{boosting,
  title={Boosting continual learning of vision-language models via mixture-of-experts adapters},
  author={Yu, Jiazuo and Zhuge, Yunzhi and Zhang, Lu and Hu, Ping and Wang, Dong and Lu, Huchuan and He, You},
  booktitle={Proceedings of the IEEE/CVF Conference on Computer Vision and Pattern Recognition},
  pages={23219--23230},
  year={2024}
}

@article{rail,
  title={Advancing cross-domain discriminability in continual learning of vision-language models},
  author={Xu, Yicheng and Chen, Yuxin and Nie, Jiahao and Wang, Yusong and Zhuang, Huiping and Okumura, Manabu},
  journal={Advances in Neural Information Processing Systems},
  volume={37},
  pages={51552--51576},
  year={2024}
}

@inproceedings{
lora,
title={Lo{RA}: Low-Rank Adaptation of Large Language Models},
author={Edward J Hu and yelong shen and Phillip Wallis and Zeyuan Allen-Zhu and Yuanzhi Li and Shean Wang and Lu Wang and Weizhu Chen},
booktitle={International Conference on Learning Representations},
year={2022},
url={https://openreview.net/forum?id=nZeVKeeFYf9}
}

@article{lora_learn_less,
  title={Lora learns less and forgets less},
  author={Biderman, Dan and Portes, Jacob and Ortiz, Jose Javier Gonzalez and Paul, Mansheej and Greengard, Philip and Jennings, Connor and King, Daniel and Havens, Sam and Chiley, Vitaliy and Frankle, Jonathan and others},
  journal={arXiv preprint arXiv:2405.09673},
  year={2024}
}

@inproceedings{
adalora,
title={Adaptive Budget Allocation for Parameter-Efficient Fine-Tuning },
author={Qingru Zhang and Minshuo Chen and Alexander Bukharin and Pengcheng He and Yu Cheng and Weizhu Chen and Tuo Zhao},
booktitle={The Eleventh International Conference on Learning Representations },
year={2023},
url={https://openreview.net/forum?id=lq62uWRJjiY}
}

@inproceedings{sora,
  title={Sparse low-rank adaptation of pre-trained language models},
  author={Ding, Ning and Lv, Xingtai and Wang, Qiaosen and Chen, Yulin and Zhou, Bowen and Liu, Zhiyuan and Sun, Maosong},
  booktitle={Proceedings of the 2023 conference on empirical methods in natural language processing},
  pages={4133--4145},
  year={2023}
}

@inproceedings{alora,
  title={Alora: Allocating low-rank adaptation for fine-tuning large language models},
  author={Liu, Zequan and Lyn, Jiawen and Zhu, Wei and Tian, Xing and Graham, Yvette},
  booktitle={Proceedings of the 2024 Conference of the North American Chapter of the Association for Computational Linguistics: Human Language Technologies (Volume 1: Long Papers)},
  pages={622--641},
  year={2024}
}

@inproceedings{moslora,
  title={Mixture-of-subspaces in low-rank adaptation},
  author={Wu, Taiqiang and Wang, Jiahao and Zhao, Zhe and Wong, Ngai},
  booktitle={Proceedings of the 2024 Conference on Empirical Methods in Natural Language Processing},
  pages={7880--7899},
  year={2024}
}

@article{loshchilov2017decoupled,
  title={Decoupled weight decay regularization},
  author={Loshchilov, Ilya and Hutter, Frank},
  journal={arXiv preprint arXiv:1711.05101},
  year={2017}
}

@article{hadsell2020embracing,
  title={Embracing change: Continual learning in deep neural networks},
  author={Hadsell, Raia and Rao, Dushyant and Rusu, Andrei A and Pascanu, Razvan},
  journal={Trends in cognitive sciences},
  volume={24},
  number={12},
  pages={1028--1040},
  year={2020},
  publisher={Elsevier}
}

@article{de2021continual,
  title={A continual learning survey: Defying forgetting in classification tasks},
  author={De Lange, Matthias and Aljundi, Rahaf and Masana, Marc and Parisot, Sarah and Jia, Xu and Leonardis, Ale{\v{s}} and Slabaugh, Gregory and Tuytelaars, Tinne},
  journal={IEEE transactions on pattern analysis and machine intelligence},
  volume={44},
  number={7},
  pages={3366--3385},
  year={2021},
  publisher={IEEE}
}

@article{pissa,
  title={Pissa: Principal singular values and singular vectors adaptation of large language models},
  author={Meng, Fanxu and Wang, Zhaohui and Zhang, Muhan},
  journal={Advances in Neural Information Processing Systems},
  volume={37},
  pages={121038--121072},
  year={2024}
}

@inproceedings{milora,
  title={MiLoRA: Efficient mixture of low-rank adaptation for large language models fine-tuning},
  author={Zhang, Jingfan and Zhao, Yi and Chen, Dan and Tian, Xing and Zheng, Huanran and Zhu, Wei},
  booktitle={Findings of the Association for Computational Linguistics: EMNLP 2024},
  pages={17071--17084},
  year={2024}
}

@article{lopez2017gradient,
  title={Gradient episodic memory for continual learning},
  author={Lopez-Paz, David and Ranzato, Marc'Aurelio},
  journal={Advances in neural information processing systems},
  volume={30},
  year={2017}
}

@inproceedings{l2p,
  title={Learning to prompt for continual learning},
  author={Wang, Zifeng and Zhang, Zizhao and Lee, Chen-Yu and Zhang, Han and Sun, Ruoxi and Ren, Xiaoqi and Su, Guolong and Perot, Vincent and Dy, Jennifer and Pfister, Tomas},
  booktitle={Proceedings of the IEEE/CVF Conference on Computer Vision and Pattern Recognition},
  pages={139--149},
  year={2022}
}

@inproceedings{dualprompt,
  title={Dualprompt: Complementary prompting for rehearsal-free continual learning},
  author={Wang, Zifeng and Zhang, Zizhao and Ebrahimi, Sayna and Sun, Ruoxi and Zhang, Han and Lee, Chen-Yu and Ren, Xiaoqi and Su, Guolong and Perot, Vincent and Dy, Jennifer and others},
  booktitle={European Conference on Computer Vision},
  pages={631--648},
  year={2022},
  organization={Springer}
}

@inproceedings{codaprompt,
  title={CODA-Prompt: COntinual Decomposed Attention-based Prompting for Rehearsal-Free Continual Learning},
  author={Smith, James Seale and Karlinsky, Leonid and Gutta, Vyshnavi and Cascante-Bonilla, Paola and Kim, Donghyun and Arbelle, Assaf and Panda, Rameswar and Feris, Rogerio and Kira, Zsolt},
  booktitle={Proceedings of the IEEE/CVF Conference on Computer Vision and Pattern Recognition},
  pages={11909--11919},
  year={2023}
}

@article{cifar,
  title={Learning multiple layers of features from tiny images},
  author={Krizhevsky, Alex and Hinton, Geoffrey and others},
  year={2009},
  publisher={Toronto, ON, Canada}
}

@article{sprompts,
  title={S-prompts learning with pre-trained transformers: An occam’s razor for domain incremental learning},
  author={Wang, Yabin and Huang, Zhiwu and Hong, Xiaopeng},
  journal={Advances in Neural Information Processing Systems},
  volume={35},
  pages={5682--5695},
  year={2022}
}

@inproceedings{luo2023class,
  title={Class-incremental exemplar compression for class-incremental learning},
  author={Luo, Zilin and Liu, Yaoyao and Schiele, Bernt and Sun, Qianru},
  booktitle={Proceedings of the IEEE/CVF Conference on Computer Vision and Pattern Recognition},
  pages={11371--11380},
  year={2023}
}

@article{aljundi2019gradient,
  title={Gradient based sample selection for online continual learning},
  author={Aljundi, Rahaf and Lin, Min and Goujaud, Baptiste and Bengio, Yoshua},
  journal={Advances in neural information processing systems},
  volume={32},
  year={2019}
}

@inproceedings{chaudhry2018riemannian,
  title={Riemannian walk for incremental learning: Understanding forgetting and intransigence},
  author={Chaudhry, Arslan and Dokania, Puneet K and Ajanthan, Thalaiyasingam and Torr, Philip HS},
  booktitle={Proceedings of the European conference on computer vision (ECCV)},
  pages={532--547},
  year={2018}
}

@inproceedings{liu2020mnemonics,
  title={Mnemonics training: Multi-class incremental learning without forgetting},
  author={Liu, Yaoyao and Su, Yuting and Liu, An-An and Schiele, Bernt and Sun, Qianru},
  booktitle={Proceedings of the IEEE/CVF conference on Computer Vision and Pattern Recognition},
  pages={12245--12254},
  year={2020}
}

@article{chaudhry2018efficient,
  title={Efficient lifelong learning with a-gem},
  author={Chaudhry, Arslan and Ranzato, Marc'Aurelio and Rohrbach, Marcus and Elhoseiny, Mohamed},
  journal={arXiv preprint arXiv:1812.00420},
  year={2018}
}

@article{kirkpatrick2017overcoming,
  title={Overcoming catastrophic forgetting in neural networks},
  author={Kirkpatrick, James and Pascanu, Razvan and Rabinowitz, Neil and Veness, Joel and Desjardins, Guillaume and Rusu, Andrei A and Milan, Kieran and Quan, John and Ramalho, Tiago and Grabska-Barwinska, Agnieszka and others},
  journal={Proceedings of the national academy of sciences},
  volume={114},
  number={13},
  pages={3521--3526},
  year={2017},
  publisher={National Acad Sciences}
}

@inproceedings{aljundi2018memory,
  title={Memory aware synapses: Learning what (not) to forget},
  author={Aljundi, Rahaf and Babiloni, Francesca and Elhoseiny, Mohamed and Rohrbach, Marcus and Tuytelaars, Tinne},
  booktitle={Proceedings of the European conference on computer vision (ECCV)},
  pages={139--154},
  year={2018}
}

@inproceedings{zenke2017continual,
  title={Continual learning through synaptic intelligence},
  author={Zenke, Friedemann and Poole, Ben and Ganguli, Surya},
  booktitle={International conference on machine learning},
  pages={3987--3995},
  year={2017},
  organization={PMLR}
}

@inproceedings{aljundi2019task,
  title={Task-free continual learning},
  author={Aljundi, Rahaf and Kelchtermans, Klaas and Tuytelaars, Tinne},
  booktitle={Proceedings of the IEEE/CVF Conference on Computer Vision and Pattern Recognition},
  pages={11254--11263},
  year={2019}
}

@inproceedings{der,
  title={Der: Dynamically expandable representation for class incremental learning},
  author={Yan, Shipeng and Xie, Jiangwei and He, Xuming},
  booktitle={Proceedings of the IEEE/CVF Conference on Computer Vision and Pattern Recognition},
  pages={3014--3023},
  year={2021}
}

@inproceedings{wang2022beef,
  title={BEEF: Bi-compatible class-incremental learning via energy-based expansion and fusion},
  author={Wang, Fu-Yun and Zhou, Da-Wei and Liu, Liu and Ye, Han-Jia and Bian, Yatao and Zhan, De-Chuan and Zhao, Peilin},
  booktitle={The Eleventh International Conference on Learning Representations},
  year={2023}
}

@inproceedings{wang2022foster,
  title={Foster: Feature boosting and compression for class-incremental learning},
  author={Wang, Fu-Yun and Zhou, Da-Wei and Ye, Han-Jia and Zhan, De-Chuan},
  booktitle={European conference on computer vision},
  pages={398--414},
  year={2022},
  organization={Springer}
}

@article{zhou2022model,
  title={A model or 603 exemplars: Towards memory-efficient class-incremental learning},
  author={Zhou, Da-Wei and Wang, Qi-Wei and Ye, Han-Jia and Zhan, De-Chuan},
  journal={arXiv preprint arXiv:2205.13218},
  year={2022}
}

@inproceedings{yan2022learning,
  title={Learning Bayesian sparse networks with full experience replay for continual learning},
  author={Yan, Qingsen and Gong, Dong and Liu, Yuhang and van den Hengel, Anton and Shi, Javen Qinfeng},
  booktitle={Proceedings of the IEEE/CVF Conference on Computer Vision and Pattern Recognition},
  pages={109--118},
  year={2022}
}

@article{nguyen2019toward,
  title={Toward understanding catastrophic forgetting in continual learning},
  author={Nguyen, Cuong V and Achille, Alessandro and Lam, Michael and Hassner, Tal and Mahadevan, Vijay and Soatto, Stefano},
  journal={arXiv preprint arXiv:1908.01091},
  year={2019}
}

@incollection{mccloskey1989catastrophic,
  title={Catastrophic interference in connectionist networks: The sequential learning problem},
  author={McCloskey, Michael and Cohen, Neal J},
  booktitle={Psychology of learning and motivation},
  volume={24},
  pages={109--165},
  year={1989},
  publisher={Elsevier}
}

@article{vaswani2017attention,
  title={Attention is all you need},
  author={Vaswani, Ashish and Shazeer, Noam and Parmar, Niki and Uszkoreit, Jakob and Jones, Llion and Gomez, Aidan N and Kaiser, {\L}ukasz and Polosukhin, Illia},
  journal={Advances in neural information processing systems},
  volume={30},
  year={2017}
}

@article{npcl,
  title={NPCL: Neural Processes for Uncertainty-Aware Continual Learning},
  author={Jha, Saurav and Gong, Dong and Zhao, He and Yao, Lina},
  journal={arXiv preprint arXiv:2310.19272},
  year={2023}
}

@article{dosovitskiy2020image,
  title={An image is worth 16x16 words: Transformers for image recognition at scale},
  author={Dosovitskiy, Alexey and Beyer, Lucas and Kolesnikov, Alexander and Weissenborn, Dirk and Zhai, Xiaohua and Unterthiner, Thomas and Dehghani, Mostafa and Minderer, Matthias and Heigold, Georg and Gelly, Sylvain and others},
  journal={arXiv preprint arXiv:2010.11929},
  year={2020}
}

@inproceedings{
jha2024clapclip,
title={{CLAP}4{CLIP}: Continual Learning with Probabilistic Finetuning for Vision-Language Models},
author={Saurav Jha and Dong Gong and Lina Yao},
booktitle={The Thirty-eighth Annual Conference on Neural Information Processing Systems},
year={2024},
url={https://openreview.net/forum?id=rF1YRtZfoJ}
}

@inproceedings{wang2024self,
  title={Self-expansion of pre-trained models with mixture of adapters for continual learning},
  author={Wang, Huiyi and Lu, Haodong and Yao, Lina and Gong, Dong},
  booktitle={Proceedings of the Computer Vision and Pattern Recognition Conference},
  pages={10087--10098},
  year={2025}
}

@article{mcdonnell2024ranpac,
  title={Ranpac: Random projections and pre-trained models for continual learning},
  author={McDonnell, Mark D and Gong, Dong and Parvaneh, Amin and Abbasnejad, Ehsan and van den Hengel, Anton},
  journal={Advances in Neural Information Processing Systems},
  volume={36},
  year={2024}
}

@inproceedings{inflora,
  title={InfLoRA: Interference-Free Low-Rank Adaptation for Continual Learning},
  author={Liang, Yan-Shuo and Li, Wu-Jun},
  booktitle={Proceedings of the IEEE/CVF Conference on Computer Vision and Pattern Recognition},
  pages={23638--23647},
  year={2024}
}

@inproceedings{zhang2024overcoming,
  title={Overcoming Generic Knowledge Loss with Selective Parameter Update},
  author={Zhang, Wenxuan and Janson, Paul and Aljundi, Rahaf and Elhoseiny, Mohamed},
  booktitle={Proceedings of the IEEE/CVF Conference on Computer Vision and Pattern Recognition},
  pages={24046--24056},
  year={2024}
}

@inproceedings{tang2025mind,
  title={Mind the interference: Retaining pre-trained knowledge in parameter efficient continual learning of vision-language models},
  author={Tang, Longxiang and Tian, Zhuotao and Li, Kai and He, Chunming and Zhou, Hantao and Zhao, Hengshuang and Li, Xiu and Jia, Jiaya},
  booktitle={European Conference on Computer Vision},
  pages={346--365},
  year={2025},
  organization={Springer}
}

@inproceedings{ease,
  title={Expandable subspace ensemble for pre-trained model-based class-incremental learning},
  author={Zhou, Da-Wei and Sun, Hai-Long and Ye, Han-Jia and Zhan, De-Chuan},
  booktitle={Proceedings of the IEEE/CVF Conference on Computer Vision and Pattern Recognition},
  pages={23554--23564},
  year={2024}
}

@article{codyra,
  title={Take Only What You Need: Rank Minimization as an Implicit Forgetting Regularizer in Continual Learning},
  author={Lu, Haodong and Zhao, Chongyang and Xue, Minhui and Yao, Lina and Moore, Kristen and Gong, Dong},
  journal={arXiv preprint arXiv:2412.01004},
  year={2024}
}

@article{kohonen1972correlation,
  title={Correlation matrix memories},
  author={Kohonen, Teuvo},
  journal={IEEE transactions on computers},
  volume={100},
  number={4},
  pages={353--359},
  year={1972},
  publisher={IEEE}
}

@article{anderson1972simple,
  title={A simple neural network generating an interactive memory},
  author={Anderson, James A},
  journal={Mathematical biosciences},
  volume={14},
  number={3-4},
  pages={197--220},
  year={1972},
  publisher={Elsevier}
}

@article{li2018measuring,
  title={Measuring the intrinsic dimension of objective landscapes},
  author={Li, Chunyuan and Farkhoor, Heerad and Liu, Rosanne and Yosinski, Jason},
  journal={arXiv preprint arXiv:1804.08838},
  year={2018}
}

@inproceedings{aghajanyan2020intrinsic,
  title={Intrinsic dimensionality explains the effectiveness of language model fine-tuning},
  author={Aghajanyan, Armen and Gupta, Sonal and Zettlemoyer, Luke},
  booktitle={Proceedings of the 59th annual meeting of the association for computational linguistics and the 11th international joint conference on natural language processing (volume 1: long papers)},
  pages={7319--7328},
  year={2021}
}

@article{de2019episodic,
  title={Episodic memory in lifelong language learning},
  author={de Masson D'Autume, Cyprien and Ruder, Sebastian and Kong, Lingpeng and Yogatama, Dani},
  journal={Advances in Neural Information Processing Systems},
  volume={32},
  year={2019}
}

@article{qin2021lfpt5,
  title={Lfpt5: A unified framework for lifelong few-shot language learning based on prompt tuning of t5},
  author={Qin, Chengwei and Joty, Shafiq},
  journal={arXiv preprint arXiv:2110.07298},
  year={2021}
}

@inproceedings{
razdaibiedina2023progressive,
title={Progressive Prompts: Continual Learning for Language Models},
author={Anastasia Razdaibiedina and Yuning Mao and Rui Hou and Madian Khabsa and Mike Lewis and Amjad Almahairi},
booktitle={The Eleventh International Conference on Learning Representations },
year={2023},
url={https://openreview.net/forum?id=UJTgQBc91_}
}

@inproceedings{wang2023orthogonal,
  title={Orthogonal subspace learning for language model continual learning},
  author={Wang, Xiao and Chen, Tianze and Ge, Qiming and Xia, Han and Bao, Rong and Zheng, Rui and Zhang, Qi and Gui, Tao and Huang, Xuan-Jing},
  booktitle={Findings of the Association for Computational Linguistics: EMNLP 2023},
  pages={10658--10671},
  year={2023}
}

@article{qiao2024learn,
  title={Learn more, but bother less: parameter efficient continual learning},
  author={Qiao, Fuli and Mahdavi, Mehrdad},
  journal={Advances in Neural Information Processing Systems},
  volume={37},
  pages={97476--97498},
  year={2024}
}

@article{lepikhin2020gshard,
  title={Gshard: Scaling giant models with conditional computation and automatic sharding},
  author={Lepikhin, Dmitry and Lee, HyoukJoong and Xu, Yuanzhong and Chen, Dehao and Firat, Orhan and Huang, Yanping and Krikun, Maxim and Shazeer, Noam and Chen, Zhifeng},
  journal={arXiv preprint arXiv:2006.16668},
  year={2020}
}

@article{shazeer2017outrageously,
  title={Outrageously large neural networks: The sparsely-gated mixture-of-experts layer},
  author={Shazeer, Noam and Mirhoseini, Azalia and Maziarz, Krzysztof and Davis, Andy and Le, Quoc and Hinton, Geoffrey and Dean, Jeff},
  journal={arXiv preprint arXiv:1701.06538},
  year={2017}
}

@inproceedings{dai2024deepseekmoe,
  title={Deepseekmoe: Towards ultimate expert specialization in mixture-of-experts language models},
  author={Dai, Damai and Deng, Chengqi and Zhao, Chenggang and Xu, RX and Gao, Huazuo and Chen, Deli and Li, Jiashi and Zeng, Wangding and Yu, Xingkai and Wu, Yu and others},
  booktitle={Proceedings of the 62nd Annual Meeting of the Association for Computational Linguistics (Volume 1: Long Papers)},
  pages={1280--1297},
  year={2024}
}

@article{fedus2022switch,
  title={Switch transformers: Scaling to trillion parameter models with simple and efficient sparsity},
  author={Fedus, William and Zoph, Barret and Shazeer, Noam},
  journal={Journal of Machine Learning Research},
  volume={23},
  number={120},
  pages={1--39},
  year={2022}
}

@article{raffel2020exploring,
  title={Exploring the limits of transfer learning with a unified text-to-text transformer},
  author={Raffel, Colin and Shazeer, Noam and Roberts, Adam and Lee, Katherine and Narang, Sharan and Matena, Michael and Zhou, Yanqi and Li, Wei and Liu, Peter J},
  journal={Journal of machine learning research},
  volume={21},
  number={140},
  pages={1--67},
  year={2020}
}

@misc{codealpaca,
  author = {Sahil Chaudhary},
  title = {Code Alpaca: An Instruction-following LLaMA model for code generation},
  year = {2023},
  publisher = {GitHub},
  journal = {GitHub repository},
  howpublished = {\url{https://github.com/sahil280114/codealpaca}},
}

@article{chen2021evaluating,
  title={Evaluating large language models trained on code},
  author={Chen, Mark and Tworek, Jerry and Jun, Heewoo and Yuan, Qiming and Pinto, Henrique Ponde De Oliveira and Kaplan, Jared and Edwards, Harri and Burda, Yuri and Joseph, Nicholas and Brockman, Greg and others},
  journal={arXiv preprint arXiv:2107.03374},
  year={2021}
}

@article{hendryckstest2021,
      title={Measuring Massive Multitask Language Understanding},
      author={Dan Hendrycks and Collin Burns and Steven Basart and Andy Zou and Mantas Mazeika and Dawn Song and Jacob Steinhardt},
      journal={International Conference on Learning Representations},
      year={2021}
    }

@article{chen2023octavius,
  title={Octavius: Mitigating task interference in mllms via lora-moe},
  author={Chen, Zeren and Wang, Ziqin and Wang, Zhen and Liu, Huayang and Yin, Zhenfei and Liu, Si and Sheng, Lu and Ouyang, Wanli and Qiao, Yu and Shao, Jing},
  journal={arXiv preprint arXiv:2311.02684},
  year={2023}
}

@article{yang2024moral,
  title={MoRAL: MoE Augmented LoRA for LLMs' Lifelong Learning},
  author={Yang, Shu and Ali, Muhammad Asif and Wang, Cheng-Long and Hu, Lijie and Wang, Di},
  journal={arXiv preprint arXiv:2402.11260},
  year={2024}
}

@article{chen2024llava,
  title={Llava-mole: Sparse mixture of lora experts for mitigating data conflicts in instruction finetuning mllms},
  author={Chen, Shaoxiang and Jie, Zequn and Ma, Lin},
  journal={arXiv preprint arXiv:2401.16160},
  year={2024}
}

@article{li2024mixlora,
  title={Mixlora: Enhancing large language models fine-tuning with lora-based mixture of experts},
  author={Li, Dengchun and Ma, Yingzi and Wang, Naizheng and Ye, Zhengmao and Cheng, Zhiyuan and Tang, Yinghao and Zhang, Yan and Duan, Lei and Zuo, Jie and Yang, Cal and others},
  journal={arXiv preprint arXiv:2404.15159},
  year={2024}
}

@article{dou2023loramoe,
  title={LoRAMoE: Alleviate world knowledge forgetting in large language models via MoE-style plugin},
  author={Dou, Shihan and Zhou, Enyu and Liu, Yan and Gao, Songyang and Zhao, Jun and Shen, Wei and Zhou, Yuhao and Xi, Zhiheng and Wang, Xiao and Fan, Xiaoran and others},
  journal={arXiv preprint arXiv:2312.09979},
  year={2023}
}

@article{garg2023tic,
  title={Tic-clip: Continual training of clip models},
  author={Garg, Saurabh and Farajtabar, Mehrdad and Pouransari, Hadi and Vemulapalli, Raviteja and Mehta, Sachin and Tuzel, Oncel and Shankar, Vaishaal and Faghri, Fartash},
  journal={arXiv preprint arXiv:2310.16226},
  year={2023}
}

@article{rusu2016progressive,
  title={Progressive neural networks},
  author={Rusu, Andrei A and Rabinowitz, Neil C and Desjardins, Guillaume and Soyer, Hubert and Kirkpatrick, James and Kavukcuoglu, Koray and Pascanu, Razvan and Hadsell, Raia},
  journal={arXiv preprint arXiv:1606.04671},
  year={2016}
}

@article{zhang2015character,
  title={Character-level convolutional networks for text classification},
  author={Zhang, Xiang and Zhao, Junbo and LeCun, Yann},
  journal={Advances in neural information processing systems},
  volume={28},
  year={2015}
}

@article{wang2018glue,
  title={GLUE: A multi-task benchmark and analysis platform for natural language understanding},
  author={Wang, Alex and Singh, Amanpreet and Michael, Julian and Hill, Felix and Levy, Omer and Bowman, Samuel R},
  journal={arXiv preprint arXiv:1804.07461},
  year={2018}
}

@article{wang2019superglue,
  title={Superglue: A stickier benchmark for general-purpose language understanding systems},
  author={Wang, Alex and Pruksachatkun, Yada and Nangia, Nikita and Singh, Amanpreet and Michael, Julian and Hill, Felix and Levy, Omer and Bowman, Samuel},
  journal={Advances in neural information processing systems},
  volume={32},
  year={2019}
}

@inproceedings{zheng2024llamafactory,
  title={LlamaFactory: Unified Efficient Fine-Tuning of 100+ Language Models},
  author={Yaowei Zheng and Richong Zhang and Junhao Zhang and Yanhan Ye and Zheyan Luo and Zhangchi Feng and Yongqiang Ma},
  booktitle={Proceedings of the 62nd Annual Meeting of the Association for Computational Linguistics (Volume 3: System Demonstrations)},
  address={Bangkok, Thailand},
  publisher={Association for Computational Linguistics},
  year={2024}
}

@misc{eval_harness,
  author       = {Gao, Leo and Tow, Jonathan and Abbasi, Baber and Biderman, Stella and Black, Sid and DiPofi, Anthony and Foster, Charles and Golding, Laurence and Hsu, Jeffrey and Le Noac'h, Alain and Li, Haonan and McDonell, Kyle and Muennighoff, Niklas and Ociepa, Chris and Phang, Jason and Reynolds, Laria and Schoelkopf, Hailey and Skowron, Aviya and Sutawika, Lintang and Tang, Eric and Thite, Anish and Wang, Ben and Wang, Kevin and Zou, Andy},
  title        = {The Language Model Evaluation Harness},
  month        = 07,
  year         = 2024,
  publisher    = {Zenodo},
  version      = {v0.4.3},
  doi          = {10.5281/zenodo.12608602},
  url          = {https://zenodo.org/records/12608602}
}

@article{zhang2025lori,
  title={Lori: Reducing cross-task interference in multi-task low-rank adaptation},
  author={Zhang, Juzheng and You, Jiacheng and Panda, Ashwinee and Goldstein, Tom},
  journal={arXiv preprint arXiv:2504.07448},
  year={2025}
}

@article{touvron2023llama,
  title={Llama 2: Open foundation and fine-tuned chat models},
  author={Touvron, Hugo and Martin, Louis and Stone, Kevin and Albert, Peter and Almahairi, Amjad and Babaei, Yasmine and Bashlykov, Nikolay and Batra, Soumya and Bhargava, Prajjwal and Bhosale, Shruti and others},
  journal={arXiv preprint arXiv:2307.09288},
  year={2023}
}

@inproceedings{
wu2024mixture,
title={Mixture of LoRA Experts},
author={Xun Wu and Shaohan Huang and Furu Wei},
booktitle={The Twelfth International Conference on Learning Representations},
year={2024}
}

@inproceedings{
theory,
title={Theory on Mixture-of-Experts in Continual Learning},
author={Hongbo Li and Sen Lin and Lingjie Duan and Yingbin Liang and Ness Shroff},
booktitle={The Thirteenth International Conference on Learning Representations},
year={2025}
}

@article{wang2023trace,
  title={Trace: A comprehensive benchmark for continual learning in large language models},
  author={Wang, Xiao and Zhang, Yuansen and Chen, Tianze and Gao, Songyang and Jin, Senjie and Yang, Xianjun and Xi, Zhiheng and Zheng, Rui and Zou, Yicheng and Gui, Tao and others},
  journal={arXiv preprint arXiv:2310.06762},
  year={2023}
}

@article{wang2022super,
  title={Super-naturalinstructions: Generalization via declarative instructions on 1600+ nlp tasks},
  author={Wang, Yizhong and Mishra, Swaroop and Alipoormolabashi, Pegah and Kordi, Yeganeh and Mirzaei, Amirreza and Arunkumar, Anjana and Ashok, Arjun and Dhanasekaran, Arut Selvan and Naik, Atharva and Stap, David and others},
  journal={arXiv preprint arXiv:2204.07705},
  year={2022}
}

@article{zhao2024sapt,
  title={Sapt: A shared attention framework for parameter-efficient continual learning of large language models},
  author={Zhao, Weixiang and Wang, Shilong and Hu, Yulin and Zhao, Yanyan and Qin, Bing and Zhang, Xuanyu and Yang, Qing and Xu, Dongliang and Che, Wanxiang},
  journal={arXiv preprint arXiv:2401.08295},
  year={2024}
}

@inproceedings{
Yoon2020Scalable,
title={Scalable and Order-robust Continual Learning with Additive Parameter Decomposition},
author={Jaehong Yoon and Saehoon Kim and Eunho Yang and Sung Ju Hwang},
booktitle={International Conference on Learning Representations},
year={2020},
}

@inproceedings{wu2025sd,
  title={SD-LoRA: Scalable Decoupled Low-Rank Adaptation for Class Incremental Learning},
  author={Wu, Yichen and Piao, Hongming and Huang, Long-Kai and Wang, Renzhen and Li, Wanhua and Pfister, Hanspeter and Meng, Deyu and Ma, Kede and Wei, Ying},
  booktitle={The Thirteenth International Conference on Learning Representations},
  year={2025}
}

@article{grattafiori2024llama,
  title={The llama 3 herd of models},
  author={Grattafiori, Aaron and Dubey, Abhimanyu and Jauhri, Abhinav and Pandey, Abhinav and Kadian, Abhishek and Al-Dahle, Ahmad and Letman, Aiesha and Mathur, Akhil and Schelten, Alan and Vaughan, Alex and others},
  journal={arXiv e-prints},
  pages={arXiv--2407},
  year={2024}
}

@article{team2024gemma,
  title={Gemma: Open models based on gemini research and technology},
  author={Team, Gemma and Mesnard, Thomas and Hardin, Cassidy and Dadashi, Robert and Bhupatiraju, Surya and Pathak, Shreya and Sifre, Laurent and Rivi{\`e}re, Morgane and Kale, Mihir Sanjay and Love, Juliette and others},
  journal={arXiv preprint arXiv:2403.08295},
  year={2024}
}

@article{brown2020language,
  title={Language models are few-shot learners},
  author={Brown, Tom and Mann, Benjamin and Ryder, Nick and Subbiah, Melanie and Kaplan, Jared D and Dhariwal, Prafulla and Neelakantan, Arvind and Shyam, Pranav and Sastry, Girish and Askell, Amanda and others},
  journal={Advances in neural information processing systems},
  volume={33},
  pages={1877--1901},
  year={2020}
}

@article{wang2025hide,
  title={HIDE-PET: continual learning via hierarchical decomposition of parameter-efficient tuning},
  author={Wang, Liyuan and Xie, Jingyi and Zhang, Xingxing and Su, Hang and Zhu, Jun},
  journal={IEEE Transactions on Pattern Analysis and Machine Intelligence},
  year={2025},
  publisher={IEEE}
}

@inproceedings{qian2025treelora,
  title={TreeLoRA: Efficient Continual Learning via Layer-Wise LoRAs Guided by a Hierarchical Gradient-Similarity Tree},
  author={Qian, Yu-Yang and Xu, Yuan-Ze and Zhang, Zhen-Yu and Zhao, Peng and Zhou, Zhi-Hua},
  booktitle={International Conference on Machine Learning},
  pages={50066--50085},
  year={2025},
  organization={PMLR}
}

@article{zhao2024learning,
  title={Learning Mamba as a continual learner: Meta-learning selective state space models for efficient continual learning},
  author={Zhao, Chongyang and Gong, Dong},
  journal={arXiv preprint arXiv:2412.00776},
  year={2024}
}

@inproceedings{zhao2026token,
  title={On Token's Dilemma: Dynamic MoE with Drift-Aware Token Assignment for Continual Learning of Large Vision Language Models},
  author={Zhao, Chongyang and Li, Mingsong and Lu, Haodong and Gong, Dong},
  booktitle={Proceedings of the IEEE/CVF Conference on Computer Vision and Pattern Recognition},
  year={2026}
}

@inproceedings{zhou2025same,
  title={Same: Learning generic language-guided visual navigation with state-adaptive mixture of experts},
  author={Zhou, Gengze and Hong, Yicong and Wang, Zun and Zhao, Chongyang and Bansal, Mohit and Wu, Qi},
  booktitle={Proceedings of the IEEE/CVF International Conference on Computer Vision},
  pages={7794--7807},
  year={2025}
}

@article{he2024mixture,
  title={Mixture of a million experts},
  author={He, Xu Owen},
  journal={arXiv preprint arXiv:2407.04153},
  year={2024}
}

@inproceedings{
ludziejewski2024scaling,
title={Scaling Laws for Fine-Grained Mixture of Experts},
author={Jan Ludziejewski and Jakub Krajewski and Kamil Adamczewski and Maciej Pi{\'o}ro and Micha{\l} Krutul and Szymon Antoniak and Kamil Ciebiera and Krystian Kr{\'o}l and Tomasz Odrzyg{\'o}{\'z}d{\'z} and Piotr Sankowski and Marek Cygan and Sebastian Jaszczur},
booktitle={Forty-first International Conference on Machine Learning},
year={2024},
url={https://openreview.net/forum?id=yoqdlynCRs}
}

@article{tong2026model,
  title={Model Inversion with Layer-Specific Modeling and Alignment for Data-Free Continual Learning},
  author={Tong, Ruilin and Lu, Haodong and Liu, Yuhang and Gong, Dong},
  journal={Advances in Neural Information Processing Systems},
  volume={38},
  pages={22091--22126},
  year={2026}
}

@inproceedings{tong2025coreset,
  title={Coreset selection via reducible loss in continual learning},
  author={Tong, Ruilin and Liu, Yuhang and Shi, Javen Qinfeng and Gong, Dong},
  booktitle={The Thirteenth International Conference on Learning Representations},
  year={2025}
}

@article{le2024mixture,
  title={Mixture of experts meets prompt-based continual learning},
  author={Le, Minh and Nguyen, An and Nguyen, Huy and Nguyen, Trang and Pham, Trang and Van Ngo, Linh and Ho, Nhat},
  journal={Advances in Neural Information Processing Systems},
  volume={37},
  pages={119025--119062},
  year={2024}
}
\bibliographystyle{icml2026}

\clearpage
\newpage
\appendix

%=============================================================
\section{Experiment Details}
\label{supp:details}
%=============================================================

\subsection{Detailed Experiment Settings}
\label{supp:settings}

\noindent\textbf{Continual Learning of CLIP on X-TAIL and MTIL.}
The MTIL setting consists of 1,201 classes drawn from 11 diverse datasets:
Aircraft~\citep{maji2013fine},
Caltech101~\citep{fei2004learning},
CIFAR100~\citep{cifar},
DTD~\citep{cimpoi2014describing},
EuroSAT~\citep{helber2019eurosat},
Flowers~\citep{nilsback2008automated},
Food~\citep{bossard2014food},
MNIST~\citep{deng2012mnist},
OxfordPet~\citep{parkhi2012cats},
Cars~\citep{krause20133d}, and SUN397~\citep{xiao2010sun}. In the X-TAIL setting, a total of 10 datasets are used, with CIFAR100~\citep{cifar} excluded to prevent domain overlap, following the protocol in~\citep{rail}. In line with \citep{rail}, we use a 5-shot split for MTIL and a 16-shot split for X-TAIL.

We follow the experimental setups in \citep{zscl, boosting, rail} and use the CLIP model with a ViT-B/16 backbone \citep{clip} for all experiments. By default, \ours is applied to every pre-trained weight matrix in both the vision and text encoders, with an initial rank of 16 per update. Each task is trained for 500 iterations using AdamW \citep{loshchilov2017decoupled} with a learning rate of $5e-4$. During continual learning, we freeze the ranks learned from previous tasks and initialize new $r=16$ ranks for each incoming task. The memory activation budget is set to 16 throughout all tasks. We set the temperature $\tau_\text{\ours}=0.01$ and the threshold $\delta=0.2$.

\noindent\textbf{Continual Learning of LLMs on TRACE.} We evaluate \ours on the TRACE benchmark \citep{wang2023trace}, a comprehensive suite designed to assess continual learning in LLMs such as LLaMA \citep{touvron2023llama,grattafiori2024llama} and Gemma \citep{team2024gemma}. TRACE standardizes eight diverse datasets spanning domain-specific understanding (C-STANCE, FOMC), reasoning (ScienceQA, NumGLUE-cm/ds), summarization (MeetingBank), code generation (Py150), and multilingual simplification (20Minuten). Following standard protocols, each task consists of 5,000 training and 2,000 testing examples. We employ task-specific metrics to capture performance nuances: Accuracy for classification and reasoning tasks; ROUGE-L for summarization; SARI for simplification; and similarity scores for code generation.

For each incoming task, we initialize $r=16$ new ranks while maintaining a fixed memory activation budget of $k=16$ across all tasks. We set the softmax temperature $\tau_\text{\ours}=0.03$ and the inference threshold $\delta=0.2$. Optimization proceeds with a learning rate of $5\times 10^{-4}$ and a batch size of 4, utilizing a maximum context length of 1024 tokens. All experiments leverage DeepSpeed ZeRO-2 with BF16 mixed-precision on a cluster of four Nvidia H100 GPUs.

\noindent\textbf{Continual Learning of LMs on language classification tasks.}
We follow the protocol of previous work in continually fine-tuning the T5-large \citep{raffel2020exploring} and LLaMA2-7B \citep{touvron2023llama} on a suite of text-classification tasks.  We train on five standard benchmarks---AG News, Amazon Reviews, Yelp Reviews, DBpedia, and Yahoo Answers---using three distinct task orderings drawn from \citep{qin2021lfpt5,razdaibiedina2023progressive,wang2023orthogonal,qiao2024learn}.  To probe longer sequences, we extend this to a 15-dataset stream (Table \ref{tab:supp_llm_data}), incorporating tasks from the original CL benchmark \citep{zhang2015character}, GLUE \citep{wang2018glue}, SuperGLUE \citep{wang2019superglue}, and the IMDB movie reviews corpus.  Natural language prompts for each task are presented in Table \ref{tab:supp_llm_prompt}, with NLI tasks (MNLI, RTE, CB), sentiment classification (Amazon, Yelp, SST-2, IMDB), and topic classification (AG News, DBpedia, Yahoo).

We evaluate three distinct task sequences for both the standard CL and 15-task benchmarks (Table \ref{tab:supp_llm_cl_order}). After completing the final task in each stream, we report the average accuracy across all tasks. All experiments use one epoch per task with DeepSpeed, a fixed learning rate of $1e-3$, batch size 64, and dropout of 0.1. \ours is applied to both the query and key projection matrices within attention layers, initializing $r=8$ new ranks for each incoming task similarly as in \citep{wang2023orthogonal,qiao2024learn}. We maintain a constant activation budget of 4 ranks throughout continual learning, set the temperature $\tau_{\text{\ours}}=0.1$, and the threshold $\delta=0.2$.

\noindent\textbf{Generalization and forgetting on unseen tasks after standard fine-tuning.}  
To assess effects on pre-trained general knowledge, we fine-tune Llama3.1-8B \citep{grattafiori2024llama} on the CodeAlpaca code-generation dataset \citep{codealpaca} using llama-Factory \citep{zheng2024llamafactory} and evaluate using lm-eval-harness \citep{eval_harness} on zero-shot in-domain performance on HumanEval \citep{chen2021evaluating}, as well as out-of-domain accuracy on a broad selection of MMLU \citep{hendryckstest2021} subjects---Formal Logic, Philosophy, World Religions, Economics, Public Relations, STEM, Physics, and Machine Learning. 

In this experiment, \ours is applied to all linear weight matrices of the pre-trained model. We fine-tune on CodeAlpaca with a batch size of 32 over 3 epochs and a cosine learning-rate schedule starting at $5e-4$. We train with $r=16$ ranks and enforce a constant activation budget of 4 ranks. The self-routed gating uses a temperature $\tau_{\text{\ours}} = 0.5$ and a threshold $\delta = 0.2$. We observe that, due to variations in hidden representations across architectures, the optimal temperature setting can differ across different pre-trained models.

\subsection{Evaluation Metrics}
\label{supp:metrics}
To strictly evaluate the plasticity and stability of our method, we employ the following metrics. Let $N$ denote the total number of learned tasks, and $A_{i,j}$ denote the performance (e.g., accuracy, ROUGE, or similarity score) on task $j$ after training on task $i$.

\paragraph{CLIP Continual Learning Metrics.}
For the CLIP experiments (MTIL and X-TAIL), we report the following metrics as in \citep{zscl,boosting,rail}:
\begin{itemize}
    \item \textbf{Transfer Accuracy (Transfer):} indicates the model's zero-shot performance on future domains before they are learned, measures the extent to which the pre-trained zero-shot ability is preserved (or improved) throughout incremental learning.
    \item \textbf{Average Accuracy (Average):} indicates the average accuracy of all learning steps across all domains. It captures the comprehensive performance stability throughout the entire incremental training process.
    \item \textbf{Last Accuracy (Last):} represent the model's performance on all seen domains after the training is fully completed. 
\end{itemize}

\paragraph{LLM Continual Learning Metrics.}
For the TRACE benchmark, we adopt the standard metrics as in \citep{wang2023trace}:
\begin{itemize}
    \item \textbf{Overall Performance (OP):} The average performance across all learned tasks at the end of training.
    \begin{equation}
        \text{OP} = \frac{1}{N} \sum_{j=1}^{N} A_{N,j},
    \end{equation}
    where $A_{i,j}$ represents the performance on task $j$ after learning task $i$.
    \item \textbf{Backward Transfer (BWT):} Measures the average performance degradation (forgetting) of previous tasks $j < N$ after learning new tasks. A lower BWT indicates less forgetting (better stability).
    \begin{equation}
        \text{BWT} = \frac{1}{N-1} \sum_{j=1}^{N-1} (A_{j,j} - A_{N,j})
    \end{equation}
\end{itemize}

\subsection{Datasets, Prompts, and Task Orderings}
\label{supp:data}

Table~\ref{tab:supp_llm_data} summarizes the 15 datasets used in our language model continual learning experiments. Table~\ref{tab:supp_llm_prompt} presents the natural language prompts for each task type. Table~\ref{tab:supp_llm_cl_order} lists the six task-sequence orderings used across our experiments.

\begin{table*}[htb]
    \centering
\caption{Details of the 15 datasets used in our continual-learning experiments using LMs. NLI denotes natural language inference, and QA denotes question-answering tasks. The first five tasks comprise the standard CL benchmark; the remaining ten tasks are used for the extended long-sequence evaluations.}
\label{tab:supp_llm_data}
\begin{tabular}{l|llll}
\hline
\textbf{Dataset name} & \textbf{Category} & \textbf{Task}             & \textbf{Domain}     & \textbf{Metric} \\ \hline
1. Yelp               & CL Benchmark      & sentiment analysis        & Yelp reviews        & accuracy        \\
2. Amazon             & CL Benchmark      & sentiment analysis        & Amazon reviews      & accuracy        \\
3. DBpedia            & CL Benchmark      & topic classification      & Wikipedia           & accuracy        \\
4. Yahoo              & CL Benchmark      & topic classification      & Yahoo Q\&A          & accuracy        \\
5. AG News            & CL Benchmark      & topic classification      & news                & accuracy        \\
6. MNLI               & GLUE              & NLI                       & various             & accuracy        \\
7. QQP                & GLUE              & paragraph detection       & Quora               & accuracy        \\
8. RTE                & GLUE              & NLI                       & news, Wikipedia     & accuracy        \\
9. SST-2              & GLUE              & sentiment analysis        & movie reviews       & accuracy        \\
10. WiC               & SuperGLUE         & word sense disambiguation & lexical databases   & accuracy        \\
11. CB                & SuperGLUE         & NLI                       & various             & accuracy        \\
12. COPA              & SuperGLUE         & QA                        & blogs, encyclopedia & accuracy        \\
13. BoolQA            & SuperGLUE         & boolean QA                & Wikipedia           & accuracy        \\
14. MultiRC           & SuperGLUE         & QA                        & various             & accuracy        \\
15. IMDB              & SuperGLUE         & sentiment analysis        & movie reviews       & accuracy        \\ \hline
\end{tabular}
\end{table*}

\begin{table*}[htb]
    \centering
    \caption{Instructions for different tasks.}
\label{tab:supp_llm_prompt}
\begin{tabular}{cl}
\hline
\textbf{Task}                                                       & \multicolumn{1}{c}{\textbf{Prompts}}                                                                                                                                  \\ \hline
NLI                                                                 & \begin{tabular}[c]{@{}l@{}}What is the logical relationship between the ``sentence 1'' and the ``sentence 2''? \\ Choose one from the option.\end{tabular}                \\ \hline
QQP                                                                 & \begin{tabular}[c]{@{}l@{}}Whether the ``first sentence'' and the ``second sentence'' have the same meaning? \\ Choose one from the option.\end{tabular}                  \\ \hline
\begin{tabular}[c]{@{}c@{}}SC\end{tabular}   & What is the sentiment of the following paragraph? Choose one from the option.                                                                                         \\ \hline
\begin{tabular}[c]{@{}c@{}}TC\end{tabular} & What is the topic of the following paragraph? Choose one from the option.                                                                                             \\ \hline
BoolQA                                                              & \begin{tabular}[c]{@{}l@{}}According to the following passage, is the question true or false? Choose one \\ from the option.\end{tabular}                             \\ \hline
MultiRC                                                             & \begin{tabular}[c]{@{}l@{}}According to the following passage and question, is the candidate answer true \\ or false? Choose one from the option.\end{tabular}        \\ \hline
WiC                                                                 & \begin{tabular}[c]{@{}l@{}}Given a word and two sentences, whether the word is used with the same sense \\ in both sentence? Choose one from the option.\end{tabular} \\ \hline
\end{tabular}
\end{table*}

\begin{table*}[h]
\centering
\caption{The six task-sequence orders used in our continual learning experiments. Sequences 1--3 follow the standard CL benchmarks employed in prior work. Sequences 4--6 extend to longer 15-task streams, as introduced in \citep{razdaibiedina2023progressive}.}
\label{tab:supp_llm_cl_order}
\small
\begin{tabular}{cc}
\hline
\textbf{Order}  & \textbf{Task Sequence}                                                                                                                                \\ \hline
1                 & dbpedia $\to$ amazon $\to$ yahoo $\to$ ag                                                                                                                         \\
2                & dbpedia $\to$ amazon $\to$ ag $\to$ yahoo                                                                                                                         \\
3                  & yahoo $\to$ amazon $\to$ ag $\to$ dbpedia                                                                                                                         \\ \hline
4                   & \begin{tabular}[c]{@{}l@{}}mnli $\to$ cb $\to$ wic $\to$ copa $\to$ qqp $\to$ boolqa $\to$ rte $\to$ imdb $\to$\\ yelp $\to$ amazon $\to$ sst-2 $\to$ dbpedia $\to$ ag $\to$ multirc $\to$ yahoo\end{tabular} \\
5                 & \begin{tabular}[c]{@{}l@{}}multirc $\to$ boolqa $\to$ wic $\to$ mnli $\to$ cb $\to$ copa $\to$ qqp $\to$ rte\\ $\to$ imdb $\to$ sst-2 $\to$ dbpedia $\to$ ag $\to$ yelp $\to$ amazon $\to$ yahoo\end{tabular} \\
6                  & \begin{tabular}[c]{@{}l@{}}yelp $\to$ amazon $\to$ mnli $\to$ cb $\to$ copa $\to$ qqp $\to$ rte $\to$ imdb $\to$\\ sst-2 $\to$ dbpedia $\to$ ag $\to$ yahoo $\to$ multirc $\to$ boolqa $\to$ wic\end{tabular} \\ \hline
\end{tabular}
\end{table*}

%=============================================================
\section{Additional Experimental Results}
\label{supp:results}
%=============================================================

\subsection{Continual Learning of CLIP}
\label{supp:clip_results}

\subsubsection{Multi-domain Task Incremental Learning (MTIL)}
\label{supp:mtil}

We evaluate \ours in the few-shot MTIL setting (Table~\ref{tab:few_mtil_full}) under the same protocols as \citep{boosting,rail,codyra}. Consistent with the results observed in the X-TAIL setting, our method demonstrates clear superiority in this scenario. 

In this challenging scenario, the model must learn 11 diverse tasks sequentially, with only five examples per class. 
These findings validate that our self-activated sparse mixture of rank-1 memories framework both facilitates continual acquisition of new knowledge and mitigates forgetting from the pre-trained model and earlier tasks.

\begin{table*}[ht]
\centering
    \caption{Comparisons on 5-shot MTIL setting. Following the same protocol as in \citep{boosting,rail,codyra}.}
    \label{tab:few_mtil_full}
    \resizebox{\linewidth}{!}{
	\begin{tabular}{l>{\centering\arraybackslash}p{1cm} >{\centering\arraybackslash}p{1cm}>{\centering\arraybackslash}p{1cm} >{\centering\arraybackslash}p{1cm} >{\centering\arraybackslash}p{1cm} >{\centering\arraybackslash}p{1cm} >{\centering\arraybackslash}p{1cm} >{\centering\arraybackslash}p{1cm} >{\centering\arraybackslash}p{1cm} >{\centering\arraybackslash}p{1cm} >{\centering\arraybackslash}p{1cm} >{\centering\arraybackslash}p{1.5cm}}
		\toprule
               {\quad} \makecell[c]{Method} & \makecell[c]{\rotatebox{90}{Aircraft}} & \makecell[c]{\rotatebox{90}{Caltech101}} & \makecell[c]{\rotatebox{90}{CIFAR100}} & \makecell[c]{\rotatebox{90}{DTD}} & \makecell[c]{\rotatebox{90}{EuroSAT}} & \makecell[c]{\rotatebox{90}{Flowers}} & \makecell[c]{\rotatebox{90}{Food}} & \makecell[c]{\rotatebox{90}{MNIST}} & \makecell[c]{\rotatebox{90}{OxfordPet}} & \makecell[c]{\rotatebox{90}{Cars}} & \makecell[c]{\rotatebox{90}{SUN397}} & \makecell[c]{\textit{Average}} \\
		\midrule
  \rowcolor{RowCOLOR}\multicolumn{13}{l}{\emph{CLIP}}\\
  {\quad}Zero-shot \citep{clip}& 24.3 & 88.4 & 68.2 & 44.6 & 54.9 & 71.0 & 88.5 & 59.4 & 89.0 & 64.7 & 65.2 & \cellcolor{CellCOLOR!50}65.3  \\
            \midrule
            \rowcolor{RowCOLOR}\multicolumn{13}{l}{\emph{Transfer}}\\
            {\quad}Zero-shot \citep{clip} & -- & {88.4}&{68.2}& 44.6 &\highred{54.9}&\highred{71.0}& \highblue{88.5} & 59.6 & {89.0} &\highblue{64.7}&\highred{65.2}& {\cellcolor{CellCOLOR!50}}{69.4}\\
            {\quad}LwF \citep{li2017learning} & -- & 72.1 & 49.2 & 35.9 & 44.5 & 41.1 & 66.6 & 50.5 & 69.0 & 19.0 & 51.7 & \cellcolor{CellCOLOR!50}50.0   \\
            {\quad}LwF-VR \citep{ding2022don} & -- & 82.2 & 62.5 & 40.1 & 40.1 & 56.3 & 80.0 & 60.9 & 77.6 & 40.5 & 60.8 & \cellcolor{CellCOLOR!50}60.1   \\
            {\quad}WiSE-FT \citep{wortsman2022robust} &--  & 77.6 & 60.0 & 41.3 & 39.4 & 53.0 & 76.6 & 58.1 & 75.5 & 37.3 & 58.2 & \cellcolor{CellCOLOR!50}57.7   \\
            {\quad}ZSCL \citep{zscl} & -- & 84.0 & 68.1 & {44.8} & 46.8 & 63.6 & 84.9 & 61.4 & 81.4 & 55.5 & 62.2 & {\cellcolor{CellCOLOR!50}}65.3\\
            {\quad}MoE-Adapter \citep{boosting} & -- & 87.9 & {68.2} & 44.1 & 48.1 & 64.7 & \highred{88.8} & \highred{69.0} & \highblue{89.1} & {64.5} & \highblue{65.1} & {\cellcolor{CellCOLOR!50}}68.9\\
            {\quad}RAIL-Primal \citep{rail} & -- & {88.4}&{68.2}& 44.6 &\highred{54.9}&\highred{71.0}& \highblue{88.5} & 59.6 & {89.0} &\highblue{64.7}&\highred{65.2}& {\cellcolor{CellCOLOR!50}}{69.4}\\
             {\quad}{CoDyRA \citep{codyra}} & -- & \highred{92.4} & \highblue{68.4} & \highred{45.8} & \highblue{54.5} &\highblue{69.6} & 87.4 &\highblue{65.2}& 88.5&{64.2} & {64.5} & {\cellcolor{CellCOLOR!50}}\highred{69.9}\\
            \midrule \rowcolor[RGB]{253, 253, 255} {\quad}{\ours} & -- & \highblue{92.0} & \highred{68.8} & \highblue{45.6} & 53.1 & 68.6 & 84.4 & 64.3 & \highred{89.8} & \highred{65.4} & 64.8 & {\cellcolor{CellCOLOR!50}}\highblue{69.7}\\
            \midrule
            \rowcolor{RowCOLOR}\multicolumn{13}{l}{\emph{Average}}\\
            {\quad}LwF \citep{li2017learning} &23.5 &77.4  &43.5 &41.7 &43.5  &52.2 &54.6 & 63.4 & 68.0& 21.3& 52.6&\cellcolor{CellCOLOR!50}49.2\\
            {\quad}LwF-VR \citep{ding2022don} &24.9 & {89.1} &64.2 &53.4 &54.3  &70.8 &79.2 &66.5  &79.2 & 44.1& 61.6&\cellcolor{CellCOLOR!50}62.5\\
            {\quad}WiSE-FT \citep{wortsman2022robust} & {32.0} & 87.7 & 61.0 &{55.8} & {68.1} & 69.3&76.8 &{71.5}  &77.6 &42.0 &59.3&\cellcolor{CellCOLOR!50}63.7  \\
            {\quad}ZSCL \citep{zscl} & 28.2& 88.6 &{66.5} & 53.5&56.3  &{73.4} &{83.1} & 56.4 & {82.4} & {57.5}&{62.9}&\cellcolor{CellCOLOR!50}{64.4} \\
            {\quad}MoE-Adapter \citep{boosting}& {30.0} & {89.6} & \highblue{73.9}& {58.7}& {69.3} &{79.3} & \highblue{88.1}& \highred{76.5} & {89.1}& {65.3}&\highred{65.8} & \cellcolor{CellCOLOR!50}{71.4} \\
            {\quad}RAIL-Primal \citep{rail}& 32.9 &{94.5}	&69.9 &58.1& \highblue{71.8} &\highred{84.4} &\highred{88.5} &70.4 &89.0 &\highblue{66.1} & \highblue{65.7} & {\cellcolor{CellCOLOR!50}}71.9\\
            \rowcolor[RGB]{253, 253, 255} {\quad}{CoDyRA \citep{codyra}} & \highblue{34.6} & \highred{95.8} & \highblue{73.9} & \highblue{60.0} & \highred{77.1} & 81.3 & 86.6 & 75.9 & \highblue{89.9} & \highblue{66.1} & 65.3 & {\cellcolor{CellCOLOR!50}}\highblue{73.3}\\
             \midrule \rowcolor[RGB]{253, 253, 255} {\quad}{\ours} & \highred{36.7} & \highblue{95.4} & \highred{74.9} & \highred{61.9} & \highred{77.1} & \highblue{82.6} & 85.3 & \highblue{76.0} & \highred{90.5} & \highred{67.0} & 65.6 & {\cellcolor{CellCOLOR!50}}\highred{73.9}\\
            \midrule
            \rowcolor{RowCOLOR}\multicolumn{13}{l}{\emph{Last}}\\
            {\quad}LwF \citep{li2017learning} & 22.1 & 58.2 & 17.9 & 32.1 & 28.1 & 66.7 & 46.0 & 84.3 & 64.1 & 31.5 & 60.1 & \cellcolor{CellCOLOR!50}46.5 \\
            {\quad}LwF-VR \citep{ding2022don} & 22.9 & {89.8} & 59.3 & 57.1 & 57.6 & 79.2 & 78.3 & 77.7 & 83.6 & 60.1 & 69.8 & \cellcolor{CellCOLOR!50}66.9  \\
            {\quad}WiSE-FT \citep{wortsman2022robust} & {30.8} & 88.9 & 59.6 & {60.3} & {80.9} & 81.7 & 77.1 & \highblue{94.9} & 83.2 & 62.8 & 70.0 & \cellcolor{CellCOLOR!50}{71.9}  \\
            {\quad}ZSCL \citep{zscl} &26.8 & 88.5 & {63.7} & 55.7 & 60.2 & {82.1} & {82.6} & 58.6 & {85.9} & {66.7} & {70.4} & \cellcolor{CellCOLOR!50}67.4 \\
         {\quad}MoE-Adapter \citep{boosting} & {30.1} & {89.3} &\highblue{74.9}  & \highblue{64.0} & {82.3} &  {89.4}&\highblue{87.1} & 89.0 & {89.1} &{69.5} &{72.5}  & \cellcolor{CellCOLOR!50}{76.1} \\ 
            {\quad}RAIL-Primal \citep{rail} & \highred{32.9}	&{95.1}	&70.3	&63.2	&81.5	&\highred{95.6}	&\highred{88.5}&	89.7	&89.0	&72.5	&71.0 & {\cellcolor{CellCOLOR!50}}77.2\\
            {\quad}{CoDyRA \citep{codyra}} & 31.6 & \highred{95.5} & 72.8 & 63.5 & \highblue{85.0} & 89.7 & 85.0 & {94.7} & \highred{93.2} & \highred{73.6} &\highblue{ 73.0} & {\cellcolor{CellCOLOR!50}}\highblue{78.0}\\
             \midrule \rowcolor[RGB]{253, 253, 255} {\quad}{\ours} & \highblue{32.5} & \highblue{95.3} & \highred{75.3} & \highred{66.6} & \highred{87.8} & \highblue{92.6} & 86.3 & \highred{96.3} & \highblue{92.6} & \highblue{73.5} & \highred{73.8} & {\cellcolor{CellCOLOR!50}}\highred{79.3}\\
		\bottomrule
	\end{tabular}}
\end{table*}

\subsubsection{Additional X-TAIL Results (Order 2)}
\label{supp:xtail_order2}

To further validate \ours's robustness, we evaluated it under other continual-learning task orderings, i.e., X-TAIL (Order 2), as shown in Table~\ref{tab:few_xtail_order_2}. The results align with those in Table~\ref{tab:few_xtail}, confirming that \ours consistently achieves state-of-the-art performance.

\begin{table*}[t]
\centering
\caption{Comparisons on X-TAIL (Order 2) for each domain in terms of ``Transfer'', ``Average'', and ``Last'' scores (\%).
   }
   \label{tab:few_xtail_order_2}
    \resizebox{0.95\linewidth}{!}{
	\begin{tabular}{l>{\centering\arraybackslash}p{1cm} >{\centering\arraybackslash}p{1cm}>{\centering\arraybackslash}p{1cm} >{\centering\arraybackslash}p{1cm} >{\centering\arraybackslash}p{1cm} >{\centering\arraybackslash}p{1cm} >{\centering\arraybackslash}p{1cm} >{\centering\arraybackslash}p{1cm} >{\centering\arraybackslash}p{1cm} >{\centering\arraybackslash}p{1cm} >{\centering\arraybackslash}p{1.6cm}}
 
		\toprule
               {\quad} \makecell[c]{Method} & \makecell[c]{\rotatebox{90}{Cars}} & \makecell[c]{\rotatebox{90}{Aircraft}}  & \makecell[c]{\rotatebox{90}{OxfordPet}} & \makecell[c]{\rotatebox{90}{Food}} & \makecell[c]{\rotatebox{90}{SUN397}} & \makecell[c]{\rotatebox{90}{MNIST}} & \makecell[c]{\rotatebox{90}{Flowers}} & \makecell[c]{\rotatebox{90}{DTD}} & \makecell[c]{\rotatebox{90}{Caltech101}} & \makecell[c]{\rotatebox{90}{EuroSAT}} & \makecell[c]{\textit{Average}} \\
  
		\midrule
            \rowcolor{RowCOLOR}\multicolumn{12}{l}{\emph{CLIP}}\\
		{\quad}Zero-shot & 66.1	&23.5&	86.7&	84&	63.7&	46.7&	63.6	&37.3	&76.8&	36.7 & {\cellcolor{CellCOLOR!50}}58.5\\
            
            \midrule
            \rowcolor{RowCOLOR}\multicolumn{12}{l}{\emph{Transfer}}\\
            {\quad}Zero-shot \citep{clip} &--	&\highblue{23.5}	&86.7	&84	&\highred{63.7}	&46.7	&63.6	&37.3	&76.8	&36.7	&{\cellcolor{CellCOLOR!50}}57.7\\
            
            {\quad}LwF \citep{li2017learning}&--	&20.0	&74.1	&79.6	&58.1	&34.1	&48.9	&27.7	&64.4	&15.1	&{\cellcolor{CellCOLOR!50}}46.9\\
            {\quad}WiSE-FT \citep{wortsman2022robust} &--	&21.3	&79.5	&83.3	&61.0	&39.9	&56.5	&29.6	&68.0	&20.8	&{\cellcolor{CellCOLOR!50}}51.1\\
            {\quad}ZSCL \citep{zscl} &--	&23.0	&84.3	&\highblue{87.2}	&\highblue{63.0}	&42.1	&65.2	&34.6	&71.4	&\highred{40.9}	&{\cellcolor{CellCOLOR!50}}56.9\\
            {\quad}MoE-Adapter \citep{boosting} &--	&17.1	&87.2	&\highred{87.5}	&58.4	&12.6	&65.5	&35.9	&70.0	&17.9	&{\cellcolor{CellCOLOR!50}}50.2\\
            {\quad}RAIL-Primal \citep{rail}&--	&\highblue{23.5}	&86.7	&84	&\highred{63.7}	&46.7	&63.6	&37.3	&76.8	&36.7	&{\cellcolor{CellCOLOR!50}}57.7\\
            {\quad}{CoDyRA \citep{codyra}} &--	&\highred{23.6}	&\highred{89.2}	&83	&62	&\highblue{51}	&\highred{71.4}	&\highblue{38}	&\highred{77.4}	&\highblue{39}	&{\cellcolor{CellCOLOR!50}}\highblue{59.4}\\
           \midrule
            \rowcolor[RGB]{253, 253, 255} {\quad}{\ours} &--	&\highred{23.6}	&\highblue{88.7}	&83.4	&62.6	&\highred{51.2}	&\highblue{69.9}	&\highred{39.3}	&\highblue{77.5}	&\highred{39}	&{\cellcolor{CellCOLOR!50}}\highred{59.5} \\
           
           \midrule
            \rowcolor{RowCOLOR}\multicolumn{12}{l}{\emph{Average}}\\
            {\quad}LwF \citep{li2017learning}&49.0	&27.4	&69.7	&83.0	&65.7	&42.2	&63.5	&33.1	&68.5	&17.5	&{\cellcolor{CellCOLOR!50}}52.0\\
            {\quad}WiSE-FT \citep{wortsman2022robust} &57.9	&29.6	&77.8	&85.4	&68.0	&51.6	&69.3	&35.5	&71.0	&23.0	&{\cellcolor{CellCOLOR!50}}56.9\\
            {\quad}ZSCL \citep{zscl} &74.4	&36.4	&86.7	&\highred{88.7}	&68.9	&50.0	&75.1	&40.1	&72.5	&43.7	&{\cellcolor{CellCOLOR!50}}63.6\\
            {\quad}MoE-Adapter \citep{boosting}&74.4	&38.6	&\highblue{87.7}	&\highblue{87.3}	&67.9	&50.6	&76.5	&43.7	&72.3	&18.8	&{\cellcolor{CellCOLOR!50}}61.8\\
            {\quad}RAIL-Primal \citep{rail} &77.9	&\highred{40.4}	&85.6	&83.3	&68.3	&62.2	&76.6	&45.8	&\highred{80.4}	&41.7	&{\cellcolor{CellCOLOR!50}}66.2\\
            % \midrule
             {\quad}{CoDyRA \citep{codyra}} &\highblue{80}	&39.2	&\highred{92.5}	&85.2	&\highblue{69.2}	&\highblue{73.7}	&\highblue{79.6}	&\highblue{46.2}	&\highblue{78.6}	&\highblue{44.1}	&{\cellcolor{CellCOLOR!50}}\highblue{68.8}\\
              \midrule
              \rowcolor[RGB]{253, 253, 255} {\quad}{\ours} &\highred{80.2}	&\highblue{40.1}	&\highred{92.5}	&84.7	&\highred{70.1}	&\highred{74}	&\highred{80.1}	&\highred{48.7}	&78.4	&\highred{44.4}	&{\cellcolor{CellCOLOR!50}}\highred{69.3} \\

             \midrule
            \rowcolor{RowCOLOR}\multicolumn{12}{l}{\emph{Last}}\\
            {\quad}LwF \citep{li2017learning}	&29.6	&17.5	&63.0	&83.8	&67.7	&44.9	&79.3	&44.8	&\highblue{84.6}	&39.0	&{\cellcolor{CellCOLOR!50}}55.4\\
            {\quad}WiSE-FT \citep{wortsman2022robust} &46.1	&23.5	&71.3	&85.7	&70.2	&59.1	&85.5	&47.9	&82.4	&42.8	&{\cellcolor{CellCOLOR!50}}61.5\\
            {\quad}ZSCL \citep{zscl} &71.7	&35.3	&86.5	&\highred{89.2}	&71.8	&52.3	&89.8	&52.0	&77.1	&68.4	&{\cellcolor{CellCOLOR!50}}69.4\\
            {\quad}MoE-Adapter  \citep{boosting} &75.1	&\highblue{41.1}	&87.9	&\highblue{87.1}	&74.1	&89.7	&92.6	&61.2	&81.0	&27.4	&{\cellcolor{CellCOLOR!50}}71.7\\
            {\quad}RAIL-Primal \citep{rail} &77.7	&\highred{41.9}	&86.1	&83.3	&71.8	&91.6	&\highred{97.3}	&\highblue{66.4}	&\highred{94.8}	&86.9	&{\cellcolor{CellCOLOR!50}}\highblue{79.8}\\
            {\quad}{CoDyRA \citep{codyra}} &\highblue{79}	&38.6	&\highblue{92.6}	&86.4	&\highblue{74.7}	&\highblue{95.2}	&93	&64.7	&81.9	&\highblue{92.2}	&{\cellcolor{CellCOLOR!50}}\highblue{79.8} \\
            \midrule
            \rowcolor[RGB]{253, 253, 255} {\quad}{\ours} &\highred{79.3}	&38.9	&\highred{93.1}	&85.4	&\highred{74.9}	&\highred{96.4}	&\highblue{94.1}	&\highred{69.9}	&82	&\highred{92.9}	&{\cellcolor{CellCOLOR!50}}\highred{80.7} \\

             \bottomrule
	\end{tabular}}
\end{table*}

\subsection{Continual Learning of Language Models}
\label{supp:lm_results}

\subsubsection{Standard CL Benchmark (T5-Large)}
\label{supp:standard_cl}

Table \ref{tab:llm_cl_main} reports results across three task orderings: \ours consistently outperforms prior methods and closely approaches the multi-task learning (MTL) upper bound. Unlike O-LoRA \citep{wang2023orthogonal} and LB-CL \citep{qiao2024learn}, which rely on orthogonality constraints or gradient projections between per-task LoRA adapters (potentially limiting adapter capacity), \ours needs no extra regularization. By decomposing each rank-r update into rank-one components and applying self-activated, sparse gating, \ours lets each component specialize on its own input distribution, reducing interference and more effectively capturing diverse patterns. We further evaluate \ours on LLaMA2-7B \citep{touvron2023llama} under the same continual-learning setup (Table~\ref{tab:llama}), \ours also outperforms O-LoRA by 2.3\% averaged over 3 task orders.

\begin{table}[H]
% \begin{wraptable}{r}{0.4\linewidth}
\centering
\small
\vspace{-0.1cm}
\caption{Summary of results on standard CL benchmarks with T5-large. We report averaged accuracy after training on the last task across three task orderings.}
\label{tab:llm_cl_main}
\vspace{-0.2cm}
\resizebox{0.9\linewidth}{!}{
\begin{tabular}{lcccc}
\toprule
\textbf{} & \multicolumn{4}{c}{{Standard CL Benchmark}} \\
\cmidrule(lr){2-5}
{Method} & {Order-1} & {Order-2} & {Order-3} & \textit{Avg.} \\
\midrule
\rowcolor{CellCOLOR}
MTL        & \multicolumn{4}{c}{{80.0}} \\
% \midrule
SeqFT   & 18.9 & 24.9 & 41.7 & 28.5 \\
SeqLoRA & 44.6 & 32.7 & 53.7 & 43.7 \\
IncLoRA & 66.0 & 64.9 & 68.3 & 66.4 \\
Replay  & 55.2 & 56.9 & 61.3 & 57.8 \\
EWC     & 48.7 & 47.7 & 54.5 & 50.3 \\
LwF     & 54.4 & 53.1 & 49.6 & 52.3 \\
L2P     & 60.3 & 61.7 & 61.1 & 60.7 \\
LFPT5   & 67.6 & 72.6 & 77.9 & 72.7 \\
InfLoRA	& 75.2 & 75.4 & 75.8 & 75.5 \\
{O-LoRA} & {75.4} & {75.7} & 76.3 & {75.8} \\
LB-CL & 76.9 & 76.5 & 76.8 & 76.7 \\
\midrule
\rowcolor{RowCOLOR}
\ours & \textbf{77.4} & \textbf{77.5} & \textbf{77.9} & \textbf{77.6} \\
\bottomrule
\end{tabular}
}
% \vspace{-0.9cm}
% \end{wraptable}
\end{table}

\begin{table}[H]
    \centering
% \begin{wraptable}{r}{0.34\linewidth}
\centering
\small
% \vspace{-0.2cm}
\caption{Continual learning results on standard CL benchmarks with the LLaMA2-7B model.}
\label{tab:llama}
% \vspace{-0.2cm}
% \resizebox{\linewidth}{!}{
\begin{tabular}{@{}lcccc}
\toprule
 Method& Order-1 & Order-2 & Order-3 & \textit{Avg.} \\ \midrule
O-LoRA & 76.8 & 75.7 & 75.7 & 76.1 \\
\rowcolor{RowCOLOR} \ours & \textbf{77.8} & \textbf{78.0} & \textbf{79.3} & \textbf{78.4}\\ \bottomrule
\end{tabular}
% \vspace{-1.6cm}
% }
% \vspace{-0.2cm}
% \end{wraptable}
\end{table}

\subsubsection{Long Task Sequences (15-task)}
\label{supp:long_seq}

In Table~\ref{tab:supp_llm_cl_long}, we extend the evaluation to challenging long task sequences using 15 datasets with 3 different orderings as in \citep{wang2023orthogonal,qiao2024learn}. Consistent with the findings in Table~\ref{tab:llm_cl_main}, \ours outperforms previous methods in terms of averaged performance across three task orders, and largely closes the gap to multi-task learning. Two key design choices drive this robust performance in the long-sequence regime. First, by decomposing each LoRA update into fine-grained rank-1 atoms and enforcing a small, fixed activation budget, \ours encourages each atom to specialize on a narrow subspace of the data manifold. At inference time, only the most relevant subspaces are activated for a given input, which preserves earlier task representations and prevents catastrophic interference. Second, our self-activated gating mechanism enables each atom to assess its own relevance on a per-token basis, yielding stable mixture patterns as the memory bank grows. Coupling with our proposed atom pruning, these mechanisms ensure that \ours continually incorporates new knowledge only when needed while robustly maintaining prior capabilities.

\begin{table}[H]
    \centering
\small
\caption{Average accuracy on T5-large continual‐learning benchmarks after the final task, evaluated over extended 15-task sequences. Results for prior methods are taken from \citep{qiao2024learn}.}\label{tab:supp_llm_cl_long}
\begin{tabular}{lcccc}
\toprule
\textbf{} & \multicolumn{4}{c}{{Large Number of Tasks}} \\
\cmidrule(lr){2-5}
{Method} & {Order-4} & {Order-5} & {Order-6} & \textit{Avg.} \\
\midrule
\rowcolor{CellCOLOR}
MTL        & \multicolumn{4}{c}{{76.5}} \\
\midrule
SeqFT   & 7.4   & 7.4   & 7.5   & 7.4 \\
SeqLoRA & 2.3   & 0.6   & 1.9   & 1.6 \\
IncLoRA & 63.3  & 58.5  & 61.7  & 61.2 \\
Replay  & {55}   & 54.6  & 53.1  & 54.2 \\
EWC     & 45.3  & 44.5  & 45.6  & 45.1 \\
LwF     & 50.1  & 43.1  & 47.4  & 46.9 \\
L2P     & 57.5  & 53.8  & 56.9  & 56.1 \\
LFPT5 & 69.8 & 67.2 & 69.2 & 68.7 \\
O-LoRA & \textbf{70.5} & 65.5 & 70.5 & 68.8 \\
LB-CL & 68.4 & 67.3 & 71.8 & 69.2 \\ 
\midrule
\rowcolor{RowCOLOR}
\ours & 68.91 & \textbf{68.32} & \textbf{71.95} & \textbf{69.72}\\
\bottomrule
\end{tabular}
\end{table}

\subsubsection{Results on SuperNI Benchmark}
\label{supp:superni}

To provide a complete comparison, we additionally evaluate \ours against SAPT~\citep{zhao2024sapt} on the SuperNI benchmark~\citep{wang2022super} using the T5-Large backbone (Table~\ref{tab:sapt}). We report results in both replay-based and replay-free settings. Under the replay-based protocol introduced by SAPT, where pseudo-samples are generated using a trained generative model, \ours attains a stronger stability--plasticity trade-off (AP: 51.79\% vs. 51.54\%; FT: 0.73\% vs. 0.91\%). In the replay-free setting, which is the primary use case for \ours, our method achieves state-of-the-art performance (AP: 39.62\%) and substantially outperforms parameter-efficient baselines such as O-LoRA (26.37\%) and L2P (15.18\%). These results highlight the complementary design choices. SAPT improves replay efficiency through shared attention prompts, while \ours relies on architectural isolation via sparse rank-1 experts. As a result, \ours remains effective without a memory buffer, yet can also incorporate replay to further improve performance beyond SAPT.

\begin{table}[H]
\centering
\caption{{Overall results on the SuperNI Benchmark using the T5-Large backbone. We report Average Performance (AP, $\uparrow$) and Forgetting (FT, $\downarrow$). The best results for the stability-plasticity trade-off are highlighted in bold for methods with and without memory replay, respectively.}}
\label{tab:sapt}
\small
\begin{tabular}{@{}lccc@{}}
\toprule
\multirow{2}{*}{Methods} & \multirow{2}{*}{Replay} & \multicolumn{2}{c}{SuperNI} \\ \cmidrule(l){3-4} 
 &  & AP & FT \\ \midrule
\rowcolor{CellCOLOR} \multicolumn{4}{l}{\textit{Replay-Based Methods}} \\ \midrule
Replay & \ding{51} & 35.37 & 16.92 \\
SAPT & \ding{51} & 51.54 & 0.91 \\ \midrule
\rowcolor{RowCOLOR} \ours & \ding{51} & \textbf{51.79} & \textbf{0.73} \\ \midrule
\rowcolor{CellCOLOR}\multicolumn{4}{l}{Replay-Free Methods} \\ \midrule
L2P & \ding{55} & 15.18 & 3.65 \\
IncLoRA & \ding{55} & 12.33 & 41.93 \\
C-LoRA & \ding{55} & 22.69 & 24.25 \\
O-LoRA & \ding{55} & 26.37 & 19.15 \\ \midrule
\rowcolor{RowCOLOR}\ours & \ding{55} & \textbf{39.62} & \textbf{5.74} \\ \bottomrule
\end{tabular}
\end{table}

\subsection{More Details of Table~\ref{tab:llm_ood_main}}
\label{supp:humaneval}

To further evaluate code generation performance, we compare \ours against LoRI-D and LoRI-S~\citep{zhang2025lori} on the HumanEval benchmark (Table~\ref{tab:supp_humaneval}). \ours consistently outperforms both LoRI variants across all metrics, achieving notable improvements in Pass@1, Pass@5, and especially Pass@10.

\begin{table}[H]
\caption{Performance comparison on the HumanEval benchmark, reported in terms of Pass@1, Pass@5, and Pass@10.}

\label{tab:supp_humaneval}
\centering
\small
\begin{tabular}{lccc}
\toprule
HumanEval & Pass@1 & Pass@5 & Pass@10 \\
\midrule
LoRI-D & 43.2  & 57.6  & 63.2  \\
LoRI-S & 41.3  & 54.4  & 59.6  \\
\rowcolor{RowCOLOR} \ours   & \textbf{47.6} & \textbf{60.9} & \textbf{70.1} \\
\bottomrule
\end{tabular}

\end{table}

\subsection{Comparisons with Additional Baselines}
\label{supp:additional_baselines}

We compare \ours against two recent PEFT-based continual learning methods: SD-LoRA~\citep{wu2025sd}, which targets class-incremental learning via scalable decoupled adaptation, and NoRGa~\citep{le2024mixture}, which combines mixture-of-experts with prompt-based continual learning. Table~\ref{tab:additional_baselines} reports results on X-TAIL under the same protocol as Table~\ref{tab:few_xtail}. \ours outperforms both methods across all three metrics, demonstrating stronger retention of both new and old task knowledge.

\begin{table*}[!t]
\centering
\caption{Comparisons with additional baselines on X-TAIL, following the same protocol as Table~\ref{tab:few_xtail}.}
\label{tab:additional_baselines}
\resizebox{0.95\linewidth}{!}{
\begin{tabular}{l>{\centering\arraybackslash}p{1cm} >{\centering\arraybackslash}p{1cm}>{\centering\arraybackslash}p{1cm} >{\centering\arraybackslash}p{1cm} >{\centering\arraybackslash}p{1cm} >{\centering\arraybackslash}p{1cm} >{\centering\arraybackslash}p{1cm} >{\centering\arraybackslash}p{1cm} >{\centering\arraybackslash}p{1cm} >{\centering\arraybackslash}p{1cm} >{\centering\arraybackslash}p{1.6cm}}
\toprule
{\quad} \makecell[c]{Method} & \makecell[c]{\rotatebox{90}{Aircraft}} & \makecell[c]{\rotatebox{90}{Caltech}}  & \makecell[c]{\rotatebox{90}{DTD}} & \makecell[c]{\rotatebox{90}{EuroSAT}} & \makecell[c]{\rotatebox{90}{Flowers}} & \makecell[c]{\rotatebox{90}{Food}} & \makecell[c]{\rotatebox{90}{MNIST}} & \makecell[c]{\rotatebox{90}{OxPet}} & \makecell[c]{\rotatebox{90}{Cars}} & \makecell[c]{\rotatebox{90}{SUN397}} & \makecell[c]{\textit{Average}} \\
\midrule
\rowcolor{RowCOLOR}\multicolumn{12}{l}{\emph{Transfer}}\\
{\quad}NoRGa \citep{le2024mixture} & -- & 73.1 & 32.7 & 43.6 & 64.7 & 82.3 & 42.4 & 87.1 & 64.9 & 60.4 & {\cellcolor{CellCOLOR!50}}61.2 \\
{\quad}SD-LoRA \citep{wu2025sd} & -- & 73.7 & 38.4 & 39.7 & 64.7 & 80.6 & 48.3 & 87.6 & 55.0 & 59.2 & {\cellcolor{CellCOLOR!50}}60.8 \\
\midrule
\rowcolor[RGB]{253, 253, 255} {\quad}{\ours} & -- & 74.5 & 38.1 & 46.9 & 65.3 & 82.9 & 45.8 & 88.2 & 65.1 & 62.9 & {\cellcolor{CellCOLOR!50}}\textbf{63.3} \\
\midrule
\rowcolor{RowCOLOR}\multicolumn{12}{l}{\emph{Average}}\\
{\quad}NoRGa \citep{le2024mixture} & 35.4 & 81.3 & 46.8 & 63.5 & 67.4 & 80.2 & 56.4 & 84.2 & 50.2 & 56.4 & {\cellcolor{CellCOLOR!50}}62.2 \\
{\quad}SD-LoRA \citep{wu2025sd} & 20.8 & 86.8 & 56.4 & 75.6 & 81.8 & 81.9 & 67.2 & 89.0 & 58.3 & 60.8 & {\cellcolor{CellCOLOR!50}}67.9 \\
\midrule
\rowcolor[RGB]{253, 253, 255} {\quad}{\ours} & 44.1 & 81.6 & 64.6 & 79.6 & 83.9 & 84.4 & 66.5 & 89.7 & 68.4 & 64.1 & {\cellcolor{CellCOLOR!50}}\textbf{72.7} \\
\midrule
\rowcolor{RowCOLOR}\multicolumn{12}{l}{\emph{Last}}\\
{\quad}NoRGa \citep{le2024mixture} & 26.6 & 78.5 & 43.2 & 76.7 & 73.9 & 82.7 & 82.8 & 88.2 & 65.0 & 72.6 & {\cellcolor{CellCOLOR!50}}69.0 \\
{\quad}SD-LoRA \citep{wu2025sd} & 19.7 & 79.9 & 59.9 & 86.2 & 87.9 & 82.6 & 94.9 & 91.9 & 64.8 & 74.8 & {\cellcolor{CellCOLOR!50}}74.3 \\
\midrule
\rowcolor[RGB]{253, 253, 255} {\quad}{\ours} & 37.7 & 81.5 & 70.7 & 92.4 & 95.0 & 86.0 & 97.6 & 92.6 & 81.0 & 74.7 & {\cellcolor{CellCOLOR!50}}\textbf{80.9} \\
\bottomrule
\end{tabular}}
\end{table*}
% TODO: INSERT brief analysis text

%=============================================================
\section{Analysis and Ablations}
\label{supp:analysis}
%=============================================================

\subsection{Robustness and Variance Analysis}
\label{supp:robustness}

\ours employs a sparse mixture of previously learned and newly introduced rank-1 experts to capture both shared and task-specific knowledge, resulting in substantially improved Last performance. To assess statistical significance and robustness, we report mean and standard deviation over three independent runs (Table~\ref{tab:supp_robustness}). \ours consistently outperforms competing methods across all metrics and exhibits lower variance, highlighting its effectiveness and stability in continual-learning scenarios.

\begin{table*}[h]
\centering
\caption{Comparison to InfLoRA and performance robustness. We report mean and standard deviation across 3 independent runs. Best performances are marked in \textbf{bold}.
   }
   \label{tab:supp_robustness}
   \small
    \resizebox{0.99\linewidth}{!}{
	\begin{tabular}{l>{\centering\arraybackslash}p{1.35cm} >{\centering\arraybackslash}p{1.35cm}>{\centering\arraybackslash}p{1.35cm} >{\centering\arraybackslash}p{1.35cm} >{\centering\arraybackslash}p{1.35cm} >{\centering\arraybackslash}p{1.35cm} >{\centering\arraybackslash}p{1.35cm} >{\centering\arraybackslash}p{1.35cm} >{\centering\arraybackslash}p{1.35cm} >{\centering\arraybackslash}p{1.35cm} >{\centering\arraybackslash}p{1.6cm}}
 
		\toprule
               {\quad} \makecell[c]{Method} & \makecell[c]{\rotatebox{90}{Cars}} & \makecell[c]{\rotatebox{90}{Aircraft}}  & \makecell[c]{\rotatebox{90}{OxfordPet}} & \makecell[c]{\rotatebox{90}{Food}} & \makecell[c]{\rotatebox{90}{SUN397}} & \makecell[c]{\rotatebox{90}{MNIST}} & \makecell[c]{\rotatebox{90}{Flowers}} & \makecell[c]{\rotatebox{90}{DTD}} & \makecell[c]{\rotatebox{90}{Caltech101}} & \makecell[c]{\rotatebox{90}{EuroSAT}} & \makecell[c]{\textit{Average}} \\

            \midrule
            \rowcolor{RowCOLOR}\multicolumn{12}{l}{\emph{Transfer}}\\
            
            {\quad}InfLoRA  & -- & $72.26^{\pm 0.56}$ & $36.19^{\pm 0.64}$ & $38.46^{\pm 0.39}$ & $55.22^{\pm 1.65}$ & $73.19^{\pm 0.55}$ & $39.32^{\pm 1.54}$ & $80.29^{\pm 0.91}$ & $51.19^{\pm 1.16}$ & $55.05^{\pm 0.51}$ & {\cellcolor{CellCOLOR!50}}$55.69^{\pm 0.24}$ \\
            
            {\quad}CoDyRA  & -- & $74.3^{\pm 0.52}$ & $36.8^{\pm 0.23}$ & $44.2^{\pm 0.56}$ & $\mathbf{69.9^{\pm 0.56}}$ & $\mathbf{83.5^{\pm 0.23}}$ & $42.8^{\pm 0.18}$ & $\mathbf{88.9^{\pm 0.42}}$ & $64.6^{\pm 0.47}$ & $\mathbf{63.4^{\pm 0.56}}$ & {\cellcolor{CellCOLOR!50}}$63.2^{\pm 0.28}$\\
           \midrule
            \rowcolor[RGB]{253, 253, 255} {\quad}{\ours} & -- & $\mathbf{74.5^{\pm 0.51}}$ & $\mathbf{38.1^{\pm 0.24}}$ & $\mathbf{46.9^{\pm 0.56}}$ & $65.3^{\pm 0.44}$ & $82.9^{\pm 0.18}$ & $\mathbf{45.8^{\pm 0.31}}$ & $88.2^{\pm 0.15}$ & $\mathbf{65.1^{\pm 0.35}}$ & $62.9^{\pm 0.10}$ & {\cellcolor{CellCOLOR!50}}$\mathbf{63.3^{\pm 0.26}}$ \\
           
           \midrule
            \rowcolor{RowCOLOR}\multicolumn{12}{l}{\emph{Average}}\\
            {\quad}InfLoRA & $20.49^{\pm 0.98}$ & $78.58^{\pm 1.02}$ & $48.5^{\pm 1.18}$ & $66.59^{\pm 1.51}$ & $71.83^{\pm 0.80}$ & $76.79^{\pm 0.34}$ & $61.45^{\pm 1.36}$ & $82.59^{\pm 0.86}$ & $55.3^{\pm 1.34}$ & $56.67^{\pm 0.59}$ & {\cellcolor{CellCOLOR!50}}$62.48^{\pm 0.31}$\\
            
             {\quad}CoDyRA  & $41.4^{\pm 0.28}$ & $81^{\pm 0.38}$ & $58.7^{\pm 0.26}$ & $77.8^{\pm 0.47}$ & $83.4^{\pm 0.39}$ & $\mathbf{84.6^{\pm 0.28}}$ & $64.5^{\pm 0.14}$ & $\mathbf{90.4^{\pm 0.40}}$ & $67.2^{\pm 0.23}$ & $\mathbf{64.4^{\pm 0.47}}$ & {\cellcolor{CellCOLOR!50}}$71.3^{\pm 0.18}$\\
              \midrule
              \rowcolor[RGB]{253, 253, 255} {\quad}{\ours} & $\mathbf{44.1^{\pm 0.24}}$ & $\mathbf{81.6^{\pm 0.34}}$ & $\mathbf{64.6^{\pm 0.34}}$ & $\mathbf{79.6^{\pm 0.37}}$ & $\mathbf{83.9^{\pm 0.36}}$ & $84.4^{\pm 0.15}$ & $\mathbf{66.5^{\pm 0.24}}$ & $89.7^{\pm 0.07}$ & $\mathbf{68.4^{\pm 0.38}}$ & $64.1^{\pm 0.09}$ & {\cellcolor{CellCOLOR!50}}$\mathbf{72.7^{\pm 0.17}}$ \\

             \midrule
            \rowcolor{RowCOLOR}\multicolumn{12}{l}{\emph{Last}}\\
            {\quad}InfLoRA 	& $18.26^{\pm 0.49}$ & $\mathbf{82.36^{\pm 0.92}}$ & $46.57^{\pm 0.89}$ & $79.38^{\pm 2.22}$ & $76.16^{\pm 1.61}$ & $79.58^{\pm 0.60}$ & $95.74^{\pm 0.44}$ & $87.78^{\pm 0.85}$ & $71.11^{\pm 0.73}$ & $73.05^{\pm 0.19}$ & {\cellcolor{CellCOLOR!50}}$70.99^{\pm 0.24}$\\
            
            {\quad}CoDyRA  & $\mathbf{37.7^{\pm 0.42}}$ & $81.5^{\pm 0.24}$ & $65.1^{\pm 0.63}$ & $89.9^{\pm 0.55}$ & $91.4^{\pm 0.38}$ & $85.5^{\pm 0.16}$ & $96.8^{\pm 0.08}$ & $\mathbf{93.3^{\pm 0.30}}$ & $77.3^{\pm 0.66}$ & $73.5^{\pm 0.21}$ & {\cellcolor{CellCOLOR!50}}$79.2^{\pm 0.18}$ \\
            \midrule
            \rowcolor[RGB]{253, 253, 255} {\quad}{\ours} & $\mathbf{37.7^{\pm 0.28}}$ & $81.5^{\pm 0.22}$ & $\mathbf{70.7^{\pm 0.49}}$ & $\mathbf{92.4^{\pm 0.20}}$ & $\mathbf{95^{\pm 0.34}}$ & $\mathbf{86^{\pm 0.13}}$ & $\mathbf{97.6^{\pm 0.19}}$ & $92.6^{\pm 0.10}$ & $\mathbf{81^{\pm 0.35}}$ & $\mathbf{74.7^{\pm 0.06}}$ & {\cellcolor{CellCOLOR!50}}$\mathbf{80.9^{\pm 0.12}}$ \\

             \bottomrule
	\end{tabular}
    }
    \vspace{-0.3cm}
\end{table*}

\subsection{Order Robustness}
\label{supp:order_robust}

To assess order robustness, we incorporate the OPD metric proposed by \citep{Yoon2020Scalable}, which measures a model's sensitivity to the sequence of arriving tasks. Following standard practice for evaluating global performance stability, we compute the standard deviation of the final average accuracy over the $K$ task orders considered. In Table~\ref{tab:order_robustness}, using results on the Standard CL Benchmark with three distinct task orders, \ours demonstrates substantially improved robustness to task ordering. The disparity across orders is 0.26 for \ours, which is approximately half of that of O-LoRA (0.46). This indicates that the Self-Activated Sparse Mixture mechanism effectively reduces task interference and mitigates the unidirectional knowledge transfer effects identified in prior work.

\begin{table}[htb]
\centering
\caption{{Order Robustness Analysis on standard CL benchmark with T5-Large. We report the accuracy for each order and the average accuracy $\pm$ the standard deviation.}}
\label{tab:order_robustness}
\small
\begin{tabular}{lcccc}
\toprule
Method & Order 1 & Order 2 & Order 3 & Avg $\pm$ Std \\ \midrule
SeqFT & 18.9 & 24.9 & 41.7 & 28.5 $\pm$ 11.82 \\
L2P & 60.3 & 61.7 & 61.1 & 61.0 $\pm$ 0.70 \\
LFPT5 & 67.6 & 72.6 & 77.9 & 72.7 $\pm$ 5.15 \\
O-LoRA & 75.4 & 75.7 & 76.3 & 75.8 $\pm$ 0.46 \\
\midrule
\rowcolor{RowCOLOR} \ours & \textbf{77.4} & \textbf{77.5} & \textbf{77.9} & \textbf{77.6 $\pm$ 0.26} \\ \bottomrule
\end{tabular}
\end{table}

\subsection{Computation Cost and Parameter Analysis}
\label{supp:compute}

Table~\ref{tab:supp_params} summarizes the per-task trainable parameters of various continual-learning methods. Standard LoRA \citep{lora}, CoDyRA \citep{codyra}, and O-LoRA \citep{wang2023orthogonal} each introduce $r(d_\text{in}+d_\text{out})$ new parameters per weight matrix. Mixture-of-Experts variants such as MoE-LoRA and MoE-Adapter \citep{boosting} additionally train a router module to control the usage of each LoRA expert, inducing $d_\text{in}$ additional parameters for each expert. LB-CL \citep{qiao2024learn} introduces $r$ additional parameters, mimicking the singular values of SVD.

By contrast, \ours requires only $r d_\text{in} + k d_\text{out}$ activated trainable parameters per task, where $k\le r$ is the activation budget, and the trainable parameter count is at most the same as a standard LoRA. Despite the small number of parameters activated and trained, \ours achieves superior continual learning performance, and reaches comparable performance in general fine-tuning with only one-third of the activated parameters of a standard LoRA.

\begin{table}[H]
    \centering
        \caption{Comparisons of trainable parameters for each pre-trained weight matrix during continual learning of each task. MoE-LoRA and MoE-Adapter trains additional router module with LoRA experts. In \ours, $k$ denotes the memory activation budget (with $k \le r$).}
    \label{tab:supp_params}
\small
\resizebox{\linewidth}{!}{
  \begin{tabular}{@{}l c@{}}
    \toprule
    \textbf{Method} 
      & \textbf{Trainable parameters per task} \\
    \midrule
    LoRA \citep{lora} 
      & $r\,(d_\text{in} + d_\text{out})$ \\
    MoE-LoRA (1 expert/task)
      & $r\,(d_\text{in} + d_\text{out})\;+\;d_\text{in}$ \\
    MoE-Adapter (2 experts/task) \citep{boosting} 
      & $2(r\,(d_\text{in} + d_\text{out}) + d_\text{in})$ \\
    CoDyRA \citep{codyra} 
      & $r\,(d_\text{in} + d_\text{out}) + r$ \\
    O-LoRA \citep{wang2023orthogonal} 
      & $r\,(d_\text{in} + d_\text{out})$ \\
      LB-CL \citep{qiao2024learn} & $r\,(d_\text{in} + d_\text{out}) + r$ \\
    \midrule
    \rowcolor{RowCOLOR}
    \ours 
      & $r d_\text{in} + k d_\text{out}$ \\
    \bottomrule
  \end{tabular}
  }
\end{table}

\noindent\textbf{Trainable parameters and training GPU memory.} Beyond the estimated parameter counts in Table~\ref{tab:supp_params}, in Table~\ref{tab:sup_params_memory}, we measured the actual trainable parameters for each continual-learning task and GPU memory usage, under the same settings as Table~\ref{tab:few_xtail} in the main paper.

LWF \citep{li2017learning} and ZSCL \citep{zscl} perform full-parameter fine-tuning, consuming the most parameters and memory. MoE-Adapters \citep{boosting} maintains a router with 22 rank-64 adapter experts (top-2 activated) and a DDAS domain predictor. CoDyRA \citep{codyra} trains a single rank-16 LoRA per task, reducing its footprint to 4.4\,M parameters. \ours introduces 16 rank-1 experts per task, with no additional router, for a total of 4.4\,M trainable parameters and keeps a low GPU memory usage, thanks to our novel self-activated sparse mixture of memories design.

\begin{table}[H]
\centering
\caption{{Trainable parameters and averaged training GPU memory per task.}}
\label{tab:sup_params_memory}
\resizebox{\linewidth}{!}{
\begin{tabular}{lcc}
\toprule
\textbf{Method} & \textbf{Trainable Params. (Million)} & \textbf{GPU Mem. (MiB)} \\
\midrule
LWF \citep{li2017learning}            & 129.6 & 32172 \\
ZSCL \citep{zscl}           & 129.6 & 26290 \\
MoE-Adapters \citep{boosting}   & 59.8  & 22358 \\
CoDyRA \citep{codyra} & 4.4   & 21770 \\
\midrule
\ours           & 4.4   & 21090 \\
\bottomrule
\end{tabular}}
\end{table}

\subsection{Latency Analysis}
\label{supp:latency}

In Table \ref{tab:latency}, we measure per-image processing time (ms) on the X-TAIL benchmark across all 10 tasks to assess the computational overhead of \ours as the memory bank grows.

\begin{table}[H]
\centering
\small
\caption{Latency analysis on X-TAIL (avg ms/image).}
\label{tab:latency}
\begin{subtable}{\linewidth}
\centering
\caption{Training}
\begin{tabular}{ccc}
\toprule
CoDyRA & MoE-Adapters & \ours \\
\midrule
6.99 & 8.56 & 8.32 \\
\bottomrule
\end{tabular}
\end{subtable}
\begin{subtable}{\linewidth}
\centering
\vspace{0.3cm}
\caption{Inference}
\resizebox{\linewidth}{!}{
\begin{tabular}{lcccc}
\toprule
 & CLIP & CoDyRA & MoE-Adapters & \ours \\
\midrule
After Task 1  & 1.95 & 1.95 & 3.25 & 2.07 \\
After Task 5  & 1.95 & 1.95 & 3.25 & 2.27 \\
After Task 10 & 1.95 & 1.95 & 3.25 & 2.52 \\
\bottomrule
\end{tabular}}
\end{subtable}
\end{table}

Training cost is comparable across methods. The variation in training time across datasets is primarily due to data I/O loading (e.g., image resolution, dataset size) rather than differences between methods. For inference, CoDyRA's weights are merged directly into the pre-trained weight matrix, so its cost equals the baseline. \ours's inference latency grows modestly as tasks accumulate, and remains substantially faster than MoE-Adapters throughout. This confirms that while relevance score computation grows with accumulated atoms, the practical overhead is modest due to fixed top-$k$ activation and threshold-based selection.

\subsection{Post-Pruning Analysis}
\label{supp:pruning}

While \ours entails linear storage growth, our atomic structure offers a distinct advantage over conventional LoRA. Unlike standard matrices that require complex post-hoc decomposition (e.g., SVD) to compress, our rank-1 experts are independent and can be individually assessed. Consequently, the model inherently supports post-pruning, retaining only atoms that exceed a cumulative activation threshold, allowing for storage reduction with minimal performance degradation if required.

We probe the effects of pruning in Table \ref{tab:prune}, where we collect the cumulative activation mass of all rank-1 memories during training and retain only the subset required to capture the top 99\% of the total activation mass, which on average prunes approximately 30\% of the parameter storage. We find that the degradation to performance is minimal, confirming that the model naturally learns a sparse representation where information is concentrated in a concise set of high-utility rank-1 memories.

\begin{table*}[t]
\centering
\caption{Analysis of post-pruning. After training of each task, we retain the subset required to capture the top 99\% of the total activation.
   }
   \label{tab:prune}
   \small
    \resizebox{0.99\linewidth}{!}{
	\begin{tabular}{l>{\centering\arraybackslash}p{1.35cm} >{\centering\arraybackslash}p{1.35cm}>{\centering\arraybackslash}p{1.35cm} >{\centering\arraybackslash}p{1.35cm} >{\centering\arraybackslash}p{1.35cm} >{\centering\arraybackslash}p{1.35cm} >{\centering\arraybackslash}p{1.35cm} >{\centering\arraybackslash}p{1.35cm} >{\centering\arraybackslash}p{1.35cm} >{\centering\arraybackslash}p{1.35cm} >{\centering\arraybackslash}p{1.6cm}}
 
		\toprule
               {\quad} \makecell[c]{Method} & \makecell[c]{\rotatebox{90}{Cars}} & \makecell[c]{\rotatebox{90}{Aircraft}}  & \makecell[c]{\rotatebox{90}{OxfordPet}} & \makecell[c]{\rotatebox{90}{Food}} & \makecell[c]{\rotatebox{90}{SUN397}} & \makecell[c]{\rotatebox{90}{MNIST}} & \makecell[c]{\rotatebox{90}{Flowers}} & \makecell[c]{\rotatebox{90}{DTD}} & \makecell[c]{\rotatebox{90}{Caltech101}} & \makecell[c]{\rotatebox{90}{EuroSAT}} & \makecell[c]{\textit{Average}} \\

            \midrule
            \rowcolor{RowCOLOR}\multicolumn{12}{l}{\emph{Transfer}}\\
            
            {\quad}{\ours} & -- & {74.5} & {38.1} & {46.9} & 65.3 & 82.9 & {45.8} & 88.2 & {65.1} & 62.9 & {\cellcolor{CellCOLOR!50}}{63.3} \\
           % \midrule
            {\quad}{\ours \textit{w}/ pruning} & -- & 75.1 & 38.0 & 43.6 & 68.4 & 83.9 & 48.0 & 88.8 & 65.3 & 62.3 & {\cellcolor{CellCOLOR!50}63.7} \\
           \rowcolor{RowCOLOR}\multicolumn{12}{l}{\emph{Average}}\\

           \midrule
              {\quad}{\ours} & {44.1} & {81.6} & {64.6} & {79.6} & {83.9} & 84.4 & {66.5} & 89.7 & {68.4} & 64.1 & {\cellcolor{CellCOLOR!50}}{72.7} \\
               % \midrule
            {\quad}{\ours \textit{w}/ pruning} & 42.3 & 81.0 & 62.9 & 75.4 & 82.7 & 83.2 & 66.3 & 89.6 & 67.6 & 63.0 &  {\cellcolor{CellCOLOR!50} 71.4} \\

            \rowcolor{RowCOLOR}\multicolumn{12}{l}{\emph{Last}}\\

             \midrule
            {\quad}{\ours} & {37.7} & 81.5 & {70.7} & {92.4} & {95.0} & {86.0} & {97.6} & 92.6 & {81.0} & {74.7} & {\cellcolor{CellCOLOR!50}}{80.9} \\
                        % \midrule
            {\quad}{\ours \textit{w}/ pruning} & 34.6 & 80.8 & 67.9 & 89.4 & 94.2 & 85.3 & 97.3 & 92.6 & 79.8 & 73.7 & {\cellcolor{CellCOLOR!50} 79.5} \\

             \bottomrule
	\end{tabular}
    }
\end{table*}
\subsection{Load-Balancing Ablation}
\label{supp:load_balance}

In standard MoE, load-balancing regularization prevents a shared external router from developing degenerate preferences. \ours mitigates this concern structurally: each atom's relevance is computed from its own frozen key rather than a shared trainable router, reducing the need for explicit load-balancing. We empirically verify that adding load-balancing regularization \emph{degrades} performance across two settings (Table~\ref{tab:load_balance}).

\begin{table}[H]
\centering
\small
\caption{Effect of load-balancing regularization.}
\label{tab:load_balance}
\resizebox{\linewidth}{!}{
\begin{tabular}{lcccc}
\toprule
 & X-TAIL & \multicolumn{3}{c}{T5-Large: 15-task} \\
\cmidrule(lr){2-2} \cmidrule(lr){3-5}
 & Last & Order-4 & Order-5 & Order-6 \\
\midrule
\ours & \textbf{80.9} & \textbf{68.91} & \textbf{68.32} & \textbf{71.95} \\
\ours w/ load-bal. & 77.3 & 68.23 & 67.86 & 71.46 \\
\bottomrule
\end{tabular}}
\end{table}

This degradation is consistent across both CLIP (10 tasks) and T5 (15 tasks) settings. Semantic information is inherently unevenly distributed in the data, so forcing each memory unit to activate equally works against the natural data distribution and hinders specialization of the key vectors, which serve dual roles in both routing and representation.

%=============================================================
\section{Visualizations}
\label{supp:vis}
%=============================================================
\subsection{Aggregated Memory Atom Activations}
\label{supp:aggre_vis}
\begin{figure*}[t]
    \centering
    \includegraphics[width=0.9\linewidth]{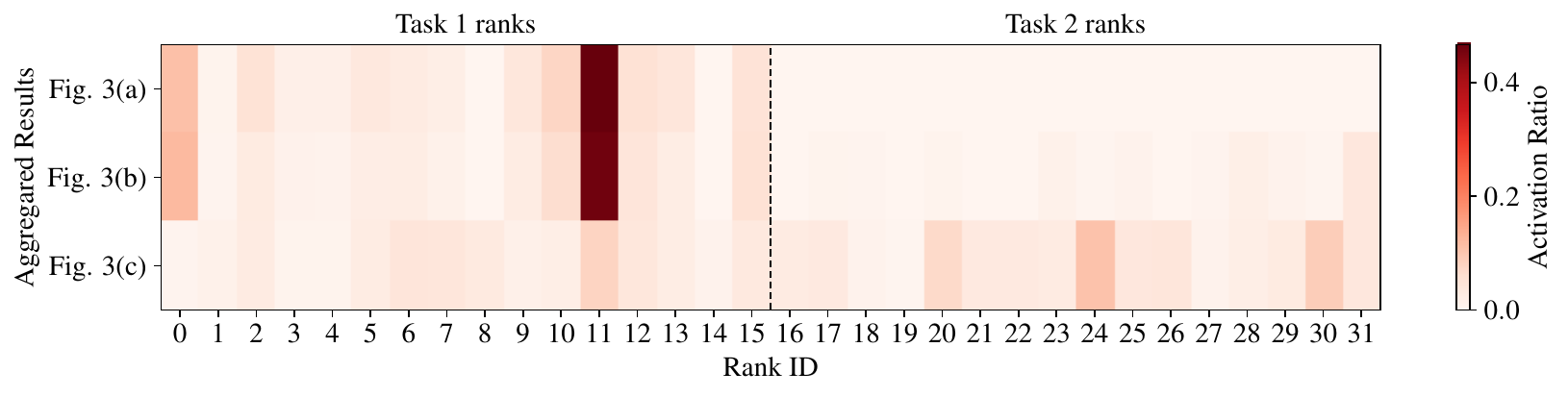}
    \caption{{Averaged activation ratio of each rank-1 expert across three scenarios in Fig.~\ref{fig:main_rank_visual}}}
    \label{supp:fig3_aggre}
\end{figure*}

To complement the qualitative examples in Fig.~\ref{fig:main_rank_visual}, we include a statistical aggregation of rank-1 atom utilization over the entire test set for each task. This heatmap (Fig.~\ref{supp:fig3_aggre}) provides a global view of the retrieval behavior and confirms that the patterns observed in Fig.~\ref{fig:main_rank_visual} are representative of the model's overall dynamics. The heatmap visualizes the activation ratio of each rank-1 atom (x-axis) across three scenarios (y-axis) in Fig.~\ref{fig:main_rank_visual}:
\begin{enumerate}
\item Task 1 Data (after Task 1): Consistent with the qualitative results in Fig.~\ref{fig:acti_1}, we observe statistically dominant usage of memory atom 0 (airplane semantics) and memory atom 11 (background/sky). Atom 11 shows higher overall activation frequency as it captures common background tokens, which constitute a larger portion of image patches than the object itself.
\item Task 1 Data (after Task 2): Crucially, the activation pattern remains virtually unchanged after training on Task 2. The heatmap shows near-zero activation for the newly introduced Task 2 atoms (atoms 16--31). This provides strong evidence that our retrieval mechanism is stable: ``old'' data does not drift to ``new'' atoms, effectively mitigating catastrophic forgetting.
\item Task 2 Data (after Task 2): We observe a distinct, dual-mode behavior:
\begin{enumerate}
\item Knowledge Reuse: The old atom 11 is reactivated, confirming that the model reuses the generic ``blue sky'' feature for the new task.
\item High Diversity: Unlike the concentrated pattern of Task 1 (Aircraft, with homogeneous airplane images), Task 2 (Caltech101, with 101 diverse categories) utilizes a broad spectrum of new atoms (e.g., atoms 20, 24, 29, 30). This aligns with our design goal: fine-grained atoms allow the model to dedicate different subspaces to the highly diverse semantics of the new task.
\end{enumerate}
\end{enumerate}
We explicitly chose to visualize activations at specific representative layers rather than averaging across the entire depth of the model. In modern deep models, different layers specialize in distinct feature types (e.g., low-level textures vs. high-level semantics). Averaging atom usage across all layers would smooth out these distinct signatures and obscure the specialized retrieval behavior we aim to demonstrate.

\subsection{Extended Memory Activation Maps}
\label{supp:rank_vis}

\begin{figure*}[t]
  \centering
  % first subfigure
  \begin{subfigure}[c]{\textwidth}
    \centering
    \begin{overpic}[width=0.92\linewidth]{figs/layer8_kproj_task1_1.pdf}
    \put(38,1){\color{orange}\framebox(3,8){}}
    \put(43,1){\color{orange}\framebox(7,8){}}
    \put(51,1){\color{orange}\framebox(5,8){}}
    \put(59,1){\color{orange}\framebox(4,8){}}
    \end{overpic}%
    \raisebox{0.1cm}{%
        \includegraphics[width=0.07\linewidth]{figs/layer8_kproj_task1_1.jpg}%
    }%
    \\
    \vspace{-0.2cm}
    \caption{Memory atom activations of \ours on data from Task 1 after learning Task 1.}
    \label{supp:acti_1}
  \end{subfigure}
  % second subfigure
  \begin{subfigure}[c]{\textwidth}
    \centering
    \begin{overpic}[width=0.92\linewidth]{figs/layer8_kproj_task2_1.pdf}
    \put(38,1){\color{orange}\framebox(3,15.4){}}
    \put(43,1){\color{orange}\framebox(7,15.4){}}
    \put(51,1){\color{orange}\framebox(5,15.4){}}
    \put(59,1){\color{orange}\framebox(4,15.4){}}
    \end{overpic}%
    \raisebox{0.6cm}{%
        \includegraphics[width=0.07\linewidth]{figs/layer8_kproj_task2_1.jpg}%
    }%
    \\ \vspace{-0.2cm}
    \caption{Memory atom activations of \ours on data from Task 1 after learning Task 2.}
    \label{supp:acti_2_2}
  \end{subfigure}
  % third subfigure
  \begin{subfigure}[c]{\textwidth}
    \centering
    \begin{overpic}[width=0.92\linewidth]{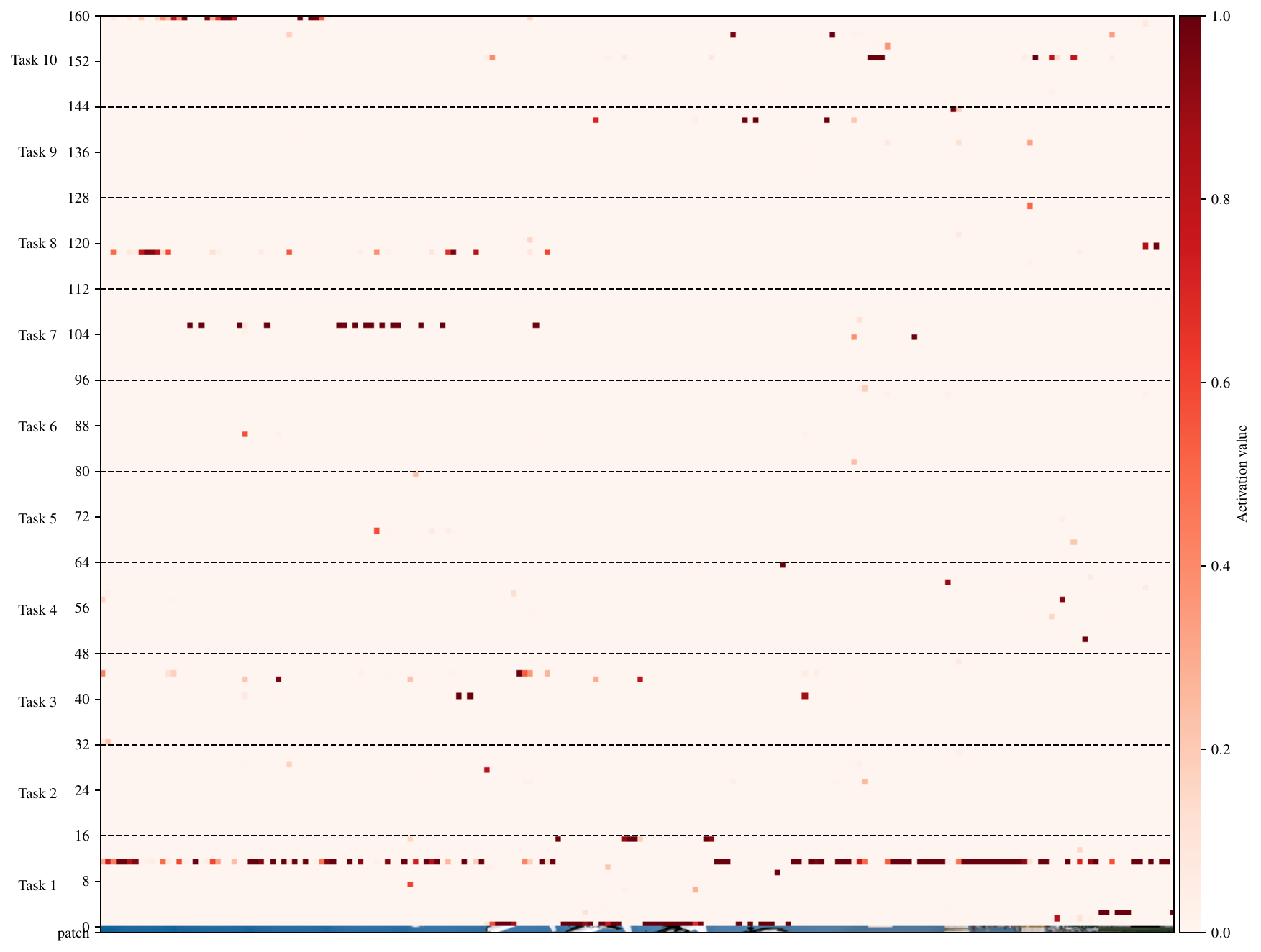}
    \put(38,1){\color{orange}\framebox(3,73.5){}}
    \put(43,1){\color{orange}\framebox(7,73.5){}}
    \put(51,1){\color{orange}\framebox(5,73.5){}}
    \put(59,1){\color{orange}\framebox(4,73.5){}}
    \end{overpic}%
    \raisebox{4.4cm}{%
        \includegraphics[width=0.07\linewidth]{figs/layer8_kproj_task1_1.jpg}%
    }%
    \\ \vspace{-0.2cm}
    \caption{Memory atom activations of \ours on data from Task 1 after learning Task 10.}
    \label{supp:acti_3}
  \end{subfigure}
    \caption{Extended view of Fig.~\ref{fig:main_rank_visual} illustrating \textbf{forgetting mitigation}. Regions corresponding to object semantics are highlighted with orange bounding boxes. Zoom in for details.}
      \label{fig:supp_rank_visual}
  \vspace{-0.3cm}
\end{figure*}

\begin{figure*}[t]
  \centering
  % first subfigure
  \begin{subfigure}[c]{\textwidth}
    % \centering
    % \includegraphics[width=0.92\linewidth]{figs/layer8_kproj_task1_1.pdf}
    \begin{overpic}[width=0.92\linewidth]{figs/layer8_kproj_task1_1.pdf}
    \put(8,1){\color{blue}\framebox(30,8){}}
    \put(62,1){\color{blue}\framebox(20,8){}}
    % \put(80,1){\color{blue}\framebox(30,8){}}
    \end{overpic}%
    \raisebox{0.1cm}{%
        \includegraphics[width=0.07\linewidth]{figs/layer8_kproj_task1_1.jpg}%
    }%
    \caption{Memory atom activations of \ours on data from Task 1 after learning Task 1.}
    \label{supp:acti_1_2}
  \end{subfigure}
  % second subfigure
  \begin{subfigure}[c]{\textwidth}
    \centering
    \begin{overpic}[width=0.92\linewidth]{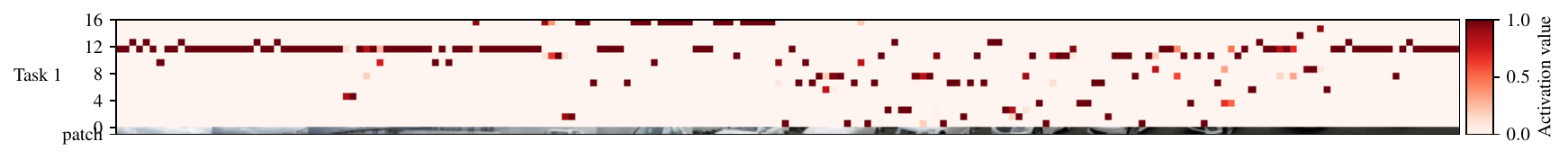}
    \put(8,1){\color{blue}\framebox(27,8){}}
    \end{overpic}%
    \raisebox{0.1cm}{%
        \includegraphics[width=0.07\linewidth]{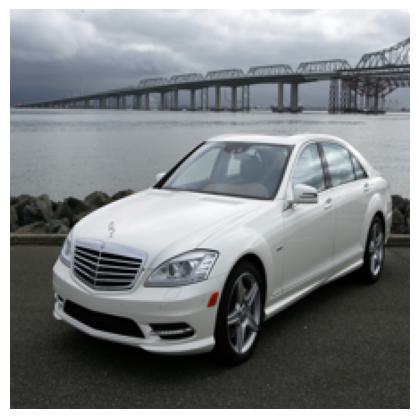}%
    }%
    \caption{Memory atom activations of \ours on data from Task 9 after learning Task 1.}
    \label{supp:acti_2}
  \end{subfigure}
  % third subfigure
  \begin{subfigure}[c]{\textwidth}
    \centering
    \begin{overpic}[width=0.92\linewidth]{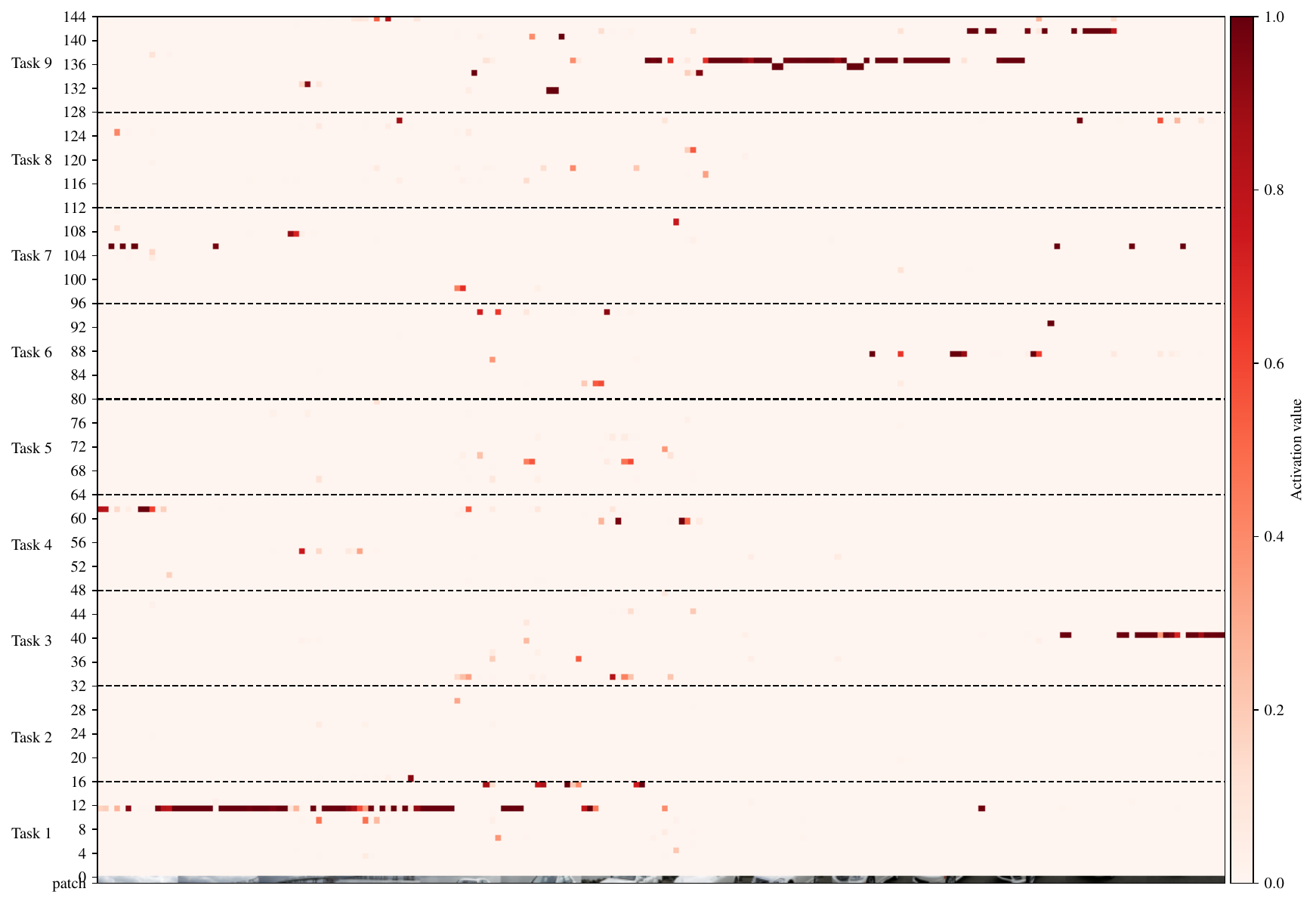}
    \put(8,1){\color{blue}\framebox(27,8){}}
    \end{overpic}%
    \raisebox{4.cm}{%
        \includegraphics[width=0.07\linewidth]{figs/layer8_kproj_task1_9.jpg}%
    }%
    \caption{Memory atom activations of \ours on data from Task 9 after learning Task 9.}
    \label{supp:acti_3_2}
  \end{subfigure}
\caption{Extended view of Fig.~\ref{fig:main_rank_visual} illustrating \textbf{knowledge reuse}. Regions corresponding to generic input tokens (\textit{e.g.} blue sky) are highlighted with blue bounding boxes. Zoom in for details.}
\label{fig:supp_rank_visual2}
  \vspace{-0.3cm}
\end{figure*}

In Sec.~\ref{sec:rank_vis} (Fig.~\ref{fig:main_rank_visual}), we illustrated memory atom activations during the learning of Task 1 and Task 2. Here, we extend these visualizations to additional tasks and scenarios in Fig.~\ref{fig:supp_rank_visual} and Fig.~\ref{fig:supp_rank_visual2}.

\noindent\textbf{\ours retains task-specific semantics without forgetting.}
Fig.~\ref{fig:supp_rank_visual} shows the activation maps for the same Task 1 image after training on Task 1 (a), Task 2 (b), and the last Task 10 (c). Patches corresponding to the airplane object are outlined in orange. In all three snapshots, the atom at index 0 remains consistently and exclusively activated for those airplane patches, demonstrating that \ours has effectively stored the airplane-specific knowledge in memory atom 0. Even after 10 subsequent tasks, this pattern remains unchanged, indicating that later updates do not overwrite or interfere with the learned airplane representations. In other words, \ours effectively memorizes and preserves task-relevant semantics, thereby mitigating catastrophic forgetting.

\noindent\textbf{\ours encodes generic semantics that are reused across tasks.}
Fig.~\ref{fig:supp_rank_visual2} examines an input image from Task 9 before and after learning Task 9. Panel (a) shows the activation map of data from Task 1 after learning Task 1: memory atom 11 (outlined in blue) already responds strongly to sky-background patches, demonstrating that \ours has stored a generic ``blue sky'' concept in this atom. In panel (b), when we infer on the Task 9 image before training on Task 9, atom 11 is again activated for the sky regions, confirming that \ours reuses this shared knowledge for unseen data. Finally, panel (c) shows the activation map after learning Task 9: atom 11 remains dedicated to the sky background, while newly initialized atoms specialize in the ``car'' object semantics. This persistent reuse of atom 11 across tasks illustrates \ours's ability to capture and retain common features as reusable memory slots, reducing redundancy and facilitating knowledge reuse.

\subsection{Statistical Analyses of Contributing Memory Atom Activations}
\label{supp:rank_stat}
\begin{figure*}
    \centering
    \includegraphics[width=0.32\linewidth]{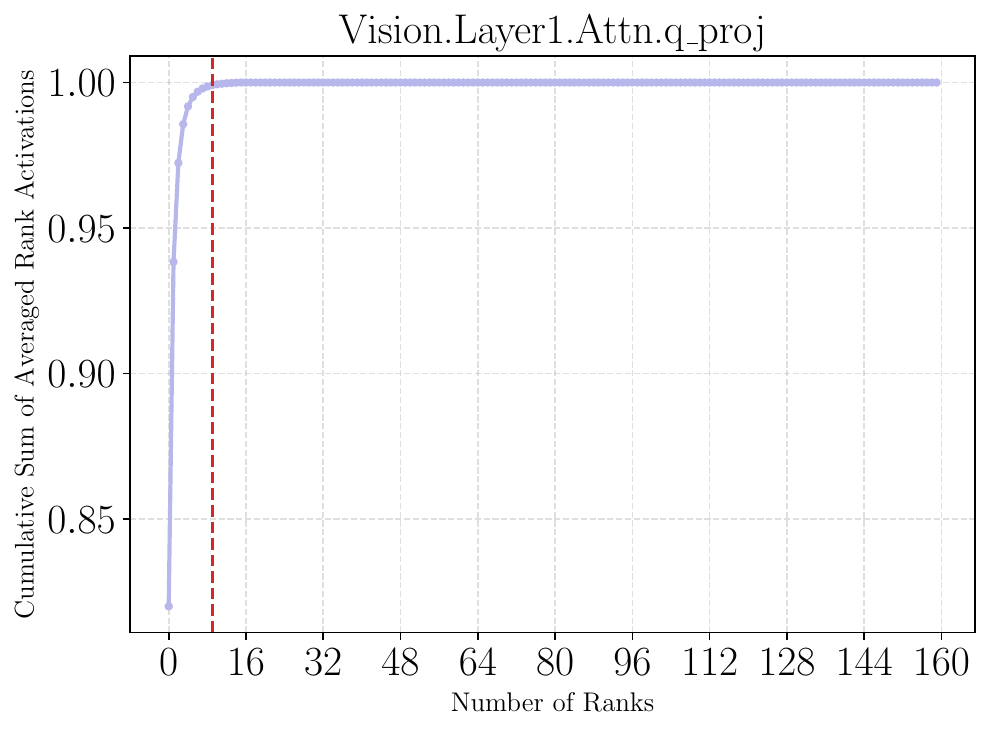} \hfill
        \includegraphics[width=0.32\linewidth]{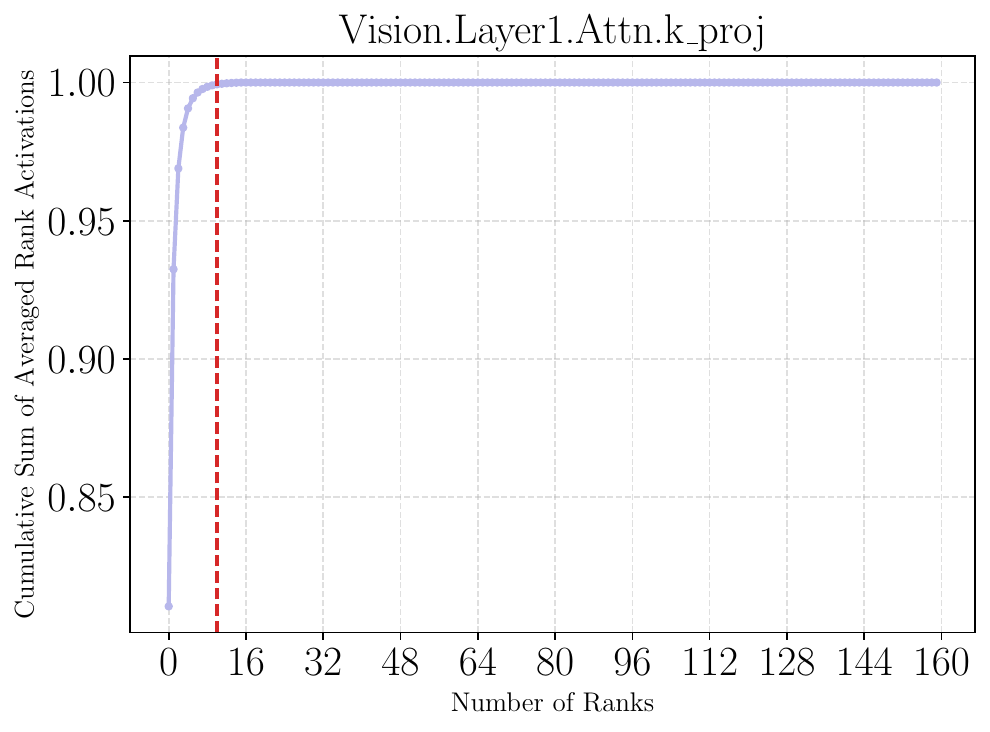} \hfill
    \includegraphics[width=0.32\linewidth]{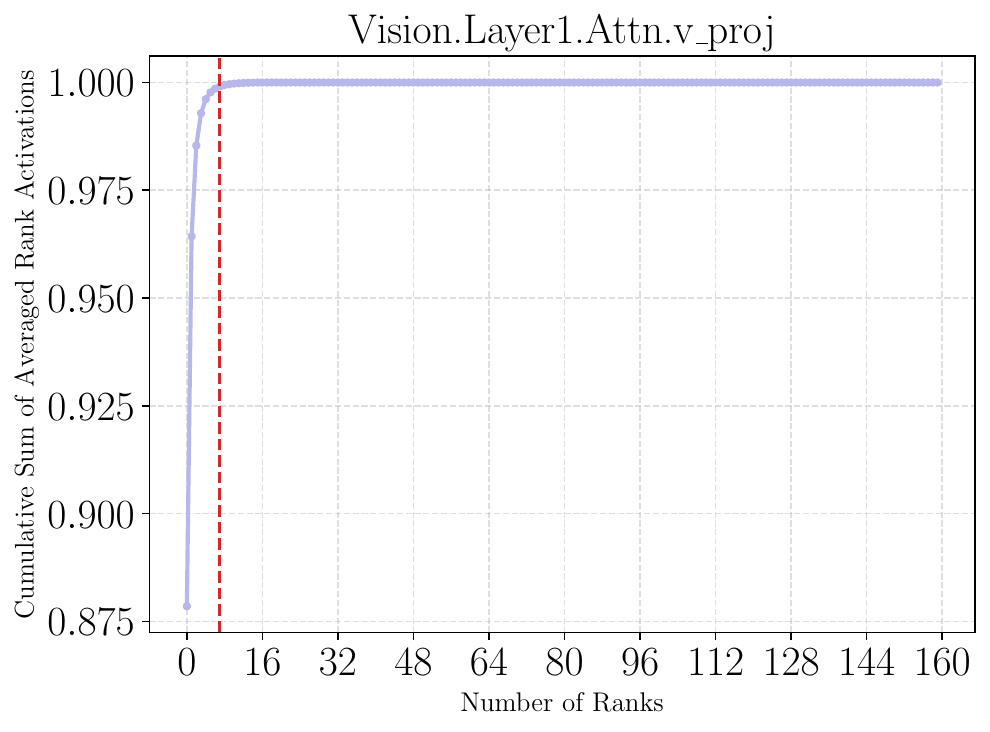}
    \includegraphics[width=0.32\linewidth]{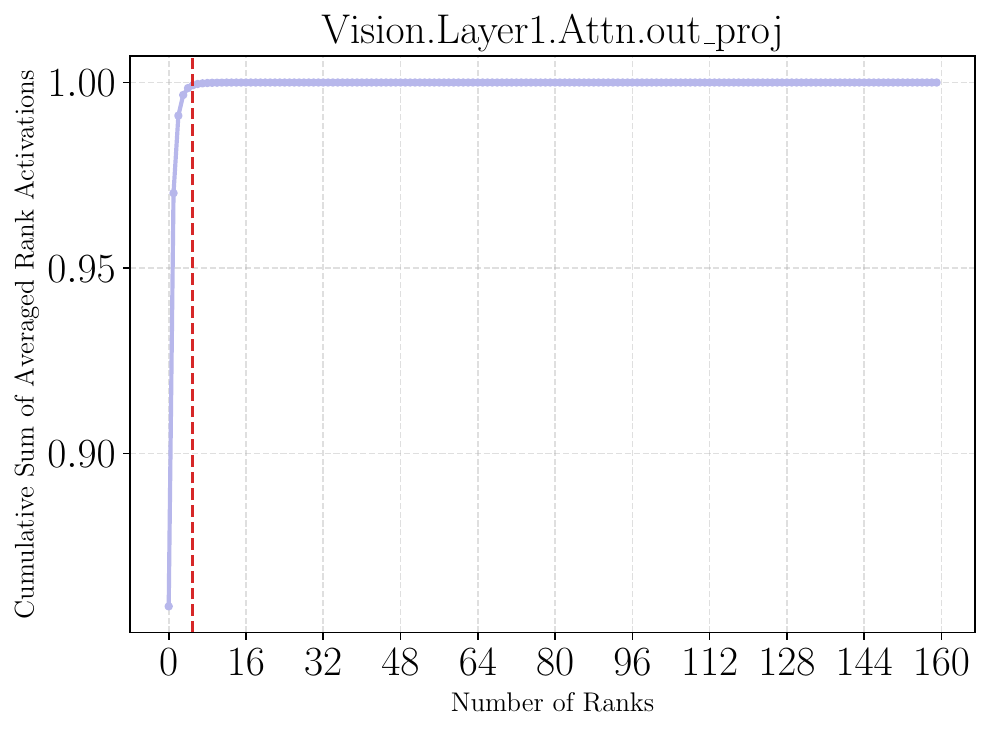} \hfill
    \includegraphics[width=0.32\linewidth]{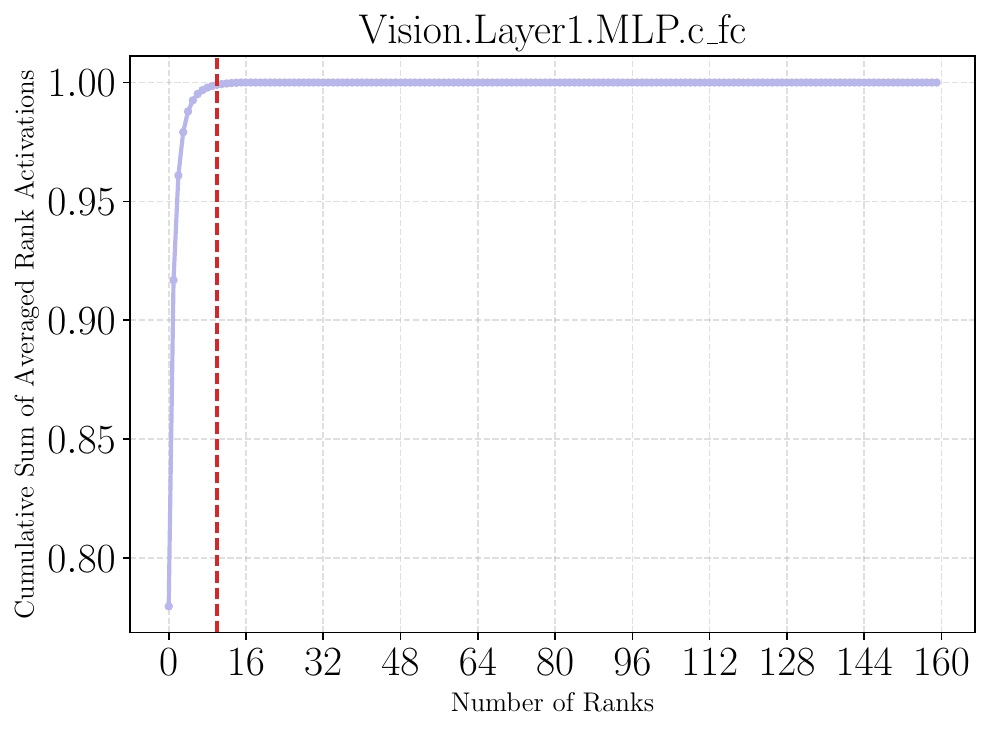} \hfill
    \includegraphics[width=0.32\linewidth]{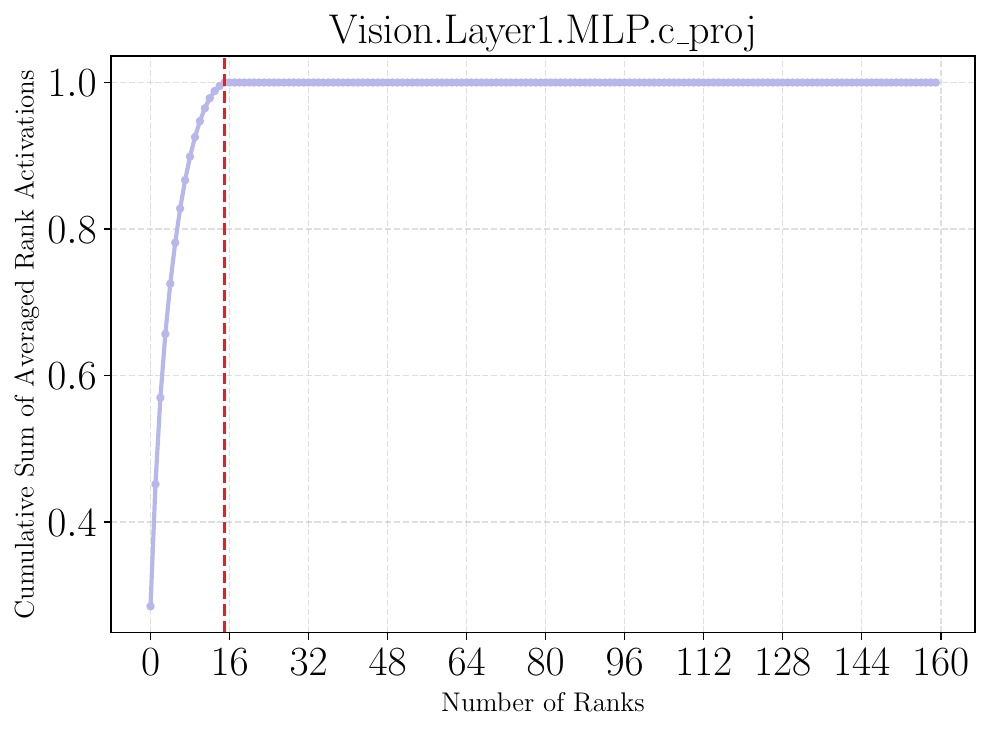}
    \includegraphics[width=0.32\linewidth]{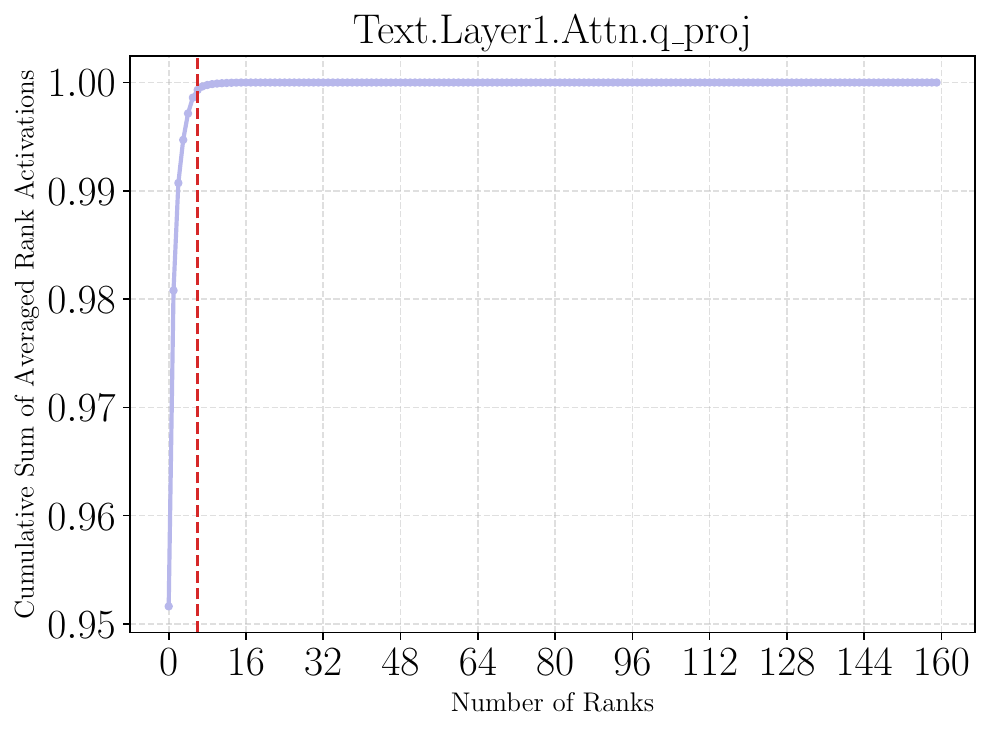} \hfill
    \includegraphics[width=0.32\linewidth]{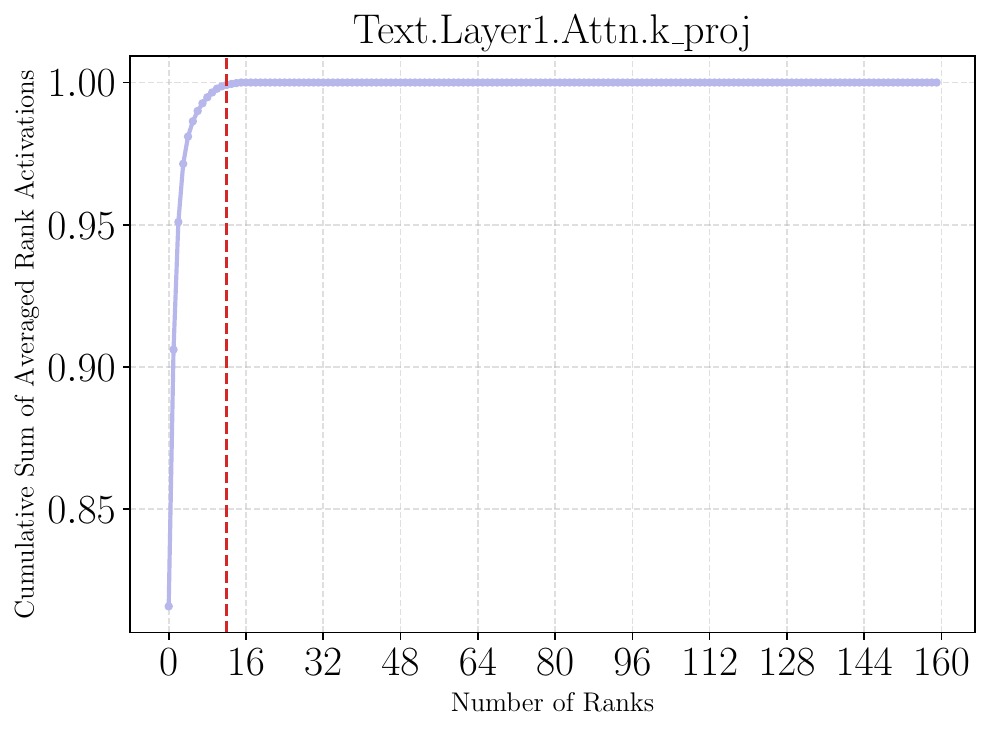} \hfill
    \includegraphics[width=0.32\linewidth]{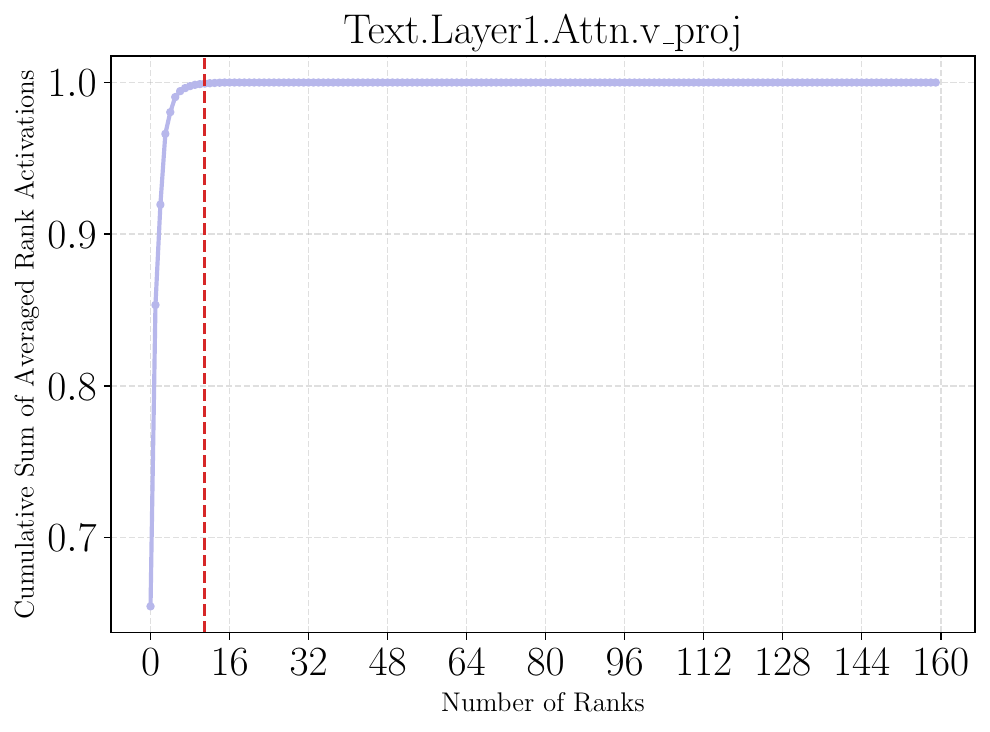}
    \includegraphics[width=0.32\linewidth]{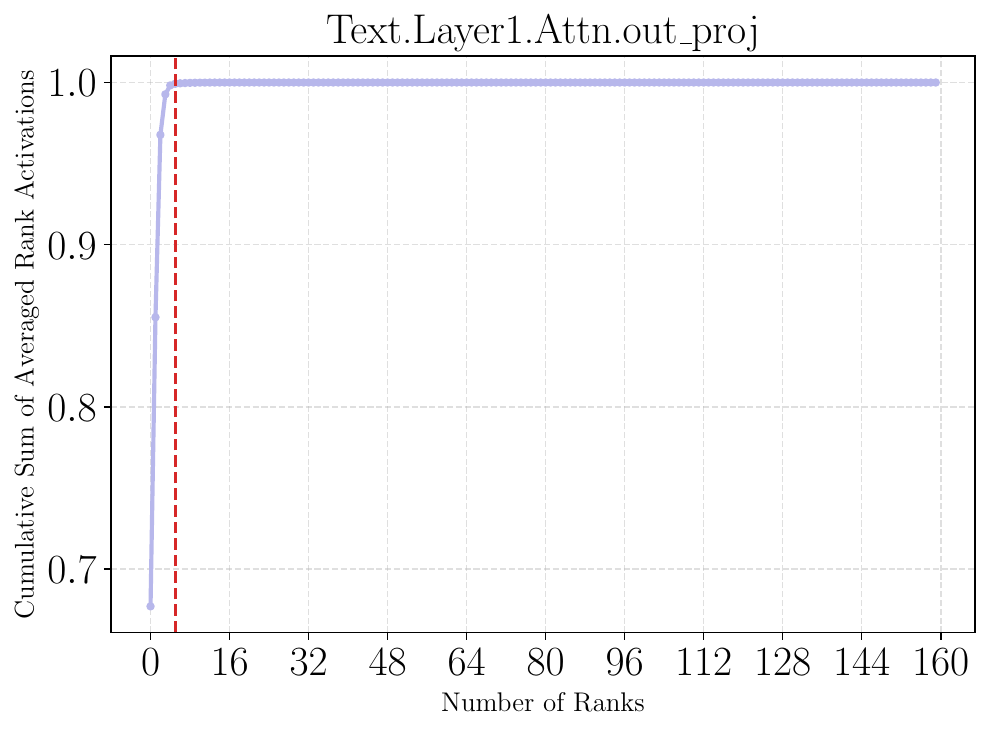} \hfill
    \includegraphics[width=0.32\linewidth]{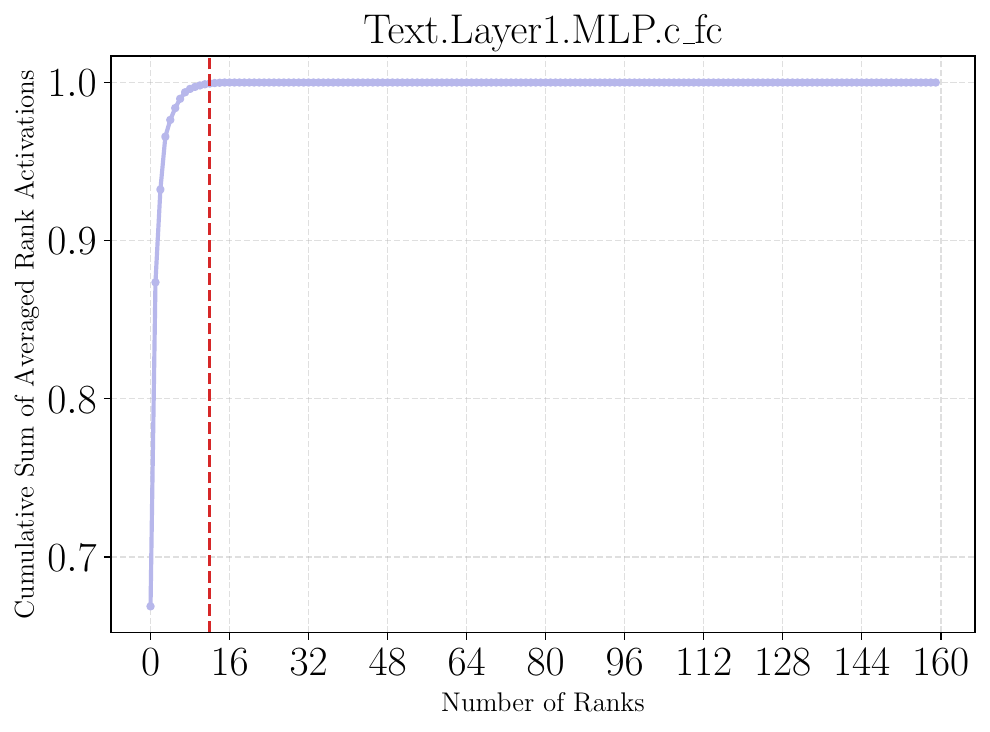} \hfill
    \includegraphics[width=0.32\linewidth]{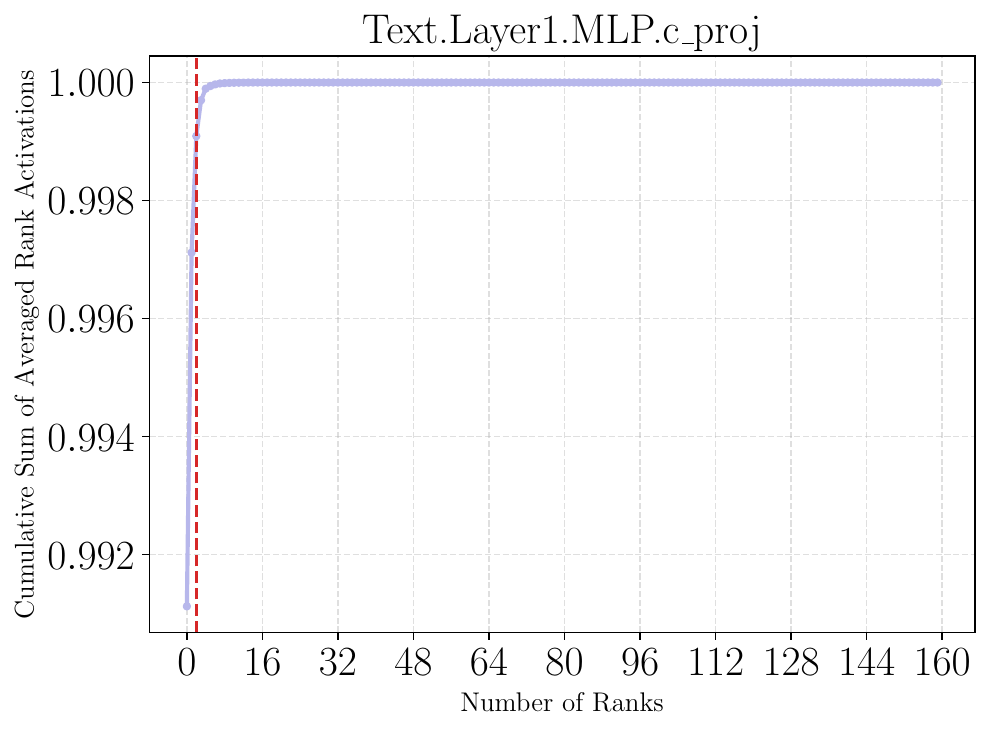}
    \caption{Statistical analyses on the number of atoms required to capture 99\% of cumulative sum (indicated in red dashed line) of all memory atom activations. Activations were gathered from the model after training on all tasks, and results are shown for a representative selection of layers and positions within the pre-trained model.}
    \label{fig:supp_rank_stat}
\end{figure*}

\begin{figure*}
    \centering
    \includegraphics[width=0.32\linewidth]{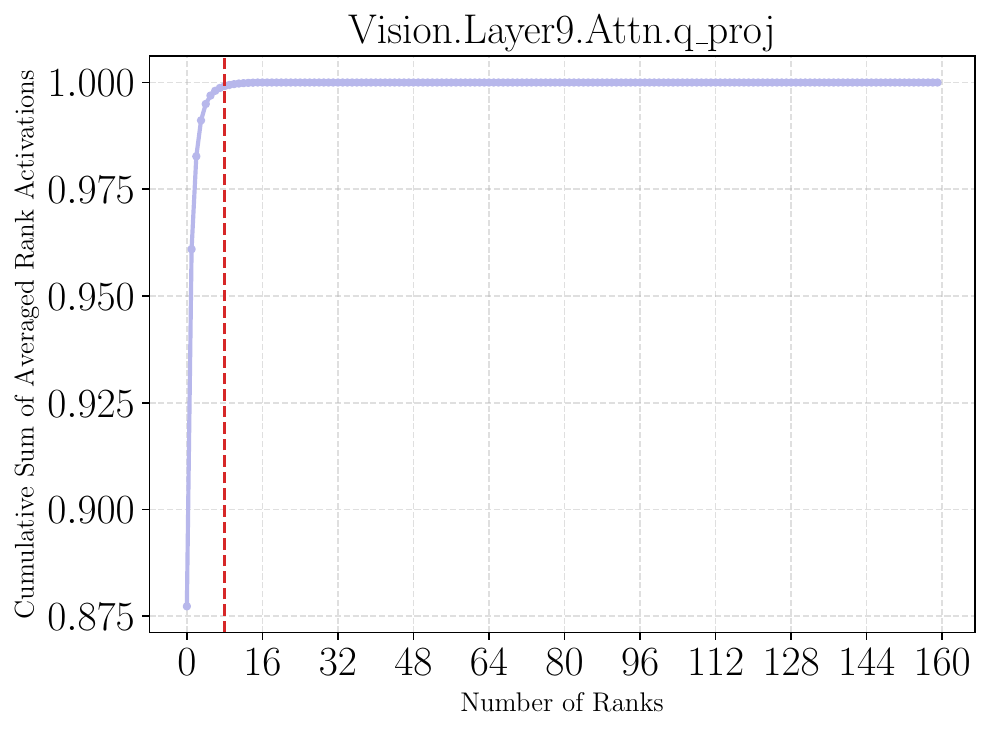} \hfill
        \includegraphics[width=0.32\linewidth]{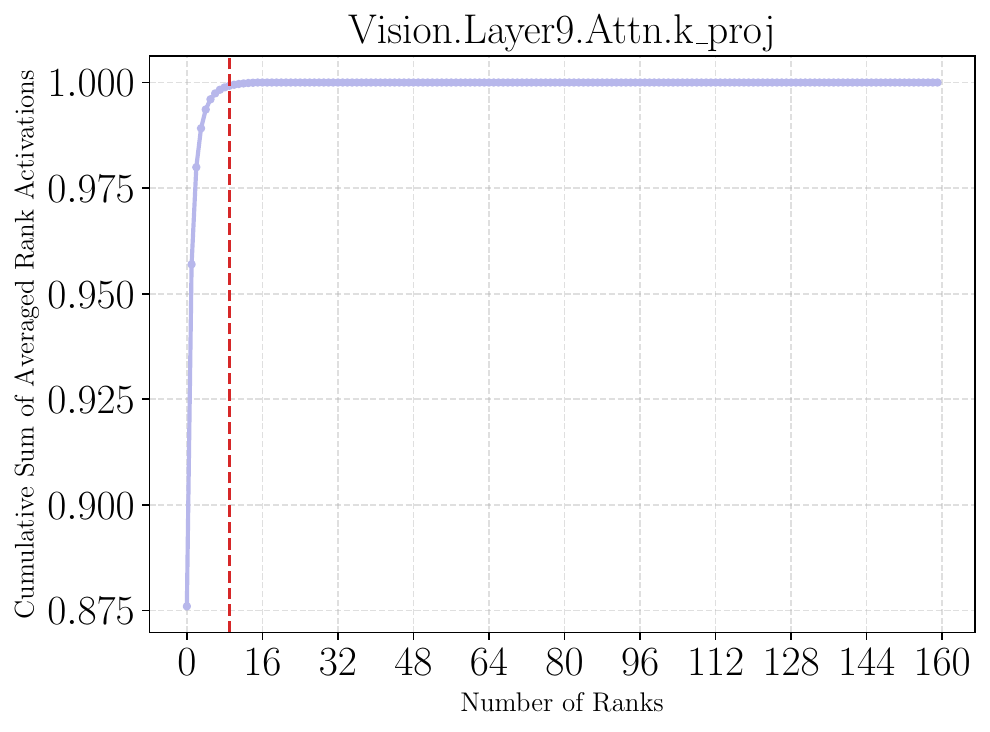} \hfill
    \includegraphics[width=0.32\linewidth]{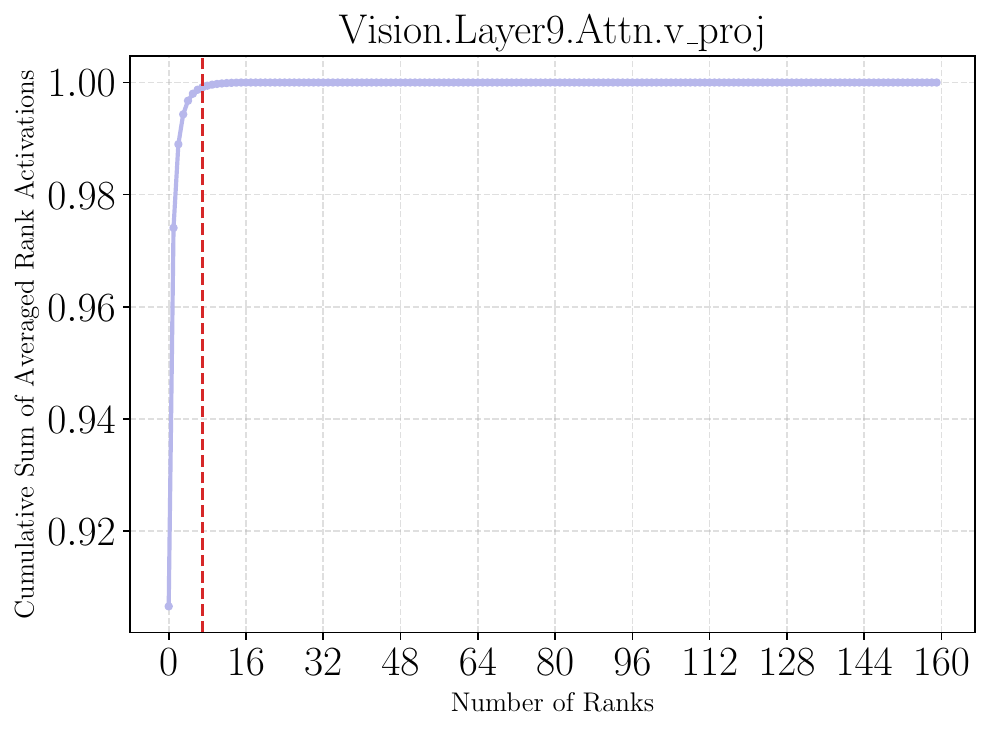}
    \includegraphics[width=0.32\linewidth]{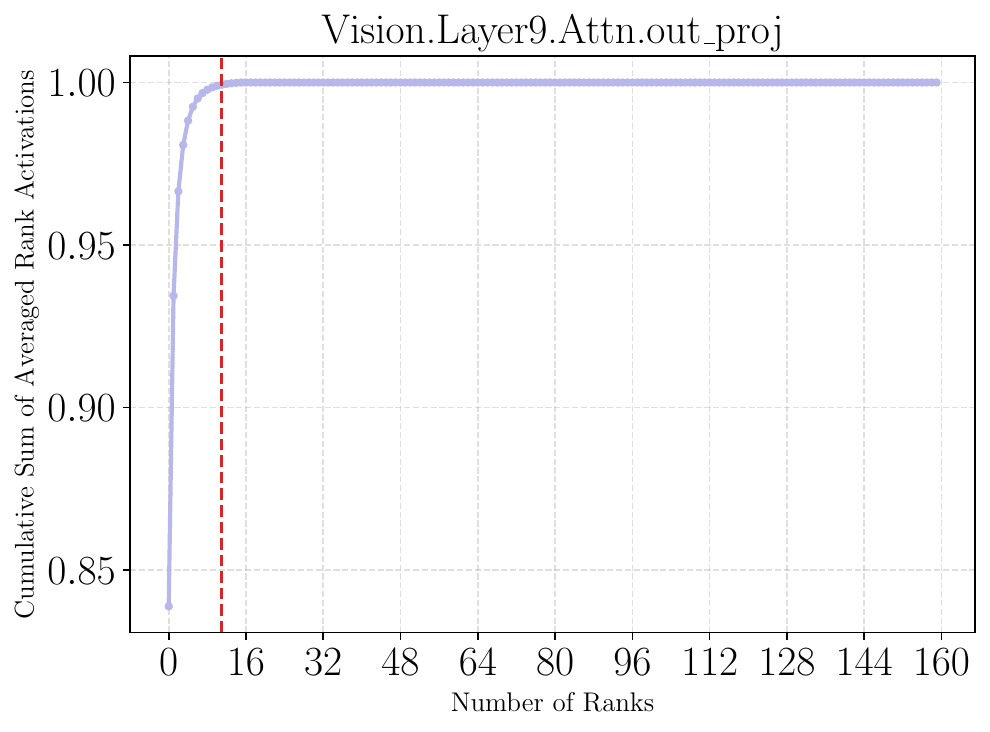} \hfill
    \includegraphics[width=0.32\linewidth]{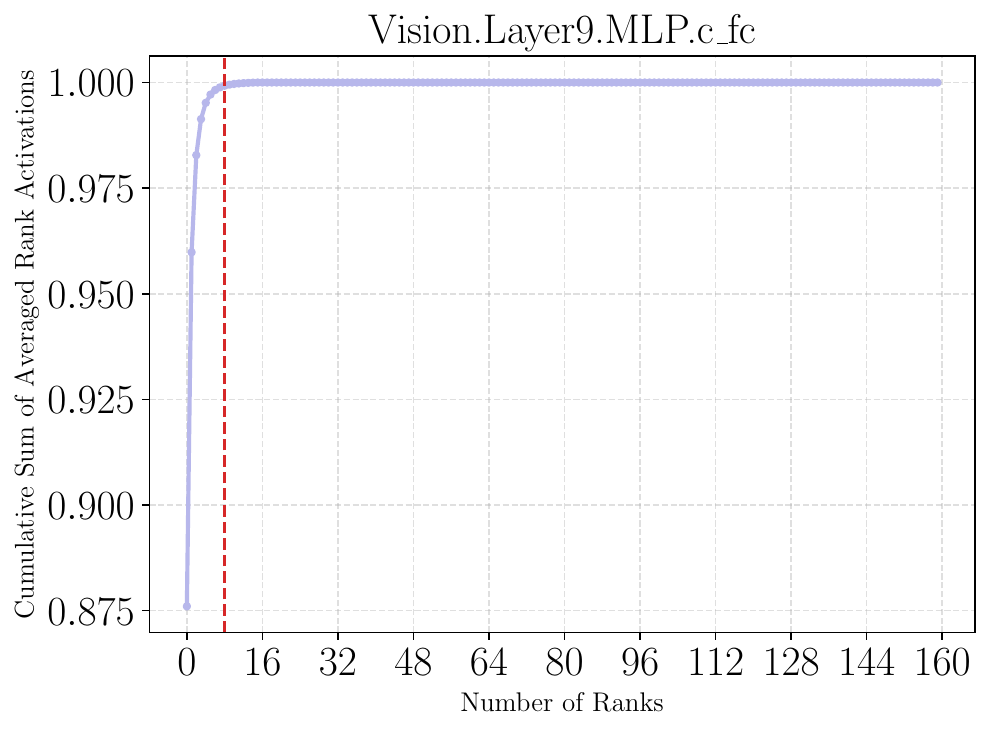} \hfill
    \includegraphics[width=0.32\linewidth]{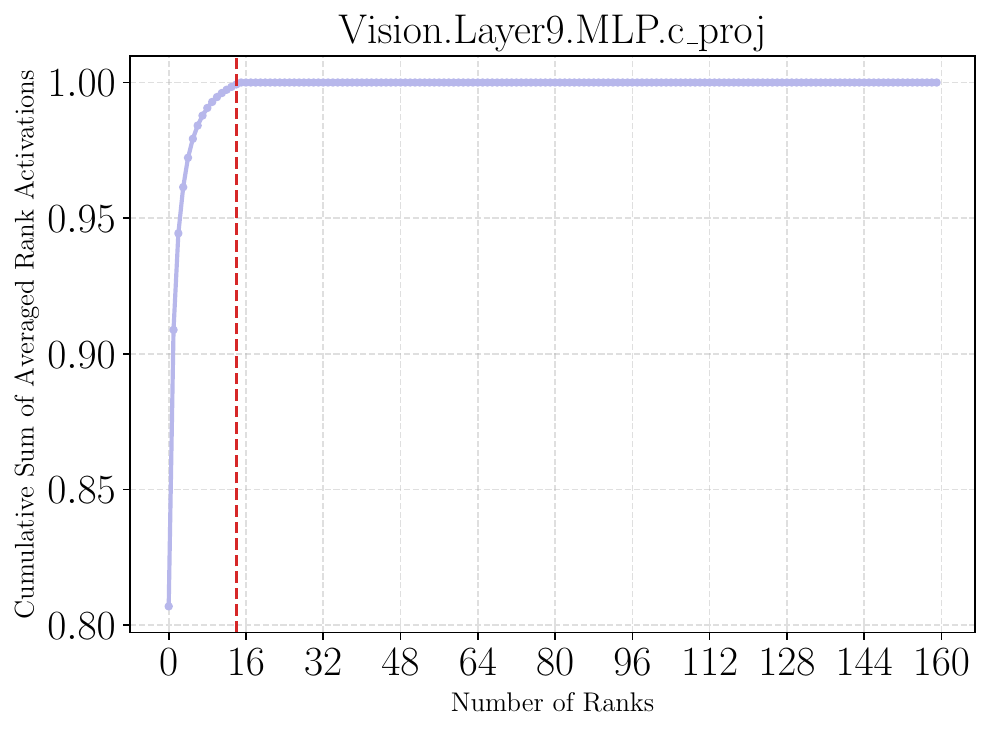}
    \includegraphics[width=0.32\linewidth]{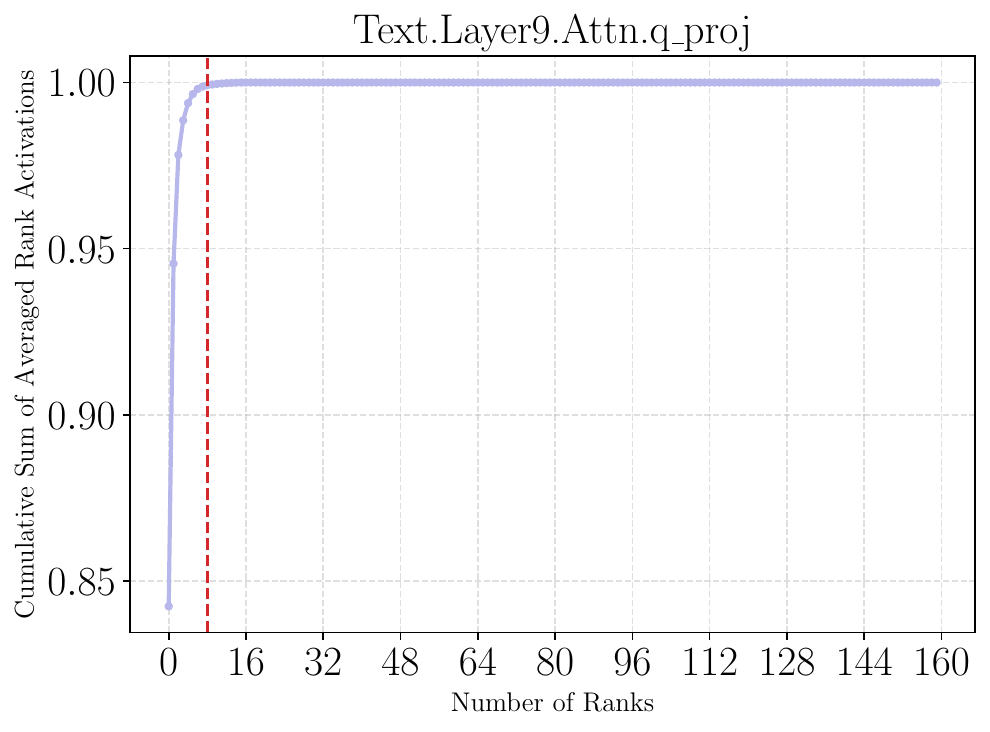} \hfill
    \includegraphics[width=0.32\linewidth]{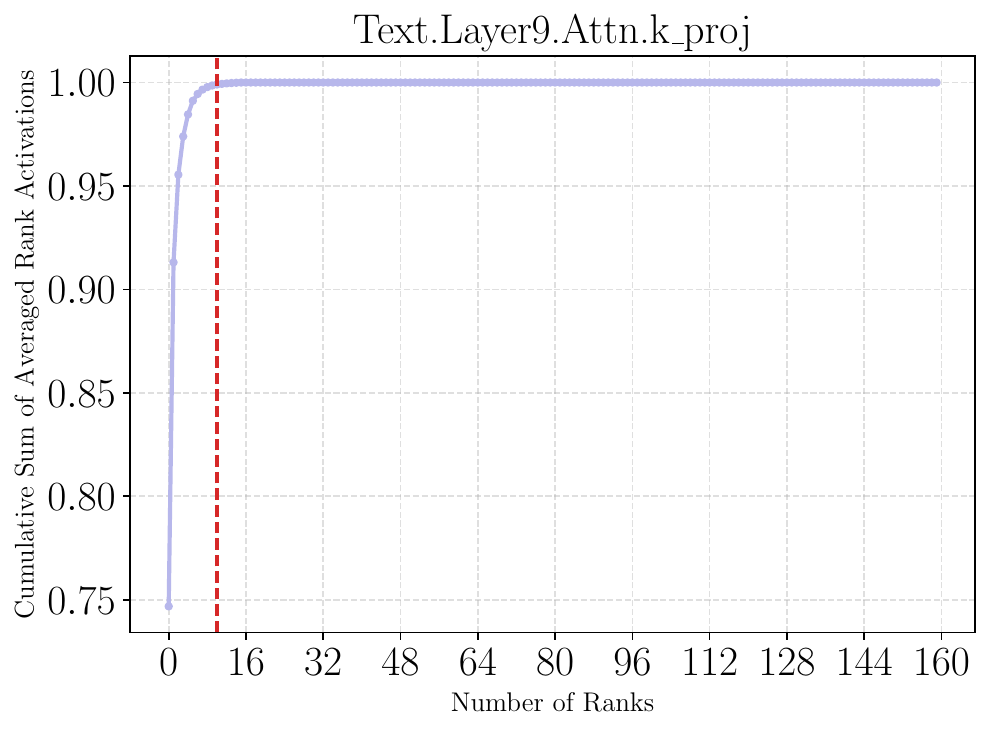} \hfill
    \includegraphics[width=0.32\linewidth]{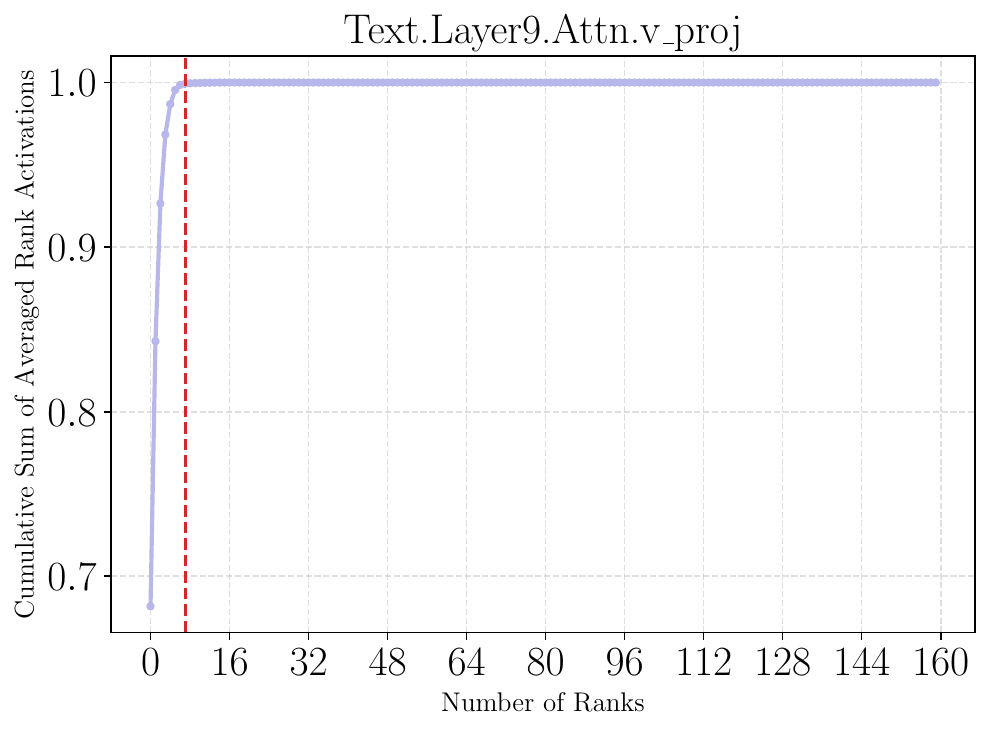}
    \includegraphics[width=0.32\linewidth]{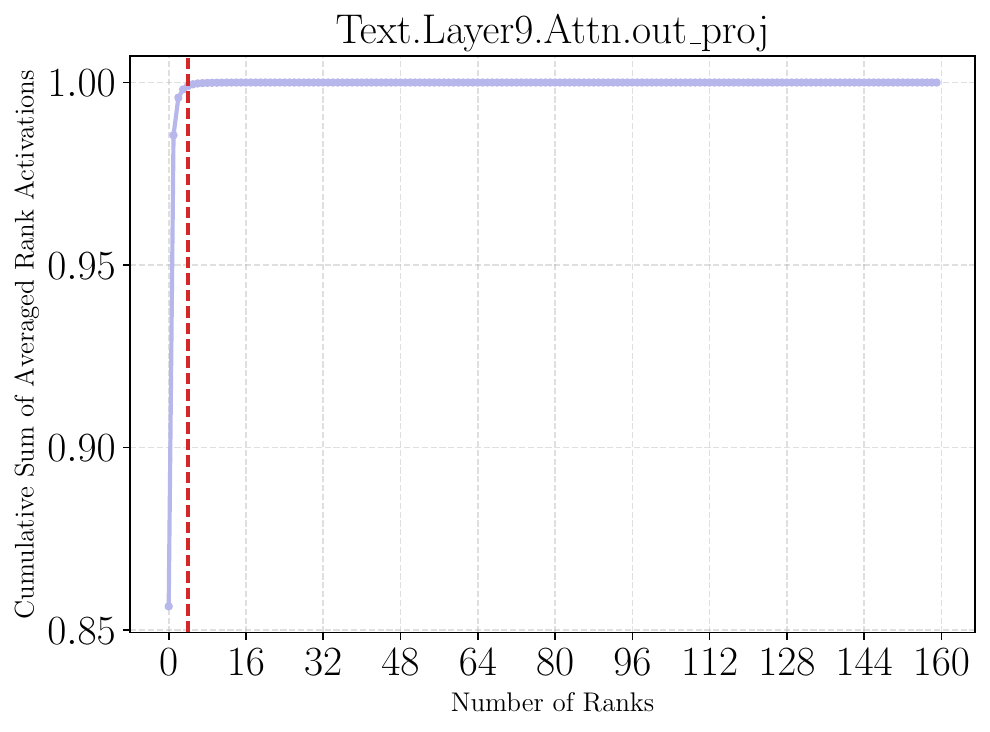} \hfill
    \includegraphics[width=0.32\linewidth]{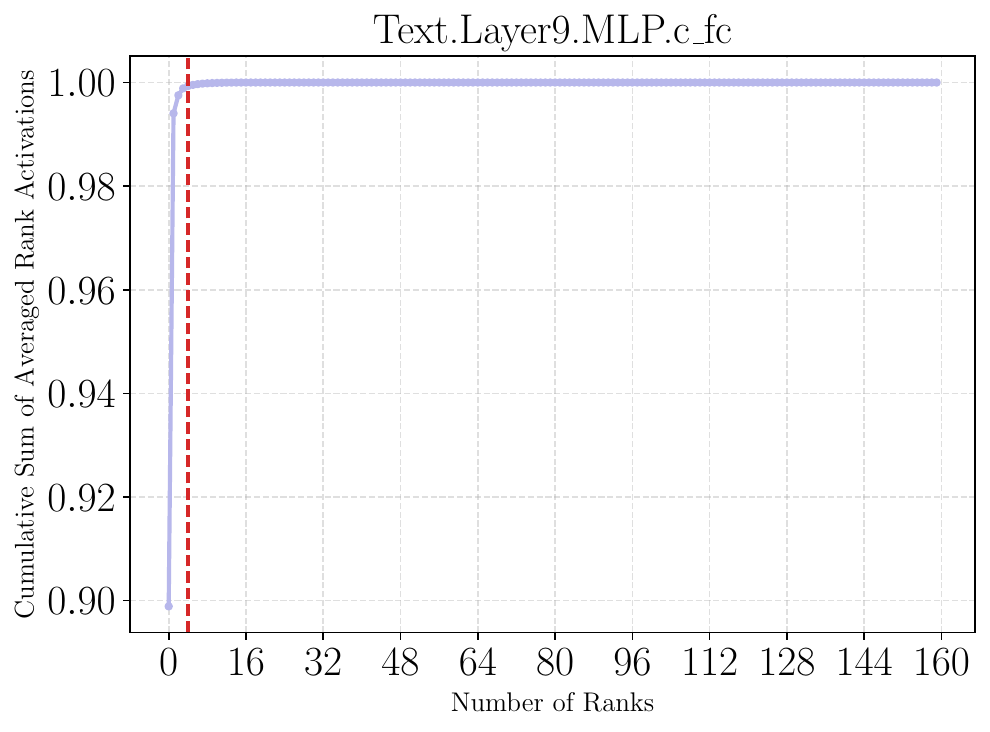} \hfill
    \includegraphics[width=0.32\linewidth]{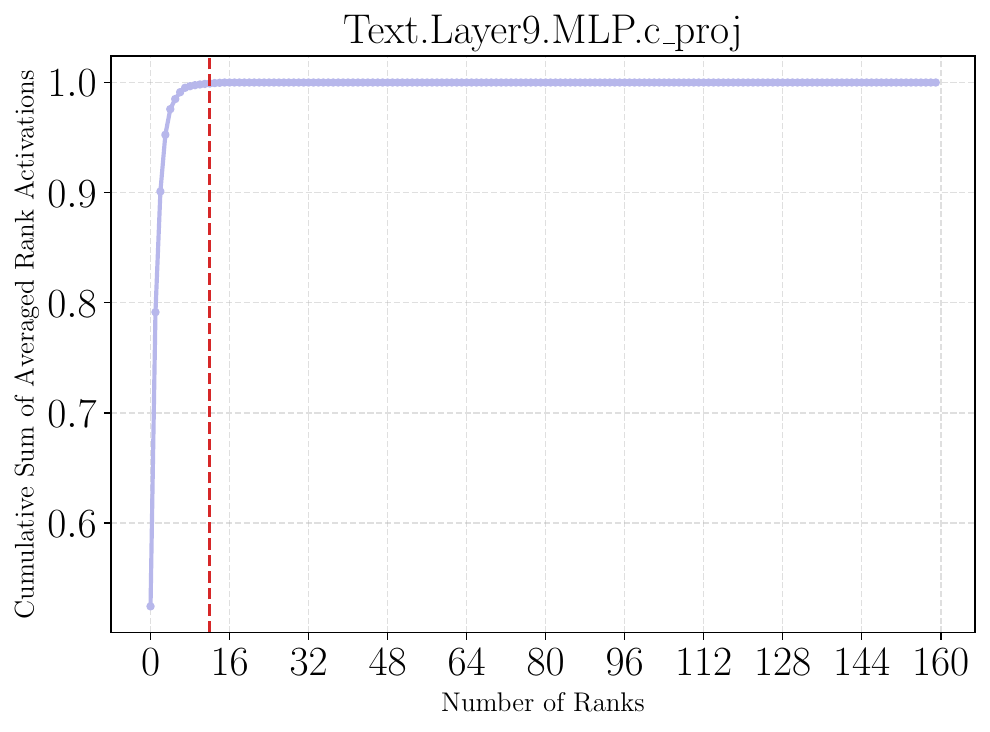}
    \caption{Statistical analyses on the number of atoms required to capture 99\% of cumulative sum (indicated in red dashed line) of all memory atom activations. Activations were gathered from the model after training on all tasks, and results are shown for a representative selection of layers and positions within the pre-trained model.}
    \label{fig:supp_rank_stat2}
\end{figure*}

Fig.~\ref{fig:supp_rank_stat} and Fig.~\ref{fig:supp_rank_stat2} plot the cumulative sum of averaged atom activations after training on all tasks, sorted in descending order, for several representative layers and locations within pre-trained models. The red dashed line marks the point at which 99\% of the total activation mass is reached, allowing us to quantify how many atoms truly contribute to the model's adaptation. Two key observations emerge:

\noindent\textbf{1. Sparse Mixture: only a small subset of atoms is needed.} Across all layers and positions, we find that fewer than 10\% of the total atoms suffice to capture 99\% of the activations. This highlights the extreme sparsity of \ours's self-activated mixture: most atoms remain dormant for any given input, while a compact set of highly relevant atoms drives the adaptation.

\noindent\textbf{2. Adaptive Activation: the number of active atoms varies across layers and modules.}
The number of atoms needed to capture 99\% of the cumulative activation mass varies across both layer depth and module type. For example, in Fig.~\ref{fig:supp_rank_stat}, the MLP's output projection (\texttt{c-proj}) in Layer 1 of the vision encoder requires 16 atoms, whereas the same module in Layer 1 of the text encoder needs only 3.

\begin{figure*}
    \centering
\begin{subfigure}[r]{\textwidth}
    \includegraphics[width=0.9\linewidth]{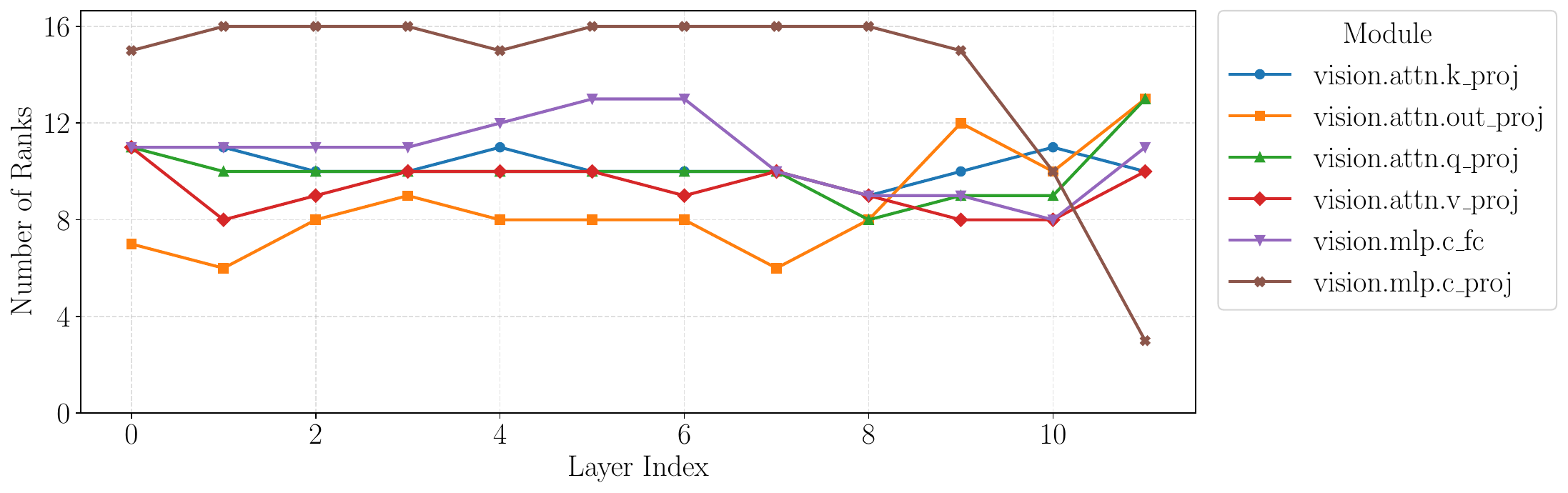}
    \caption{Vision Encoder}
\end{subfigure}
    \begin{subfigure}[c]{\textwidth}
    \includegraphics[width=0.9\linewidth]{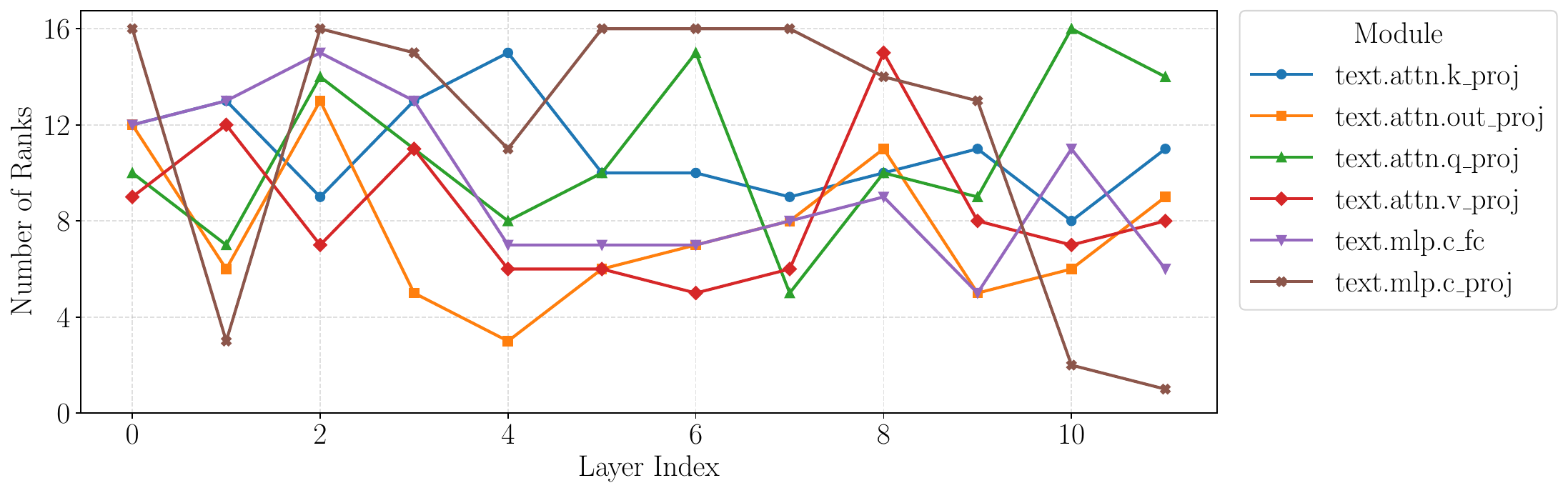}
    \caption{Text Encoder}
\end{subfigure}
\caption{Required atoms to capture 99 \% of cumulative activations, shown across different pre-trained model layers and projection locations.}    
\label{fig:supp_topr}
\end{figure*}

To provide a broader view, Fig.~\ref{fig:supp_topr} shows the required atom counts for every module in the pre-trained model. We observe that most attention modules require around 6--12 atoms, while the second MLP projection generally demands more atoms in early layers, peaking in the first few blocks, and then steadily declining in deeper layers.

Coupling the atom activation budget with atom pruning, \ours adapts the number of atoms activated at each layer and module. This adaptive sparsity maximizes the efficient use of newly acquired knowledge during continual learning.

\end{document}